%% file: main.tex
\documentclass[preprint,5p,times,authoryear]{elsarticle}

\usepackage{amssymb}
\usepackage{amsmath}

\usepackage{setspace}
\usepackage{epstopdf}
\usepackage{amsfonts}
\usepackage{mathtools}
\usepackage{sansmath}
\usepackage{algorithm}
\usepackage{algpseudocode}

\usepackage{tabularx}
\usepackage{booktabs} 
\usepackage{siunitx}
\robustify\bfseries  
\usepackage{multirow} 
\usepackage{array}
\usepackage{multicol}
\usepackage{dcolumn}
\usepackage{microtype}
\usepackage{url}
\usepackage{makecell} 
\usepackage{graphicx}
\usepackage[dvipsnames]{xcolor}
\usepackage[table]{xcolor}
\usepackage{subcaption} 
\usepackage{threeparttable}  
\usepackage[labelsep=period]{caption}  
\usepackage[british]{babel} 
\usepackage[hang]{footmisc}
\usepackage[
    colorlinks=true, 
    linkcolor=blue, 
    breaklinks=true, 
    urlcolor=blue,
]{hyperref}
\usepackage[capitalize,noabbrev,nameinlink]{cleveref} 
\usepackage{csquotes}

\usepackage{pgfplots,pgfplotstable}
\pgfplotsset{compat=1.18}

\definecolor{colorBE}{HTML}{1f77b4}
\definecolor{colorEE}{HTML}{ff7f0e}
\definecolor{colorLV}{HTML}{2ca02c}
\definecolor{colorPT}{HTML}{d62728}
\definecolor{colorSK}{HTML}{9467bd}
\definecolor{colorLT}{HTML}{8c564b}
\definecolor{colorSI}{HTML}{e377c2}
\definecolor{colorES}{HTML}{17becf}

\usepackage{enumerate}

\usepackage{tumabbrev}

\usepackage{tikz}
\usetikzlibrary{calc, positioning}
\usepackage{pgfplots,pgfplotstable}
\usepgfplotslibrary{groupplots}
\pgfplotsset{compat=newest}
\usepgfplotslibrary{fillbetween}
\usepackage[
  acronym,
  automake,
]{glossaries-extra}
\makeglossaries

\setabbreviationstyle[acronym]{long-short-user}
\renewcommand*{\glsxtruserparen}[2]{
  \glsxtrfullsep{#2}%
  \glsxtrparen
   {#1\ifglshasfield{\glsxtruserfield}{#2}{;\xspace%
     \expandafter\citealp\expandafter{\glscurrentfieldvalue}%
   }{}%
   }%
}

\glsdefpostdesc{acronym}{
 \ifglshasfield{\glsxtruserfield}{\glscurrententrylabel}%
 {~\expandafter\citep\expandafter{\glscurrentfieldvalue}}%
 {}%
}

\glsdisablehyper
\defglsentryfmt[acronym]{\ifglsused{\glslabel}{\glsgenentryfmt}{\emph{\glsgenentryfmt}}}

\glsdisablehyper
\defglsentryfmt{\ifglsused{\glslabel}{\glsgenentryfmt}{\emph{\glsgenentryfmt}}}

\input{./acronyms}

\usepackage{xspace}
\usepackage{relsize}  

\usepackage{placeins}
\usepackage{float}

\definecolor{mediummidnightblue}{RGB}{72,121,195}
\definecolor{lightseagreen}{RGB}{62,186,103}

\newcommand{\dataset}[1]{\textsc{#1}\xspace}
\newcommand{\class}[1]{{\smaller{\texttt{#1}}}\xspace}%
\newcommand{\EuroCrops}{\dataset{EuroCrops}}
\newcommand{\EuroCropsML}{\dataset{EuroCropsML}}
\newcommand{\EuroCropsMLNew}{\dataset{EuroCropsML 2.0}}
\newcommand{\maize}{\class{grain maize corn popcorn}}
\newcommand{\meadow}{\class{pasture meadow grassland grass}}
\newcommand{\winterwheat}{\class{winter common soft wheat}}
\newcommand{\forest}{\class{tree wood forest}}
\newcommand{\grapes}{\class{vineyards wine vine rebland grapes}}
\newcommand{\wheat}{\class{wheat}}
\newcommand{\parsley}{\class{parsley}}

\newcommand{\channel}[1]{{\small{\texttt{#1}}}\xspace}%
\newcommand{\Bone}{\channel{B01}}

\newcommand{\Bnine}{\channel{B09}}
\newcommand{\Bten}{\channel{B10}}

\newcommand{\colorblue}{\color{blue}}

\newcommand{\colorteal}{\color{teal}}

\newcolumntype{X}{>{\scriptsize}l}
\newcolumntype{L}{>{\scriptsize\raggedright\arraybackslash}X}

\begin{document}

\begin{frontmatter}



\title{DirPA: Addressing Prior Shift in Imbalanced Few-shot Crop-type Classification} 

\author[tum]{Joana Reuss\corref{cor1}}
\author[tum]{Ekaterina Gikalo}
\author[tum,mdsi,ellis]{Marco K\"orner}
\affiliation[tum]{
    organization={Technical University of Munich (TUM), TUM School of Engineering and Design, Department of Aerospace and Geodesy, Chair of Remote Sensing Technology}, 
    city={Munich},
    postcode={80333}, 
    country={Germany}
}
\affiliation[mdsi]{
    organization={Technical University of Munich (TUM), Munich Data Science Institute (MDSI)}, 
    city={Garching},
    postcode={85748}, 
    country={Germany}
}
\affiliation[ellis]{
    organization={ELLIS Unit Jena, University of Jena, Jena, Germany}, 
    city={Jena},
    postcode={07743}, 
    country={Germany}
}      

\cortext[cor1]{\hspace*{0.1em}Corresponding author. \\
\hspace*{0.1em}\textit{E-Mail address:} \href{mailto:joana.reuss@tum.de}{joana.reuss@tum.de} (J{.} Reuss)}

\begin{abstract}
Real-world agricultural monitoring is often hampered by severe class imbalance and high label acquisition costs, resulting in significant data scarcity.
In \gls{gl:fsl}---a framework specifically designed for data-scarce settings---, training sets are often artificially balanced.
However, this creates a disconnect from the long-tailed distributions observed in nature, leading to a distribution shift that undermines the model's ability to generalize to real-world agricultural tasks.
We previously introduced \gls{gl:dirpa} to proactively mitigate the effects of such label distribution skews during model training.
In this work, we extend the original study's geographical scope.
Specifically, we evaluate this extended approach across multiple countries in the \gls{gl:eu}, moving beyond localized experiments to test the method's resilience across diverse agricultural environments.
Our results demonstrate the effectiveness of \gls{gl:dirpa} across different geographical regions.
We show that \gls{gl:dirpa} not only improves system robustness and stabilizes training under extreme long-tailed distributions, regardless of the target region, but also substantially improves individual class-specific performance by proactively simulating priors.
\end{abstract}


\begin{highlights}
\item Extensive validation of Dirichlet Prior Augmentation for prior shift in FSL
\item Cross-dataset stability shown across eight European countries
\item Strong correlation between class imbalance, prior shift, and Dirichlet gains
\end{highlights}


\end{frontmatter}


\glsreset{gl:fsl}
\glsreset{gl:dirpa}

\section{Introduction}\label{sec:intro}
With climate change on the rise, today's agricultural practices, as we know them, are severely at risk.
Therefore, not only recently, food security has been appointed as one of the \num{17} \glspl{gl:sdg} of the 2030 Agenda for Sustainable Development by the \gls{gl:un}.
As a result, agriculture-related tasks such as crop mapping, yield estimation, water stress monitoring, and tracking pest or disease outbreaks are becoming increasingly important.
The sheer amount of \gls{gl:rs} data provided by \gls{gl:eo} and meteorological satellites, as well as the latest advancements in \gls{gl:ml} and, particularly, \gls{gl:dl}, have significantly contributed to solving such tasks, for instance in \citet{Garnot20:TAE,Bermudez17:RNNsforCrops,Russwurm20:selfattention,Saini18:cropclassificationRFSVM}.
However, data-driven \gls{gl:dl} models require vast amounts of annotated data to achieve acceptable performances.
This frequently poses a significant challenge in agricultural applications, as labeled crop data is often rare and unevenly distributed globally, due to high acquisition costs and the impracticality of manual annotation.
Consequently, the need for \gls{gl:dl} approaches that rely less on substantial amounts of data has emerged as a major research direction.
A \gls{gl:sota} framework that aims to increase model performance on a target task with very limited amounts of data, is \gls{gl:fsl}.
\Gls{gl:fsl} is typically defined as an $n$-way $k$-shot problem, where $n$ defines the number of distinct classes and $k$ the number of samples per class.
Therefore, in common \gls{gl:fsl}, the target dataset consists of a balanced training (support) set and a corresponding, also balanced, test (query) set.
Conversely, real-world agricultural data is characterized by high class imbalance, with the majority of samples belonging to a limited number of classes (\eg \wheat) and a considerable amount of underrepresented crops (\eg \parsley).
Therefore, various studies have recommended the use of arbitrary and imbalanced test sets to better reflect real-world scenarios \citep{Reuss26:MTLSSL,Ochael23:classimbalanceFSL,Mohammadi24:fewshotcropmappingdirichlet}.
In doing so, we induce a prior distribution shift, where the training prior $p_\text{train}(y)$ differs from the final real-world test prior $p_\text{test}(y)$, potentially resulting in the model learning a strong, incorrect bias, which often leads to poor generalization \citep{Reuss26:MTLSSL,Reuss26:DirPA}.

In the past, such prior shifts have traditionally been addressed, for instance, via the adjustment of class probabilities at inference-time \citep{Lipton18:BlackBox,Kluger21:FPSA}, \cf \cref{sec:relatedwork}.
We previously demonstrated, however, that our novel \gls{gl:dirpa} method proactively addresses prior shifts during training, leading to improved model performance and robustness without requiring additional inference-time adjustments \citep{Reuss26:DirPA}.
However, while \gls{gl:dirpa} improved \emph{overall classification accuracy} $a_\text{OA}$ and \emph{Cohen's kappa} $\kappa$, initial experiments showed inferior performance on macro metrics.

Building upon our previous work \citep{Reuss26:DirPA}, this study represents a major expansion.
We now apply \gls{gl:dirpa} across an extensive geographical area to assess its operational stability and reliability under diverse environmental and agricultural conditions.

The main contributions are:
\begin{enumerate}
    \item \textbf{Geographical expansion}: We apply the \gls{gl:dirpa} method to a large-scale, real-world dataset covering eight diverse European countries, demonstrating cross-dataset stability.
    \item \textbf{Imbalance-dependent performance}: We reveal a strong correlation between the dataset's structural class imbalance and the magnitude of performance gains.
    \item \textbf{Hierarchical robustness}: We show that \gls{gl:dirpa} enhances classification reliability by boosting parent-level accuracy and F1-scores, ensuring accurate land-cover identification even when fine-grained distinctions are constrained by data scarcity.
\end{enumerate}


\section{Related work}\label{sec:relatedwork}
We study few-shot crop-type classification under realistic long-tailed label distributions, where balanced training sets induce an artificial prior shift between training and evaluation. 
In this section, we review the few-shot paradigm, imbalance-aware training frameworks, and prior-shift handling.

\subsection{Few-shot learning via knowledge transfer}
A common approach in \gls{gl:fsl} is to leverage prior knowledge gained from training on a distinct source task.
Various approaches have been explored to equip a model with relevant knowledge before being trained on a data-scarce target task, among the most prominent being \emph{transfer learning}, \emph{meta-learning}, and \gls{gl:ssl}.
While both meta-learning and \gls{gl:ssl} have been shown to achieve good performance, they come with high computational costs \citep{Reuss26:MTLSSL}.
Although \gls{gl:fsl} and the concept of meta-learning are frequently used as synonyms, \gls{gl:fsl} may also denote a task defined strictly by its low-data constraints \citep{Reuss26:DirPA}.
We adopt this broader definition of \gls{gl:fsl}, acknowledging recent findings on transfer learning---via regular pre-training and subsequent fine-tuning---achieving competitive results compared to complex meta-learning algorithms across a wide range of agricultural applications \citep{Vinija24:transferlearning,Alem22:transferlearningLCLU,Wang18:cropyieldprediction,Rouba23:heterogenoustransferlearning}.
Transfer learning leverages the transferability of latent features, assuming that the source-domain knowledge generalizes sufficiently to overcome the lack of samples from the target domain.

\subsubsection{Class imbalance}

\Citet{Ochael23:classimbalanceFSL} provide a detailed comparison and evaluation of various existing \glsxtrlong{gl:fsl} methods under class imbalance. 
They found that classical data-level rebalancing, in particular, random oversampling applied to the support set, yields consistent improvements across different model architectures and imbalance distributions. 
In contrast, loss-based reweighting strategies, including weighted loss and \gls{gl:fl}, provide limited or inconsistent gains and generally underperform random oversampling.

\subsubsection{FSL in remote sensing}
In recent years, researchers have widely applied \gls{gl:fsl} in agricultural applications to address data scarcity. 
First, \citet{Russwurm20:MTL} and \citet{Wang20:MTL} demonstrated the superiority of algorithms based on \gls{gl:maml} compared to regular transfer learning on land-cover classification.
\Citet{Tseng22:TIML} extended the original \gls{gl:maml} algorithm to explicitly support agricultural monitoring by incorporating additional metadata, such as spatial coordinates.
Furthermore, \citet{Keraani22:FSLTAE} proposed a temporal attention-based encoder for few-shot crop classification and employed transfer learning, training on an imbalanced dataset using \gls{gl:fl} and data augmentation.
Addressing the need for realistic \gls{gl:fsl} benchmarks, we provided a comprehensive cross-regional benchmark study using the few-shot crop-type dataset \EuroCropsML \citep{Reuss26:MTLSSL,Reuss25:EML}.
Their findings show that, while meta-learning achieves superior performances compared to regular transfer learning and self-supervised learning, it comes at the expense of substantially increased computational costs.
Moreover, they highlight that none of the evaluated methods were able to overcome the distributional discrepancy between the balanced training set and the imbalanced test set.

\subsection{Prior shift in FSL}
\begin{figure}[t]
    \centering
    \input{images/priorshift}
    \caption{Visualization of a label (prior) distribution shift between train dataset and a long-tailed test dataset. 
    }
    \label{fig:priorshift}
\end{figure}
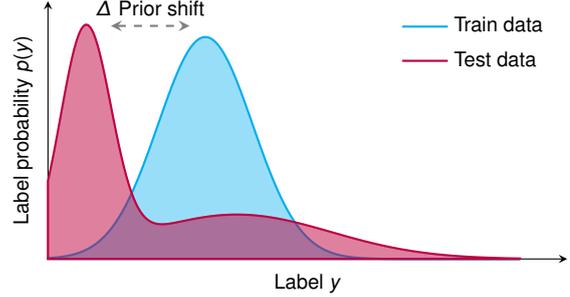

In \gls{gl:fsl}, training episodes are typically constructed with balanced class priors, whereas deployment data is often long-tailed. 
This induces a shift in the label (prior) distribution, as visualized in \cref{fig:priorshift}, which biases posterior predictions and can degrade performance. 
Prior-shift methods, therefore, either estimate the test distribution and recalibrate predictions at inference time or aim to reduce sensitivity to unknown priors during training.

Recent studies often address the problem of prior distribution shifts at inference time.
\Gls{gl:bbse} estimates the test distribution $p_\text{test}(y)$ to improve generalization for symptom-diagnosis detection.
\Citet{Sipka22:hitchiker} present a novel prior estimation approach based on confusion matrices, addressing infeasible solutions arising from inconsistent decision probability  and confusion matrix estimates.

\citet{Kluger21:FPSA} directly tackles the problem of label (prior) and feature (covariate) distribution shift in few-shot crop-type classification using crop statistics, assuming that the distribution of the test set is known.
Specifically, to address the prior distribution shift, they reweigh the posterior probabilities.

Our focus is on demonstrating that \gls{gl:dirpa}'s training-time regularization yields a robust pipeline that requires no post-hoc adjustment.
While existing inference-time approaches are complementary to our work, a direct comparison is therefore beyond the scope of this study.

\subsection{Dirichlet priors and distribution augmentation}
The \gls{gl:DirDist} $\operatorname{Dir}(\boldsymbol{\alpha})$ is often used to model the prior in Bayesian statistics \citep{Reuss26:DirPA}.
Among others, previous work addressed supervised clustering \citep{Daume05:DirichletClustering} and the utilization of Dirichlet priors within a Bayesian framework for regression \citep{Rademacher21:Dirichletpriors}. 
The \gls{gl:DirDist} offers several desirable properties as a Bayesian prior \citep[\cf][]{Rademacher21:Dirichletpriors}.
First, it has full support over the space of probability distributions, ensuring that the true data-generating distribution remains learnable regardless of the initial prior specification.
Second, it is the conjugate prior for multinomial data (\eg categorical counts), yielding closed-form posterior distributions that simplify mathematical derivations and improve computational efficiency.
Finally, it provides controllable informativeness through the localization parameter $\alpha_0$, which explicitly governs the bias-variance trade-off, and allows the prior to range from \emph{highly opinionated} to \emph{effectively non-informative}, depending on the degree of prior knowledge.

Most previous few-shot learning studies employing Dirichlet priors typically assume a balanced training (support) and test (query) set.
Consequently, they often use Dirichlet sampling primarily to simulate imbalanced test distributions.
Specifically, \citet{Veilleux21:dirichletevaluation} model the class marginal distribution of the query set as a random variable drawn from a \gls{gl:DirDist}.
Their analysis shows that several \gls{gl:sota} transductive few-shot methods implicitly encode a class-balance prior and suffer significant performance degradation under Dirichlet-sampled query sets. 
Similarly, \citet{Mohammadi24:fewshotcropmappingdirichlet} adopt Dirichlet-distributed query sampling to evaluate few-shot learning methods for crop mapping.
They demonstrate that commonly used balanced-query evaluations systematically overestimate model performance.
Using realistic Dirichlet-sampled query sets, they report consistent performance drops across methods.
These approaches essentially serve as a few-shot evaluation method, as they solely focus on simulating artificially imbalanced test sets.
Contrarily, we introduced Dirichlet priors during training, based on the assumption of having a balanced training set and an imbalanced test set \citep{Reuss26:DirPA}.
They employ \gls{gl:dirpa}, which proactively tackles this distribution shift directly during training by sampling pseudo-priors from the \emph{symmetric} \gls{gl:DirDist}.
By learning across a wide variety of priors, the model learns a \emph{prior-agnostic} representation, ensuring stable generalization to real-world imbalanced distributions without requiring prior knowledge of the target domain at test time.

\section{Dataset}\label{sec:dataset}
We conduct all training and evaluation on \EuroCropsMLNew, an extension and refinement of the \EuroCropsML dataset \citep{Reuss25:EML}.
\EuroCropsML provides parcel-level, multi-temporal Sentinel-2 L1C (top-of-atmosphere) observations paired with multi-class crop annotations originating from the \EuroCrops reference dataset \citep{Schneider2023:EuroCrops}. 
All crop labels follow the \gls{gl:hcat}, where, for instance, level \num{3} represents general land cover types and level \num{6} the most-granular crop type, distinguished by seasonality, if applicable.

We extend the geographic coverage of the original \EuroCropsML with 2021 parcel-level crop data from Austria, Belgium (Flanders and Wallonia), Estonia, Germany (\gls{gl:ls} and \gls{gl:nrw}), Lithuania, Latvia, Portugal, Slovakia, Slovenia, and Spain (Navarra).
While we adhere to the original processing pipeline for median-based parcel aggregation, all observations have been reprocessed using the Sentinel-2 L1C Collection 1 \citep{Enache23:S2C1,ESA21:S2C1}.
This ensures a harmonized data standard across the full geographic coverage.
Furthermore, we refine the cloud-removal step by also accounting for thin cirrus and snow or ice, following the classification approach of the L2A algorithm \citep{ESA_Sentinel2_ATBD}.
An overview of the dataset details is given in \cref{tab:EuroCropsML2stats}.

\begin{table} 
\caption{%
Dataset details of \EuroCropsMLNew.
The \emph{\# Crop classes} column refers to the number of unique crop types in \gls{gl:hcat} level \num{6}.
The \emph{\# Parcels} column refers to unique parcel geometries after removing potential duplicates from the \EuroCrops reference data and filtering out clouds, cirrus, snow, and ice. \\ 
} 
\label{tab:EuroCropsML2stats} 
\centering
\begin{tabular}[]{@{}Xrr@{}} 
    \toprule 
    \multicolumn{1}{@{}X}{Country} & \multicolumn{1}{X}{\# Crop classes} & \multicolumn{1}{X@{}}{\# Parcels} \\ 
    \cmidrule(r){1-1} \cmidrule(lr){2-2} \cmidrule(l){3-3} 
    \href{https://github.com/maja601/EuroCrops/wiki/Austria}{Austria} & \num{98} & \num{2589192} \\ 
    \href{https://github.com/maja601/EuroCrops/wiki/Belgium}{Belgium} & \num{136} & \num{560671} \\
    \smaller\hspace{1em} Flanders & \smaller \num{120} & \smaller \num{223269} \\ 
    \smaller\hspace{1em} Wallonia & \smaller \num{91} & \smaller \num{337402} \\ 
    \href{https://github.com/maja601/EuroCrops/wiki/Estonia}{Estonia} & \num{127} & \num{175905} \\ 
    \href{https://github.com/maja601/EuroCrops/wiki/Germany}{Germany} & \num{198} & \num{1632620} \\ 
    \smaller\hspace{1em} Lower Saxony & \smaller \num{150} & \smaller \num{901364} \\ 
    \smaller\hspace{1em} North Rhine-Westphalia & \smaller \num{175} & \smaller \num{731256} \\ 
    \href{https://github.com/maja601/EuroCrops/wiki/Latvia}{Latvia} & \num{108} & \num{431143} \\ 
    \href{https://github.com/maja601/EuroCrops/wiki/Lithuania}{Lithuania} & \num{22} & \num{1101839} \\ 
    \href{https://github.com/maja601/EuroCrops/wiki/Portugal}{Portugal} & \num{84} & \num{99612} \\ 
    \href{https://github.com/maja601/EuroCrops/wiki/Slovakia}{Slovakia} & \num{121} & \num{251250} \\ 
    \href{https://github.com/maja601/EuroCrops/wiki/Slovenia}{Slovenia} & \num{112} & \num{818902} \\ 
    \href{https://github.com/maja601/EuroCrops/wiki/Spain}{Spain} & 10 & \num{962111} \\ 
    \smaller\hspace{1em} Navarra & \smaller 10 & \smaller \num{962111} \\ 
    \bottomrule 
\end{tabular} 
\end{table}

\subsection{Dataset split}
\begin{figure}[t]
    \centering
    \includegraphics[width=0.95\linewidth]{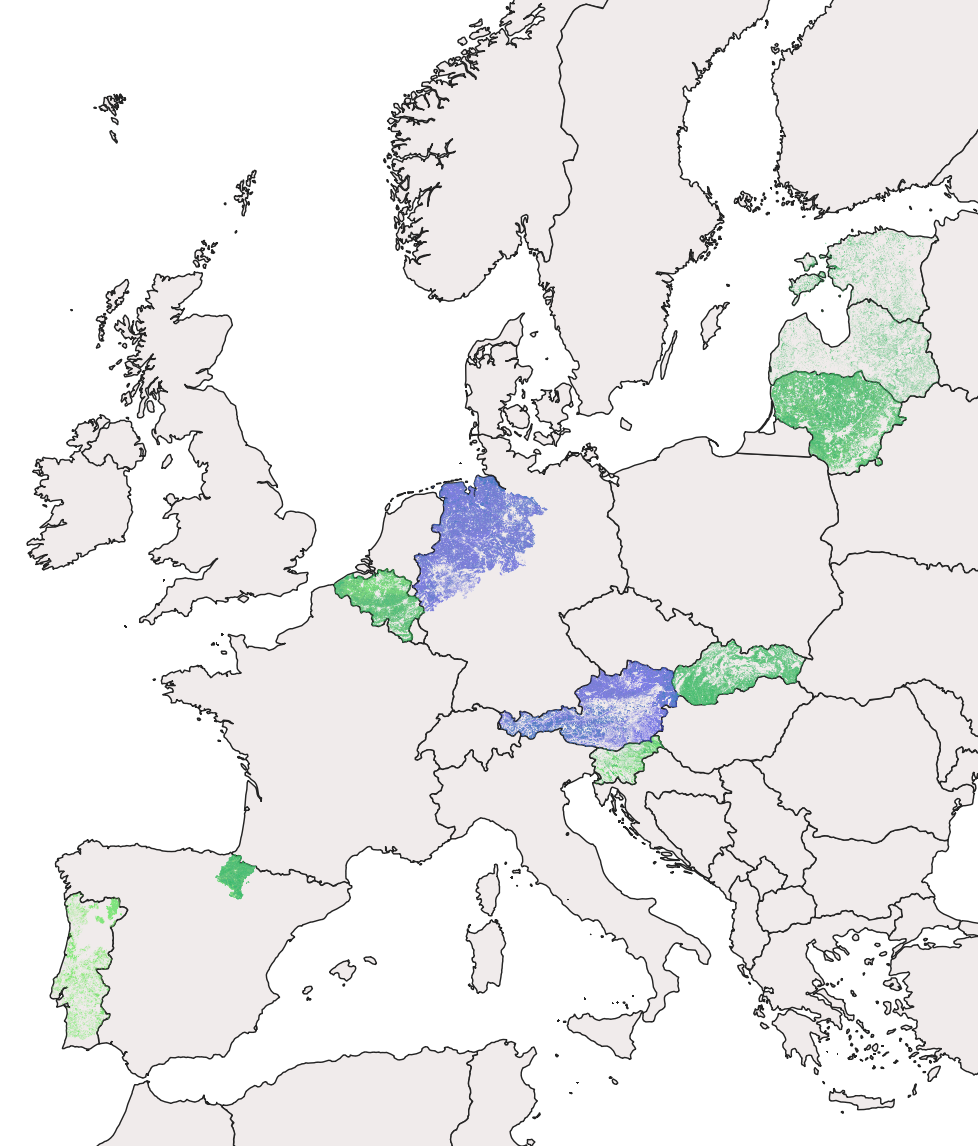}
    \caption{Pre-training and fine-tuning country split.
    Countries colored in {\color{mediummidnightblue}blue} are used for pre-training, whereas those colored in {\color{lightseagreen}green} are used for fine-tuning. 
    The darker the coloring, the denser the labeled agricultural fields.}
    \label{fig:dataset_split}
\end{figure}

We define a universal pre-training dataset comprising Austria and Germany.
All other remaining countries are utilized as individual fine-tuning sets.
\Cref{fig:dataset_split} illustrates the pre-training and fine-tuning countries in detail.
Pre-training on a joint Austrian-German dataset offers several advantages and specific characteristics, which we outline below.

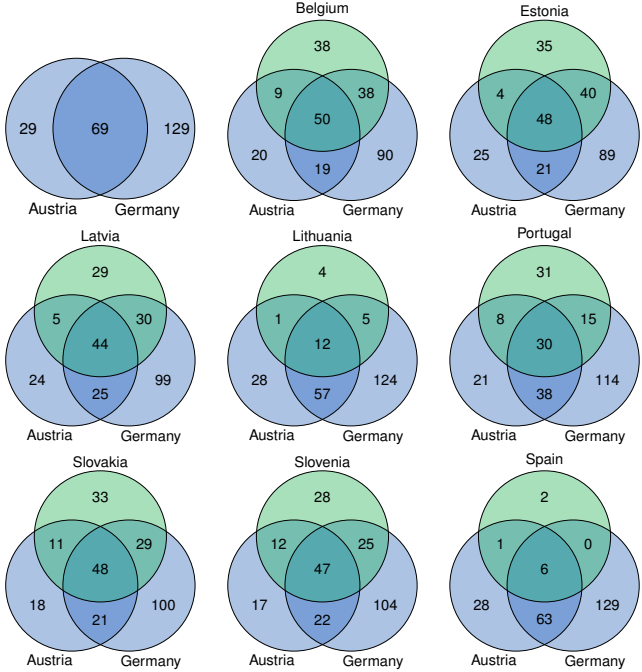
\begin{figure}
    \centering
    \begin{minipage}[t]{0.32\linewidth}
        \centering
        \resizebox{.9\linewidth}{!}{\input{images/classoverlap/AT_DE}}
    \end{minipage}
    \begin{minipage}[t]{0.32\linewidth}
        \centering
        \resizebox{.9\linewidth}{!}{\input{images/classoverlap/AT_DE_BE}}
    \end{minipage}
    \begin{minipage}[t]{0.32\linewidth}
        \centering
        \resizebox{.9\linewidth}{!}{\input{images/classoverlap/AT_DE_EE}}
    \end{minipage}
    \begin{minipage}[t]{0.32\linewidth}
        \centering
        \resizebox{.9\linewidth}{!}{\input{images/classoverlap/AT_DE_LV}}
    \end{minipage}
    \begin{minipage}[t]{0.32\linewidth}
        \centering
        \resizebox{.9\linewidth}{!}{\input{images/classoverlap/AT_DE_LT}}
    \end{minipage}
    \begin{minipage}[t]{0.32\linewidth}
        \centering
        \resizebox{.9\linewidth}{!}{\input{images/classoverlap/AT_DE_PT}}
    \end{minipage}
    \begin{minipage}[t]{0.32\linewidth}
        \centering
        \resizebox{.9\linewidth}{!}{\input{images/classoverlap/AT_DE_SK}}
    \end{minipage}
    \begin{minipage}[t]{0.32\linewidth}
        \centering
        \resizebox{.9\linewidth}{!}{\input{images/classoverlap/AT_DE_SI}}
    \end{minipage}
    \begin{minipage}[t]{0.32\linewidth}
        \centering
        \resizebox{.9\linewidth}{!}{\input{images/classoverlap/AT_DE_ES}}
    \end{minipage}
    \caption{The number of annotated crop classes that are shared and distinct between the pre-training countries---Austria and Germany---and the respective fine-tuning country.}
    \label{fig:overlap_diagrams}
\end{figure}

\paragraph{Class distribution}
\begin{figure}[t]
    \input{tables/at_de_classes_logbin}
    \caption{\EuroCropsMLNew binned class distribution for Germany (\gls{gl:ls} and \gls{gl:nrw}) and Austria, shown on a logarithmic scale, with the \meadow class removed.}
    \label{fig:at_de_classes_logbin}
\end{figure}
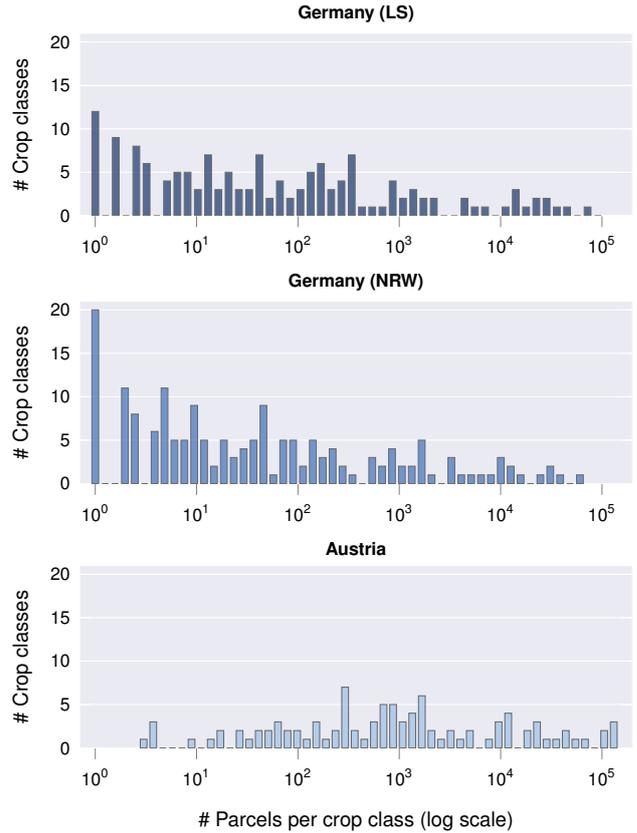
Both countries exhibit diverse class distributions and highly diverse reporting structures, resulting in very fine-grained crop classes.
This enables the training of a rich feature extractor and is necessary for a successful transfer to fine-grained target tasks.
Specifically, Austria contains \num{2589192} samples, while \gls{gl:ls} and \gls{gl:nrw} comprise \num{901364} and \num{731256} samples, respectively.
Despite the larger total sample count in Austria, Germany exhibits a higher number of unique crop classes with very few samples, as shown in \cref{fig:at_de_classes_logbin}.
Across both Austria and Germany, the \meadow class dominates by a wide margin, accounting for 42\% of samples in Germany (\gls{gl:ls}: 41\%, \gls{gl:nrw}: 43\%) and 51\% in Austria.
When excluding the dominant \meadow class, crop distributions in Germany are similar across \gls{gl:ls} and \gls{gl:nrw}.
In \gls{gl:ls}, \maize is the most frequent crop class at 22\%, followed by \winterwheat at 14\%.
In \gls{gl:nrw}, \maize contributes 18\%, while \winterwheat accounts for 16\%.
In contrast, Austria exhibits a different composition, with the most prevalent classes being \forest (13\%) and \grapes (13\%).
A detailed visualization of overlapping classes between the pre-training countries and each fine-tuning country is shown in \cref{fig:overlap_diagrams}.
With regard to Spain, the class \class{not known or other class} is excluded, given its inclusion of all elements not categorized as agricultural fields.

\paragraph{Biogeographical regions}
\begin{figure}[t]
    \centering
    \includegraphics[width=0.95\linewidth]{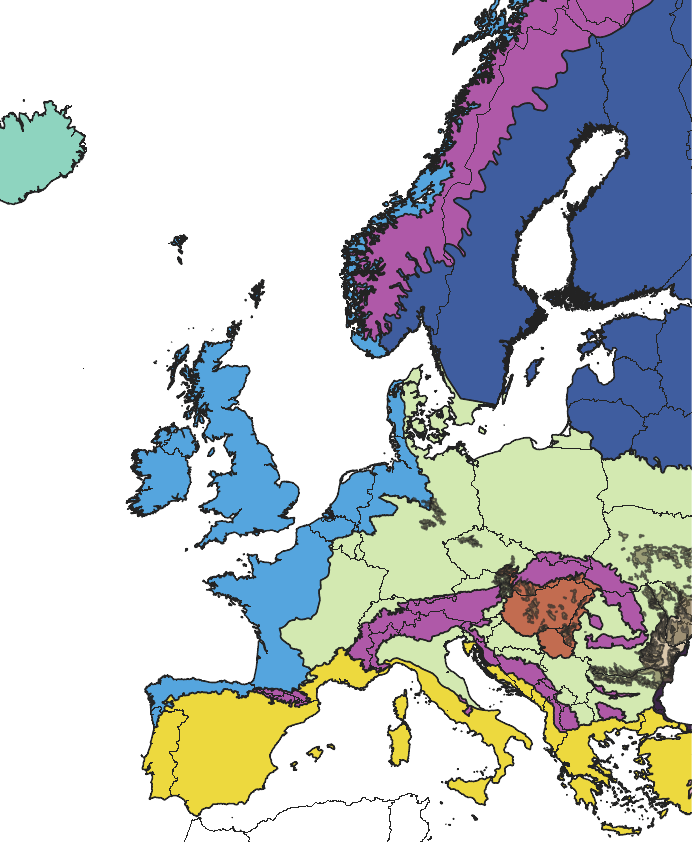}
    \vspace{10pt}
    \resizebox{0.95\linewidth}{!}{\input{images/biogeographical_regions_legend}}
    \caption{Biogeographical regions present in \EuroCropsMLNew, as defined by the \citep{EEA16:biogeographical}. 
    }
    \label{fig:biogeographical_regions}
\end{figure}
Both countries are located in central Europe.
Hence, they are mostly characterized by the \emph{Continental} biogeographical zone, with marginal overlaps from the \emph{Alpine} or \emph{Atlantic} regions \citep{EEA16:biogeographical}, \cf \cref{fig:biogeographical_regions}.
This establishes a \emph{central European baseline} that avoids severe climatic conditions and permits the investigation of varying levels of domain shift.

\paragraph{Agricultural practices}
Austria is characterized by a higher number of small-scale agricultural fields, with a median field area of \SI{0.46}{\hectare} compared to a mean of \SI{1.22}{\hectare}.
In contrast, \gls{gl:nrw} and \gls{gl:ls} exhibit larger typical field sizes, with medians of \SI{1.18}{\hectare} and \SI{1.86}{\hectare} and means of \SI{2.04}{\hectare} and \SI{2.87}{\hectare}, respectively.
As a result, the per-field median-pixel estimates of \EuroCropsMLNew in Austria are derived from a much smaller number of pixels, thereby increasing their sensitivity to outliers.
Conversely, the larger parcel sizes observed in Germany allow for aggregation over a much greater number of pixels, thereby yielding more robust spectral reflectance values.

\subsection{Fine-tuning split}
\label{sec:dataset_split}
Most fine-tuning countries exhibit a long-tailed data distribution.
Therefore, simply applying a random train, validation, and test split would result in zero-shot classification and non-representative validation sets.
Our objective is to establish a robust and realistic training-validation environment while reserving a large, representative set for the final evaluation.
Consequently, the overall split was prioritized as follows:
\begin{description} 
    \item[Test set] 20\% of the total dataset for the final, unbiased performance evaluation.
    \item[Validation set] A small, efficient set of \num{5000} samples for hyperparameter tuning and early stopping.
    \item[Training set] The remaining samples for training. 
\end{description}
We employ a class-aware splitting strategy to ensure that all fine-tuning classes are represented in the training set, serving as a prerequisite for assuming known crop types in real-world scenarios.
Let $N_c$ be the total number of available samples for class $c$.
We employ the following allocation protocol:
\begin{enumerate}
    \item \textbf{$N_c=1$}: The sample is allocated to the training set.
    \item \textbf{$N_c=2$}: One sample is allocated to the training and one to either validation or test.
    \item \textbf{$N_c=3$}: One sample is assigned to each subset.
    \item \textbf{$N_c>3$}: The samples are randomly distributed between the test and validation sets to fulfill the remaining capacity required to meet the 20\% test target and \num{5000} validation target\footnote{%
    This is constrained by the class distribution and, consequently, on occasion, results in slightly lower or higher values.
    These values are generated to meet the criterion of maximizing the class overlap across all three subsets.
    }, while maintaining the class distribution of the original pool.
    All samples not selected for the validation or test sets are added to the training set.
\end{enumerate}
This strategy ensures that the final validation set, although small, contains a proportional representation of all classes with $N_c\geq3$, thereby preventing evaluation biases. 
\Cref{tab:finetuningsplit_numbers} gives an overview of the final number of distinct crop classes in each validation and test set.
Furthermore, \cref{tab:gini} shows the \emph{Gini coefficient} $C_\text{G}$ \citep{Ceriani11:Gini} of each fine-tuning test set, representing the level of class imbalance. 
The Gini coefficient provides a single bounded and interpretable scalar that quantifies the class imbalance, enabling a fair comparison across datasets with different numbers of classes.

\begin{table}
    \caption{%
        Class distribution of the \EuroCropsMLNew fine-tuning sets, showing the number of distinct crop classes.
        The total number of classes in the training set equals the number of available crop types and can be extracted from \cref{tab:EuroCropsML2stats}.
    }
    \label{tab:finetuningsplit_numbers}
    \centering
    \begin{tabular}[t]{@{}Xrrrr@{}}
      \toprule
      \multicolumn{1}{@{}X}{Country} & \multicolumn{4}{X@{}}{\# Crop classes} \\
      \cmidrule(l){2-5}
      & \multicolumn{1}{X}{Train $\setminus$ (validation $\cup$ test)} & \multicolumn{1}{X}{Validation} & \multicolumn{1}{X}{Test} & \multicolumn{1}{X@{}}{Validation $\cap$ test} \\
      \cmidrule(r){1-1} \cmidrule(lr){2-2} \cmidrule(lr){3-3} \cmidrule(lr){4-4}  \cmidrule(l){5-5} 
     Belgium & \num{2} & \num{133} & \num{133} & \num{132} \\  
     Estonia & \num{15} & \num{107} & \num{108} & \num{103} \\
     Latvia & \num{2} & \num{106} & \num{105} & \num{105} \\  
     Lithuania & \num{0} & \num{22} & \num{22} & \num{22} \\  
     Portugal & \num{10} & \num{72}  & \num{71} & \num{69} \\  
     Slovakia & \num{10} & \num{107}  & \num{107} & \num{103} \\  
     Slovenia & \num{5} & \num{105} & \num{105} & \num{103} \\
     Spain & \num{0} & \num{9} & \num{9} & \num{9} \\  
\bottomrule
    \end{tabular}
\end{table}

We evaluate various few-shot settings with 
\numlist[list-final-separator = {, or }]{1;5;10;20;100} shots
\citep[\cf][]{Reuss25:EML,Reuss26:DirPA}, where \enquote{$k$-shot} denotes a maximum of $k$ samples per class, using all available samples when a class contains fewer than $k$.
This allows us to determine the performance impact of the prior adjustment across different degrees of prior shifts.
For each country, we compared the class distributions between the few-shot training set and the final test set to gauge the extent of the prior shift.
In particular, we calculated the \emph{Bhattacharyya distance} $D_\text{B}$ \citep{Bhattacharyya46} to quantify the statistical overlap between crop classes, as shown in \cref{tab:dBEuroCropsML}.

\begin{table}[t]
    \caption{%
        Gini coefficient for the test sets of the \EuroCropsMLNew fine-tuning countries, a measure of dataset skewness.
        A value of \num{0} indicates a perfectly balanced class distribution, while a value approaching \num{1} means almost all samples belong to a single majority class.      
    }
    \label{tab:gini}
    \centering
    \begin{tabular}[t]{@{}Xr@{}}
      \toprule
       \multicolumn{1}{@{}X}{Country} & \multicolumn{1}{X@{}}{Gini 
       coefficient $C_\text{G}$} \\
      \cmidrule(r){1-1} \cmidrule(l){2-2} 
      Belgium &  \num{0.939}  \\
      Estonia &  \num{0.919}  \\
      Latvia &  \num{0.910}  \\
      Lithuania &  \num{0.761}  \\
      Portugal &  \num{0.881}  \\
      Slovakia &  \num{0.904}  \\
      Slovenia &  \num{0.928}  \\
      Spain &  \num{0.634}  \\
\bottomrule
    \end{tabular}
\end{table}

\begin{table}[t]
    \caption{%
        Bhattacharyya distance for fine-tuning countries in \EuroCropsMLNew, calculated between the fixed test set and the few-shot training sets.
        The distance is \num{0} if the two distributions are identical and increases as they become more distinct.        
    }
    \label{tab:dBEuroCropsML}
    \centering
    \begin{tabular}[t]{@{}Xrrrrr@{}}
      \toprule
       \multicolumn{1}{@{}X}{Country} & \multicolumn{5}{X@{}}{$k$-shot training set} \\
       \cmidrule(l){2-6}
       & \multicolumn{1}{X}{\num{1}} & \multicolumn{1}{X}{\num{5}} & \multicolumn{1}{X}{\num{10}} & \multicolumn{1}{X}{\num{20}} & \multicolumn{1}{X@{}}{\num{100}} \\
      \cmidrule(r){1-1} \cmidrule(lr){2-2} \cmidrule(lr){3-3} \cmidrule(lr){4-4} \cmidrule(lr){5-5} \cmidrule(l){6-6} 
      Belgium &  \num{0.880} & \num{0.850} & \num{0.827} & \num{0.786} & \num{0.637} \\
      Estonia &  \num{0.857} & \num{0.740} & \num{0.691} & \num{0.629} & \num{0.450} \\
      Latvia &  \num{0.736} & \num{0.716} & \num{0.688} & \num{0.655} & \num{0.539}  \\
      Lithuania &  \num{0.365} & \num{0.365} & \num{0.365} & \num{0.365} & \num{0.365} \\
      Portugal &  \num{0.705} & \num{0.588} & \num{0.552} & \num{0.495} & \num{0.370} \\
      Slovakia & \num{0.773} & \num{0.680} & \num{0.645} & \num{0.605} & \num{0.462}  \\
      Slovenia & \num{0.866} & \num{0.804} & \num{0.782} & \num{0.750} & \num{0.621}  \\
      Spain &  \num{0.213} & \num{0.213} & \num{0.213} & \num{0.213} & \num{0.213} \\
\bottomrule
    \end{tabular}
\end{table}

\section{Methodology}\label{sec:methods}
We adopt the \gls{gl:dirpa} method \citep{Reuss26:DirPA}, where class-prior uncertainty is modeled via a symmetric \gls{gl:DirDist} and injected into the classifier logits during the training process. 

\subsection{DirPA: Dirichlet prior augmentation}
\label{sec:dirpa}

\begin{figure*}[t]
  \centering
  \subcaptionbox{$\boldsymbol{\alpha}=(0.5, 0.5, 0.5)$ \label{fig:dir_a05}}[.21\linewidth]{\includegraphics[width=\linewidth, keepaspectratio]{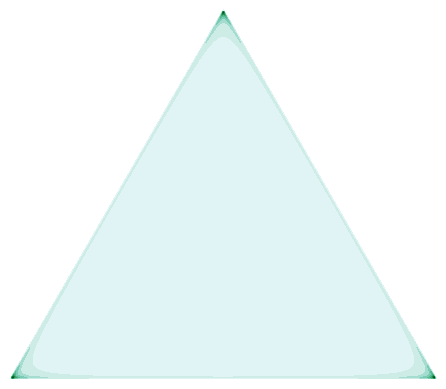}}
  \hspace{1em}
  \subcaptionbox{$\boldsymbol{\alpha}=(30, 30, 30)$ \label{fig:dir_a30}}[.21\linewidth]{\includegraphics[width=\linewidth, keepaspectratio]{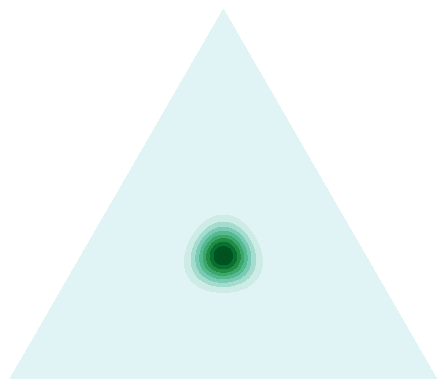}}
  \hspace{1em}
    \subcaptionbox{$\boldsymbol{\alpha}=(5, 5, 5)$ \label{fig:dir_a5}}[.21\linewidth]{\includegraphics[width=\linewidth, keepaspectratio]{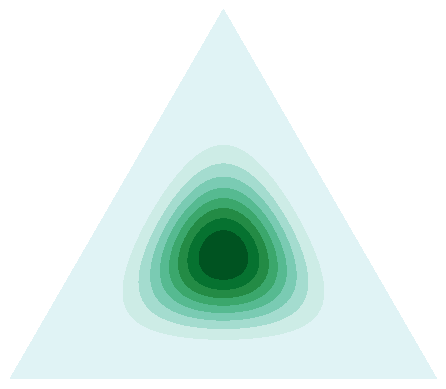}}
  \hspace{1em}
  \subcaptionbox*{\label{fig:dir_cbar}}[.05\linewidth]{%
    \resizebox{!}{3.5cm}{\input{images/dirichlet/dirichlet_densitycolorbar}}
  }\hspace*{0.04\linewidth}
  \caption{Dirichlet density for $K=3$ classes, defined over the $(K-1)=2$-simplex.
  The plots show different concentration parameters $\boldsymbol{\alpha} = \alpha \cdot \boldsymbol{1}$, where $\boldsymbol{1} = (1, 1, 1)^\top$ \citep{Reuss26:DirPA}.}
  \label{fig:dirichlet_density}
\end{figure*}
\begin{figure}[t]
    \centering
    \input{images/dirichlet/dirichlet_bars}
    \caption{Visualization of four possible draws of the pseudo-prior $\tilde{\boldsymbol{\pi}}^{(s)} \sim \operatorname{Dir}(\alpha \cdot \boldsymbol{1})$ for $K=5$ classes at consecutive training iterations $s$.
    The orange sampling ($\alpha = 0.5$) produces sparse draws, with a different class dominating at each iteration.
    With $\alpha = 5$ (teal), the draws remain closer to the uniform distribution, indicated by the dashed line.
    Resampling the pseudo-prior at each iteration exposes the model to a diverse set of imbalance patterns without assuming any fixed test distribution.
    }
    \label{fig:dirichlet_sampling}
\end{figure}
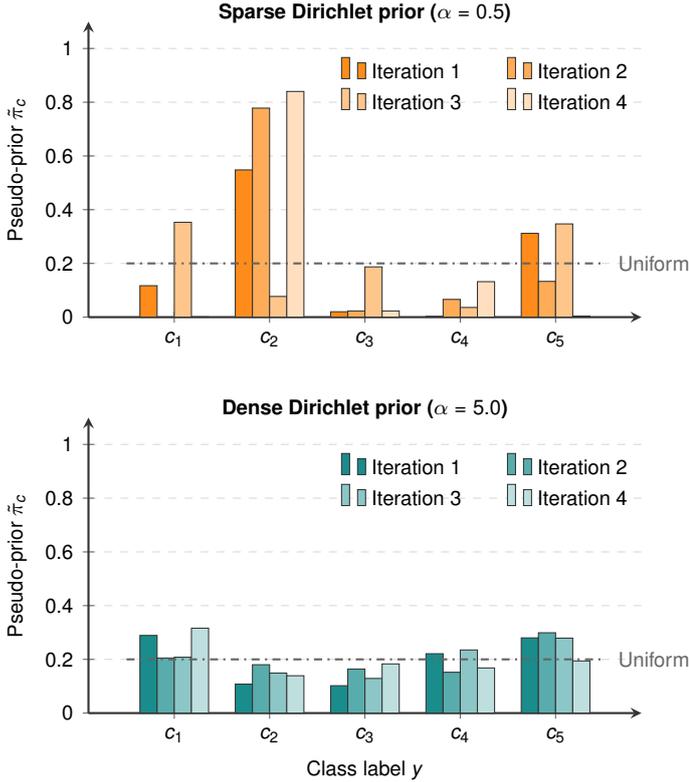

Let $f_{\boldsymbol{\theta}}$ be a model parameterized by $\boldsymbol{\theta}$ and let $\boldsymbol{z}_i = f_{\boldsymbol{\theta}}(\boldsymbol{x}_i) \in \mathbb{R}^K$ denote logits for sample $\boldsymbol{x}_i$ over $K$ classes. 
At each iteration $s$, a pseudo-prior 
\begin{align*}
  \tilde{\boldsymbol{\pi}}^{(s)} &\sim \operatorname{Dir}(\alpha \cdot \boldsymbol{1}), \quad \tilde{\boldsymbol{\pi}}^{(s)} \in \Delta^{K-1},
\end{align*}
is sampled from the symmetric \gls{gl:DirDist} defined over the $(K-1)$-dimensional simplex $\Delta^{K-1} = \{\boldsymbol{\pi} \in \mathbb{R}^K: \sum_{c=1}^K \pi_c = 1, \pi_c \ge 0\}$.
The parameter $\alpha \in \mathbb{R}_+$ controls the degree of imbalance.
A small value ($\alpha < 1$) samples highly skewed (imbalanced) distributions, while $\alpha \geq 1$ samples distributions closer to uniform, as visualized in \cref{fig:dirichlet_density}.
The sampled prior is incorporated through a logit adjustment, 
\begin{align*}
\boldsymbol{z}_i^\prime \leftarrow \boldsymbol{z}_i + \tau \log (\tilde{\boldsymbol{\pi}}^{(s)}),
\end{align*}
with $\tau \in \mathbb{R}_+$ being a scaling factor. 
The modified logits $\boldsymbol{z}_i^\prime$ are subsequently passed through the \emph{Softmax} function to obtain class probabilities
\begin{align*}
  \hat{p}_{i,c} = \sigma\left(z_{i,c}^\prime\right) = \frac{\exp(z_{i,c}^\prime)}{\sum_{k=1}^{K} \exp(z_{i,k}^\prime)}, \quad c = 1, \dots, K.
\end{align*}
%
%
\Cref{fig:dirichlet_sampling} illustrates the sampling process for $K=5$ classes and different values of $\alpha$.



\begin{algorithm}
\caption{Dirichlet Prior Augmentation}\label{alg:sym_dirpa}
\begin{algorithmic}[1]
\Require $\alpha, \tau \in \mathbb{R}_+$
\Require $\sigma(\cdot)$ (Softmax activation function)
\Require $f_{\boldsymbol{\theta}}$ (model parameterized by $\boldsymbol{\theta}$)
\For{each training step $s = 1$ to $S$}
\State sample mini-batch of data points $\mathcal{D}^{(s)} = \{(\boldsymbol{x}_i, y_i)\}_{i=1}^{b}$
\State sample pseudo-prior $\tilde{\boldsymbol{\pi}}^{(s)} \sim \operatorname{Dir}(\alpha \cdot \boldsymbol{1})$
\For{each data point $(\boldsymbol{x}_i, y_i)$ in $\mathcal{D}^{(s)}$}
    \State compute base logits: $\boldsymbol{z}_i \leftarrow f_{\boldsymbol{\theta}}(\boldsymbol{x}_i)$
    \State adjust logits: $\boldsymbol{z}_i' \leftarrow \boldsymbol{z}_i + \tau \log (\tilde{\boldsymbol{\pi}}^{(s)})$
    \State compute predictive distribution: $\hat{\boldsymbol{p}}_i \leftarrow \sigma(\boldsymbol{z}_i')$
\EndFor
\State compute mini-batch loss $\mathcal{L}_{\text{batch}} \leftarrow \frac{1}{b}\sum_{i=1}^{b} \mathcal{L}(\hat{\boldsymbol{p}}_i, y_i)$
\State update $\boldsymbol{\theta}$ via backpropagation
\EndFor
\end{algorithmic}
\end{algorithm}

\subsection{Transformer model}
\label{sec:transformer}
In alignment with \citet{Reuss26:DirPA}, all experiments from \cref{sec:experiments} are conducted using a state-of-the-art Transformer encoder architecture \citep{vaswani_vanillatransformer} stacked with a linear classification layer on top.
To encode the temporal information, we apply sinusoidal positional encodings with a maximum sequence length of $T_\text{max}=366$ days, representing a full leap year. 
While the encoder serves as the model's feature extractor, the linear layer maps these features to the final class log-probabilities.
We refer to the encoder 
\begin{align*}
  f^\text{backbone}_{\boldsymbol{\theta}_\text{backbone}} &\colon \mathbb{R}^{T_\text{max} \times d} \to \mathbb{R}^{n_e}
\end{align*}
as the \emph{backbone} and to the classification layer 
\begin{align*}
  f^\text{head}_{\boldsymbol{\theta}_\text{head}} &\colon \mathbb{R}^{n_e} \to \mathbb{R}^{K}
\end{align*}
as the \emph{head}.
$d \in \{1, \ldots, 13\}$ denotes the number of bands, $n_e \in \mathbb{N}$ the Transformer's embedding dimension, and $\boldsymbol{\theta}_\text{backbone}$ the backbone's and $\boldsymbol{\theta}_\text{head}$ the head's trainable parameter.

Subsequently, the full model is decomposed as 
\begin{align*}
  f_{\boldsymbol{\theta}} &= f^\text{head}_{\boldsymbol{\theta}_\text{head}} \circ f^\text{backbone}_{\boldsymbol{\theta}_\text{backbone}},
\end{align*}
where $\boldsymbol{\theta}=\left[\boldsymbol{\theta}_\text{backbone},\boldsymbol{\theta}_\text{head}\right]$ represents all trainable model parameters.

\section{Experiments}\label{sec:experiments}
Across all experiments, we employ a standard Transformer encoder consisting of four layers.
Each layer features eight attention heads and a feed-forward network with a hidden layer of dimension \num{1024}.
Initially, input tokens are projected into a latent space of dimension \num{256} using internal embeddings.
Regarding the spectral resolution, we retain \num{12} of the \num{13} Sentinel-2 bands, excluding \Bten (cirrus), which is utilized solely for cloud detection during pre-processing, \cf \cref{sec:dataset}. 
To obtain reflectance values, the raw data is normalized by a factor of $\num{e4}$.
The resulting values are clipped to the range $[0.0, 1.2]$ and subsequently adjusted by an offset of $\num{e{-4}}$.
\emph{No-data} values are either masked out or set to \num{0.0}.
For both pre-training and fine-tuning, all hyperparameters are optimized on the respective validation set.
For the tuning trial runs, we employ a multi-variate \gls{gl:tpe} with percentile pruning within the \texttt{Optuna} framework \citep{Akiba19:Optuna}.

All code used for training and evaluation will be made available upon publication.

\subsection{Additional baselines and experiments}
In addition to the standard \gls{gl:ce} and \acrfull{gl:fl} experiments, we include \gls{gl:ros} as a baseline for handling imbalanced support sets \citep{Ochael23:classimbalanceFSL}.
For each $k$-shot task, if a class contains fewer than $k$ samples, we randomly duplicate existing samples until exactly $k$ are available.
Furthermore, to demonstrate that \gls{gl:dirpa}'s benefits generalize across different architectures, we conduct additional experiments using the \gls{gl:presto}, an established model explicitly designed for pixel-wise time series and downstream few-shot classification.
Although also Transformer-based, \gls{gl:presto} differs in its use of self-supervised pre-training and per-channel tokenization with learned channel embeddings.
In alignment with \citet{Tseng23:PRESTO}, we remove bands \Bone, \Bnine, and \Bten.
We use the pre-trained weights to initialize the model backbone and evaluate \gls{gl:presto} with and without \gls{gl:dirpa} on a subset of countries.

\subsection{Pre-training}
First, we pre-train the model using downsampled parcel-level data from Austria and Germany.
We downsample all \num{226} distinct crop classes to at most \num{10000} samples per class.
If fewer samples are available for a given class, all available samples are retained for that class.
We use the standard \gls{gl:ce} with label smoothing and one cosine annealing cycle.
We train the model for a total of \num{100} epochs and validate after each epoch on the full pre-training validation set.
If the validation loss does not decrease for more than \num{15} epochs, we stop training.
For details on hyperparameter tuning, please refer to \ref{sec:hyperparametertuning_pretrain}.

\subsection{Fine-tuning}
Pre-training is followed by fine-tuning on each of the remaining \EuroCropsMLNew countries separately. For each fine-tuning case, we reinitialize the linear head $f^\text{head}$ with the number of fine-tuning classes, \cf \cref{tab:EuroCropsML2stats}.
For all experiments, we train for up to \num{200} epochs with a batch size of \num{16}.
Training is terminated if the validation loss fails to improve for \num{20} consecutive epochs.
We use \emph{Cohen's kappa} $\kappa$ evaluated on the validation set as the primary criterion for checkpoint selection, prioritizing models with robust inter-class reliability. 
This model state is then evaluated on the held-out test set.
As a chance-corrected metric, $\kappa$ ensures that the selected hyperparameters (\cf \ref{sec:hyperparametertuning_finetune}) and checkpoints reflect genuine discriminative performance across all classes, particularly in the presence of significant class imbalance.
This approach mitigates the accuracy paradox, in which high overall performance can be misleadingly achieved by frequent exploitation of the majority class. 
Furthermore, we favor Cohen's kappa over \emph{macro-averaged F1-scores}, as the latter tends to exhibit high variance across episodes in very low-shot regimes (1-shot, 5-shot)---a single misclassified sample in a rare class can produce large swings in the score, making it an unreliable signal for model selection.
Apparent improvements in F1-score in such settings are often attributable to chance rather than genuine model improvement.
Cohen's kappa, computed over the full confusion matrix, is statistically more stable and therefore a more robust criterion for optimization.
All fine-tuning experiments are conducted a total of five times with different random seeds $r \in \{0, 1, 42, 123, 1234\}$.
The chosen random seed influences, for instance, the initialization of random parameters and mini-batch sampling.
We conduct all fine-tuning experiments with both \glsxtrlong{gl:ce} and \glsxtrlong{gl:fl} (without a class-imbalance factor) in order to validate \gls{gl:dirpa} across varying optimization landscapes.
By utilizing \gls{gl:fl}, we can analyze the synergy between Dirichlet priors and difficulty-awareness.
Both loss functions are trained with and without Dirichlet priors.
We append the postfix \enquote{\gls{gl:dirpa}} whenever the \gls{gl:dirpa} method is used.

\section{Results}\label{sec:results}
\begin{figure*}[t]
  \centering 
  \subcaptionbox{Performance gain for \gls{gl:ce} \gls{gl:dirpa} compared to its \gls{gl:ce} baseline. \label{fig:ce_results_lv3}}[.48\linewidth]
  {%
    \centering
    \input{images/bubble_plots/newformat/bubble_ce_lv3_acc}\\[0.5em]
    \input{images/bubble_plots/newformat/bubble_ce_lv3_f1}
  }
  \subcaptionbox{Performance gain for \gls{gl:fl} \gls{gl:dirpa}  compared to its \gls{gl:fl} baseline. \label{fig:fl_results_lv3}}[.48\linewidth]
  {%
    \centering
    \input{images/bubble_plots/newformat/bubble_fl_lv3_acc}\\[0.5em]
    \input{images/bubble_plots/newformat/bubble_fl_lv3_f1}
  }
  \par
  \centering
  \vspace{1em}
  \subcaptionbox*{}[\linewidth]
  {\input{images/bubble_plots/legend}}  
  \vspace{-1em}
  \caption{%
    Performance gains on \gls{gl:hcat} level \num{3} achieved by applying \gls{gl:dirpa}, visualized relative to the Bhattacharyya distance (\cf \cref{tab:dBEuroCropsML}) and Gini coefficient (\cf \cref{tab:gini}).
    Circular markers denote performance gains, while crosses represent losses of the \gls{gl:dirpa} method.
    The marker sizes indicate the magnitude of gains or losses in metrics compared to the respective baseline.
    An increasing $D_\text{B}$ is equivalent to a decreasing number of $k$ in the few-shot scenario and represents a bigger prior shift.
     For Lithuania and Spain, $D_\text{B}$ stays constant since the relative class distribution remains the same in each few-shot task.
  }
  \label{fig:results_bubbles_lv3}
\end{figure*}
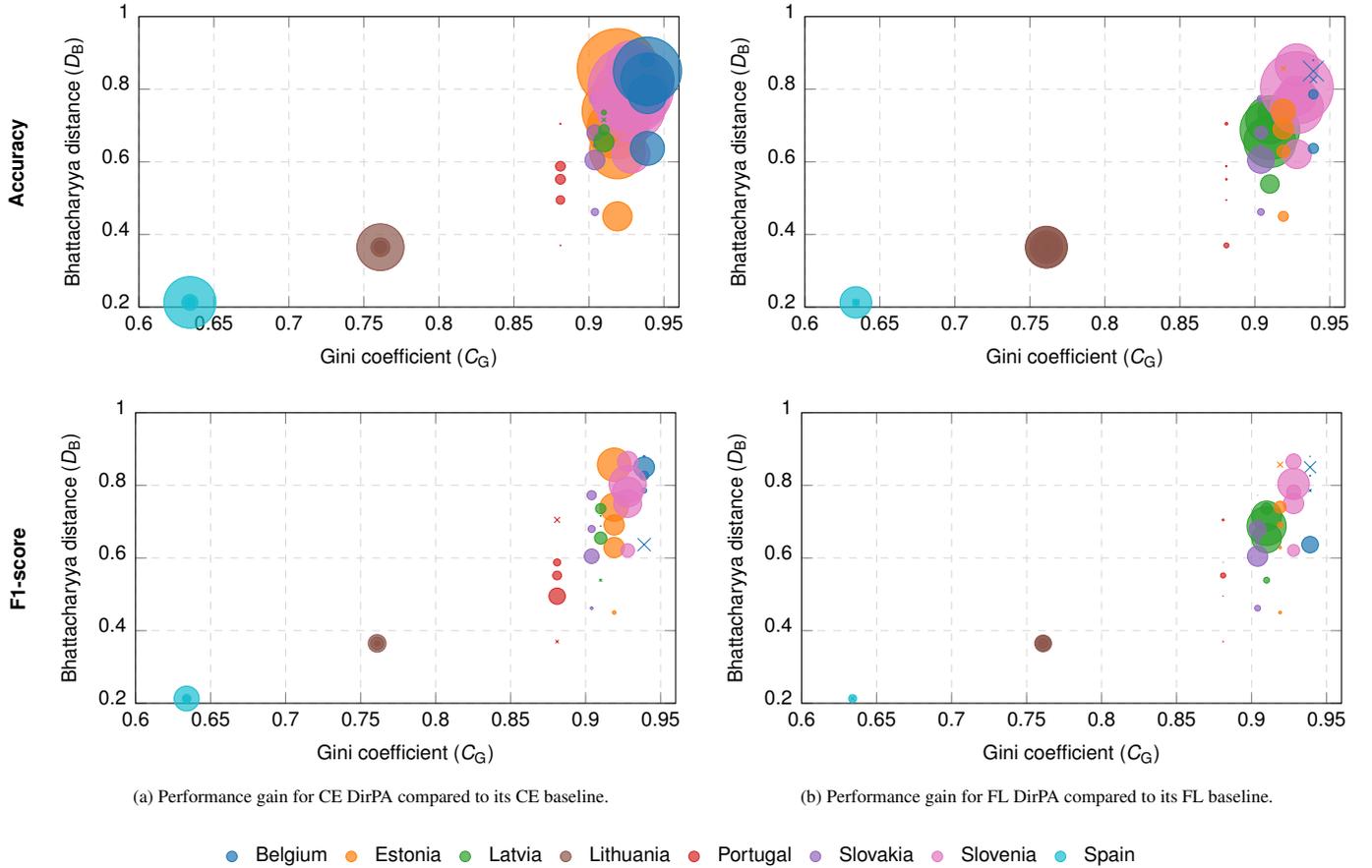

Cohen's kappa serves as our main validation and evaluation metric.
In addition, in order to analyze the trade-off between overall system robustness, overall performance, and class-specific performance, we also report \emph{overall classification accuracy} $a_\text{OA}$ (micro-averaged) and \emph{F1-score} (macro-averaged).
\Cref{tab:results_BE,tab:results_EE,tab:results_LV,tab:results_LT,tab:results_PT,tab:results_SI,tab:results_SK,tab:results_ES} show the test metrics for each country and algorithm.
Furthermore, a visual representation of performance gains in relation to a country's Gini coefficient (\cf \cref{tab:gini}) and the Bhattacharyya distance between the test set and each few-shot training set (\cf \cref{tab:dBEuroCropsML}) is shown in \cref{fig:results_bubbles,fig:results_bubbles_lv3}.
The postfix \gls{gl:dirpa} is appended to the name of the loss function if the \gls{gl:dirpa} method has been utilized.
All test metrics are reported using the best-performing model, measured by validation kappa.
In addition to the most-granular classification results on \gls{gl:hcat} level \num{6}, we also report test accuracies and F1-scores on broader land cover classes (\gls{gl:hcat} level \num{3}).
Note that the training process does not directly leverage the hierarchical structure of the dataset and is performed only at the granular class level \num{6}.
Predictions at level \num{3} are derived by mapping the predicted level \num{6} crop type to its corresponding parent category within the hierarchical structure.
In the $1$-shot scenarios, \gls{gl:ce} and \gls{gl:ce} \gls{gl:ros} yield identical results, since no oversampling occurs when each class is expected to have only one sample.
Given the architectural differences between \gls{gl:presto} and our pre-trained model, we evaluate the results independently from the main comparison.

\begin{table*}
    \centering
    \caption{%
    Few-shot classification results for \textbf{Belgium}.
    We show the results for the most-granular crop types, as presented in \cref{tab:finetuningsplit_numbers}, as well as for the broader land cover classes (\gls{gl:hcat} level \num{3}).
    We report the metrics' means $\pm$ standard deviations on the test set for each variant and few-shot task, computed over five runs, \cf \cref{sec:experiments}. 
    The best result for each few-shot scenario is highlighted in {\color{blue}\textbf{bold blue}}.
    Depending on which loss function achieves the best result, the top result of the other loss function is shown in \textbf{bold black}. 
    The best-performing \gls{gl:presto} model is highlighted in {\color{teal}\textbf{bold teal}}.}
    
    \scriptsize
    \begin{subtable}{\linewidth}
        \centering
        \caption{%
            Fine-grained classification results on \gls{gl:hcat} level \num{6}.
            \label{tab:results_BE_lv6}
        }
        \input{tables/results_revision/BE} 
        \vspace{10pt}
    \end{subtable}%
    \hfill
    \begin{subtable}{\linewidth}
        \centering
        \caption{%
            Classification results on \gls{gl:hcat} level \num{3}.
            \label{tab:results_BE_lv3}
        }
        \input{tables/results_revision/parentlv3/BE} 
    \end{subtable}%
    \label{tab:results_BE}
\end{table*}

\begin{table*}
    \centering
    \caption{%
    Few-shot classification results for \textbf{Estonia}.
    We show the results for the most-granular crop types, as presented in \cref{tab:finetuningsplit_numbers}, as well as for the broader land cover classes (\gls{gl:hcat} level \num{3}).
    We report the metrics' means $\pm$ standard deviations on the test set for each variant and few-shot task, computed over five runs, \cf \cref{sec:experiments}. 
    The best result for each few-shot scenario is highlighted in {\color{blue}\textbf{bold blue}}.
    Depending on which loss function achieves the best result, the top result of the other loss function is shown in \textbf{bold black}. 
    The best-performing \gls{gl:presto} model is highlighted in {\color{teal}\textbf{bold teal}}.
    }    
    \scriptsize
    \begin{subtable}{\linewidth}
        \centering
    \caption{%
        Fine-grained classification results on \gls{gl:hcat} level \num{6}.
        \label{tab:results_EE_lv6}
    }
        \input{tables/results_revision/EE} 
        \vspace{10pt}
    \end{subtable}%

    \begin{subtable}{\linewidth}
        \centering
    \caption{%
        Classification results on \gls{gl:hcat} level \num{3}.
        \label{tab:results_EE_lv3}
    }
        \input{tables/results_revision/parentlv3/EE} 
    \end{subtable}%
    \label{tab:results_EE}
\end{table*}

\begin{table*}
    \centering
    \caption{%
    Few-shot classification results for \textbf{Latvia}.
    We show the results for the most-granular crop types, as presented in \cref{tab:finetuningsplit_numbers}, as well as for the broader land cover classes (\gls{gl:hcat} level \num{3}).
    We report the metrics' means $\pm$ standard deviations on the test set for each variant and few-shot task, computed over five runs, \cf \cref{sec:experiments}. 
    The best result for each few-shot scenario is highlighted in {\color{blue}\textbf{bold blue}}.
    Depending on which loss function achieves the best result, the top result of the other loss function is shown in \textbf{bold black}. 
    }    
   \scriptsize
    \begin{subtable}{\linewidth}
        \centering
    \caption{%
        Fine-grained classification results on \gls{gl:hcat} level \num{6}.
        \label{tab:results_LV_lv6}
    }
        \input{tables/results_revision/LV} 
        \vspace{10pt}
    \end{subtable}%

    \begin{subtable}{\linewidth}
        \centering
    \caption{%
        Classification results on \gls{gl:hcat} level \num{3}.
        \label{tab:results_LV_lv3}
    }
        \input{tables/results_revision/parentlv3/LV} 
        \vspace{10pt}
    \end{subtable}%
    \label{tab:results_LV}
\end{table*}

\begin{table*}
    \centering
    \caption{%
    Few-shot classification results for \textbf{Lithuania}.
    We show the results for the most-granular crop types, as presented in \cref{tab:finetuningsplit_numbers}, as well as for the broader land cover classes (\gls{gl:hcat} level \num{3}).
    We report the metrics' means $\pm$ standard deviations on the test set for each variant and few-shot task, computed over five runs, \cf \cref{sec:experiments}. 
    The best result for each few-shot scenario is highlighted in {\color{blue}\textbf{bold blue}}.
    Depending on which loss function achieves the best result, the top result of the other loss function is shown in \textbf{bold black}. 
    }    
    \scriptsize
    \begin{subtable}{\linewidth}
        \centering
    \caption{%
        Fine-grained classification results on \gls{gl:hcat} level \num{6}.
        \label{tab:results_LT_lv6}
    }
        \input{tables/results_revision/LT} 
        \vspace{10pt}
    \end{subtable}%
    \hfill
    \begin{subtable}{\linewidth}
        \centering
    \caption{%
        Classification results on \gls{gl:hcat} level \num{3}.
    \label{tab:results_LT_lv3}
    }
        \input{tables/results_revision/parentlv3/LT} 
        \vspace{10pt}
    \end{subtable}%
    \label{tab:results_LT}
\end{table*}

\begin{table*}
    \centering
    \caption{%
    Few-shot classification results for \textbf{Portugal}.
    We show the results for the most-granular crop types, as presented in \cref{tab:finetuningsplit_numbers}, as well as for the broader land cover classes (\gls{gl:hcat} level \num{3}).
    We report the metrics' means $\pm$ standard deviations on the test set for each variant and few-shot task, computed over five runs, \cf \cref{sec:experiments}. 
    The best result for each few-shot scenario is highlighted in {\color{blue}\textbf{bold blue}}.
    Depending on which loss function achieves the best result, the top result of the other loss function is shown in \textbf{bold black}. 
    }    
    \scriptsize
    \begin{subtable}{\linewidth}
    \caption{%
        Fine-grained classification results on \gls{gl:hcat} level \num{6}.
    }
    \label{tab:results_PT_lv6}
        \centering
        \input{tables/results_revision/PT} 
        \vspace{10pt}
    \end{subtable}%
    \hfill
    \begin{subtable}{\linewidth}
        \caption{%
            Classification results on \gls{gl:hcat} level \num{3}.
        }
        \label{tab:results_PT_lv3}
        \centering
        \input{tables/results_revision/parentlv3/PT} 
        \vspace{10pt}
    \end{subtable}%
    \label{tab:results_PT}
\end{table*}

\begin{table*}
    \centering
    \caption{%
    Few-shot classification results for \textbf{Slovakia}.
    We show the results for the most-granular crop types, as presented in \cref{tab:finetuningsplit_numbers}, as well as for the broader land cover classes (\gls{gl:hcat} level \num{3}).
    We report the metrics' means $\pm$ standard deviations on the test set for each variant and few-shot task, computed over five runs, \cf \cref{sec:experiments}. 
    The best result for each few-shot scenario is highlighted in {\color{blue}\textbf{bold blue}}.
    Depending on which loss function achieves the best result, the top result of the other loss function is shown in \textbf{bold black}. 
    }    
    \scriptsize    
    \begin{subtable}{\linewidth}
        \centering
        \caption{%
            Fine-grained classification results on \gls{gl:hcat} level \num{6}.
            \label{tab:results_SK_lv6}
        }
        \input{tables/results_revision/SK} 
        \vspace{10pt}
    \end{subtable}%

    \begin{subtable}{\linewidth}
        \centering
        \caption{%
            Classification results on \gls{gl:hcat} level \num{3}.
        \label{tab:results_SK_lv3}
        }
        \input{tables/results_revision/parentlv3/SK} 
        \vspace{10pt}
    \end{subtable}%
    \label{tab:results_SK}
\end{table*}

\begin{table*}
    \centering
    \caption{%
    Few-shot classification results for \textbf{Slovenia}.
    We show the results for the most-granular crop types, as presented in \cref{tab:finetuningsplit_numbers}, as well as for the broader land cover classes (\gls{gl:hcat} level \num{3}).
    We report the metrics' means $\pm$ standard deviations on the test set for each variant and few-shot task, computed over five runs, \cf \cref{sec:experiments}. 
    The best result for each few-shot scenario is highlighted in {\color{blue}\textbf{bold blue}}.
    Depending on which loss function achieves the best result, the top result of the other loss function is shown in \textbf{bold black}. 
   The best-performing \gls{gl:presto} model is highlighted in {\color{teal}\textbf{bold teal}}.
    }    
    \scriptsize
    \begin{subtable}{\linewidth}
        \centering
        \caption{%
            Fine-grained classification results on \gls{gl:hcat} level \num{6}.
            \label{tab:results_SI_lv6}
        }
        \input{tables/results_revision/SI} 
        \vspace{10pt}
    \end{subtable}%
    \hfill
    \begin{subtable}{\linewidth}
        \centering
        \caption{%
            Classification results on \gls{gl:hcat} level \num{3}.
            \label{tab:results_SI_lv3}
        }
        \input{tables/results_revision/parentlv3/SI} 
        \vspace{10pt}
    \end{subtable}%
    \label{tab:results_SI}
\end{table*}

\begin{table*}
    \centering
    \caption{%
    Few-shot classification results for \textbf{Spain}.
    We show the results for the most-granular crop types, as presented in \cref{tab:finetuningsplit_numbers}, as well as for the broader land cover classes (\gls{gl:hcat} level \num{3}).
    We report the metrics' means $\pm$ standard deviations on the test set for each variant and few-shot task, computed over five runs, \cf \cref{sec:experiments}. 
    The best result for each few-shot scenario is highlighted in {\color{blue}\textbf{bold blue}}.
    Depending on which loss function achieves the best result, the top result of the other loss function is shown in \textbf{bold black}. 
    The best-performing \gls{gl:presto} model is highlighted in {\color{teal}\textbf{bold teal}}.
    }    
    \scriptsize
    \begin{subtable}{\linewidth}
        \centering
        \caption{%
            Fine-grained classification results on \gls{gl:hcat} level \num{6}.
            \label{tab:results_ES_lv6}
        }
        \input{tables/results_revision/ES}

        \vspace{10pt}
    \end{subtable}%
    \hfill
    \begin{subtable}{\linewidth}
        \centering
        \caption{%
            Classification results on \gls{gl:hcat} level \num{3}.
            \label{tab:results_ES_lv3}
        }
        \input{tables/results_revision/parentlv3/ES} 
        \vspace{10pt}
    \end{subtable}%
    \label{tab:results_ES}
\end{table*}


\begin{figure*}[t!]
  \centering
  \subcaptionbox{Performance gain for \gls{gl:ce} \gls{gl:dirpa} compared to its \gls{gl:ce} baseline. \label{fig:ce_results}}[.48\linewidth]
  {%
    \centering
    \input{images/bubble_plots/newformat/bubble_ce_ck}\\[0.5em]
    \input{images/bubble_plots/newformat/bubble_ce_acc}\\[0.5em]
    \input{images/bubble_plots/newformat/bubble_ce_f1}
  }
  \subcaptionbox{Performance gain for \gls{gl:fl} \gls{gl:dirpa} compared to its \gls{gl:fl} baseline. \label{fig:fl_results}}[.48\linewidth]
  {%
    \centering
    \input{images/bubble_plots/newformat/bubble_fl_ck}\\[0.5em]
    \input{images/bubble_plots/newformat/bubble_fl_acc}\\[0.5em]
    \input{images/bubble_plots/newformat/bubble_fl_f1}
  }
  \par
  \centering
  \vspace{1em}
  \subcaptionbox*{}[\linewidth]
  {\input{images/bubble_plots/legend}}  
  \vspace{-1em}
  \caption{%
    Performance gains achieved by applying \gls{gl:dirpa}, visualized relative to the Bhattacharyya distance (\cf \cref{tab:dBEuroCropsML}) and Gini coefficient (\cf \cref{tab:gini}).
    Circular markers denote performance gains, while crosses represent losses of the \gls{gl:dirpa} method.
    The marker sizes indicate the magnitude of gains or losses compared to the respective baseline.
    An increasing $D_\text{B}$ is equivalent to a decreasing number of $k$ in the few-shot scenario and represents a bigger prior shift.
     For Lithuania and Spain, $D_\text{B}$ stays constant since the relative class distribution remains the same in each few-shot task.
  }
  \label{fig:results_bubbles}
\end{figure*}

\begin{description}
\item[Belgium] 
\Cref{tab:results_BE_lv6} shows the classification results in Belgium.
Across all few-shot settings, the combination of \gls{gl:ce} \gls{gl:dirpa} consistently outperforms the baseline (and all other methods) in both Cohen's kappa and overall accuracy.
For kappa, \gls{gl:dirpa} achieves a relative performance gain of up to \num{33}\% ($5$-shot), maintaining a steady improvement of \num{10}\% even in the $100$-shot.
Furthermore, it also achieves the highest F1-scores for $1$- and $5$-shot.
While \gls{gl:fl} with prior augmentation generally improves performance, it falls short of the baseline in the $5$-shot benchmark, though it remains effective for F1-scores for $10$- and $100$-shot.
In terms of broader parent-level metrics, presented in \cref{tab:results_BE_lv3}, \gls{gl:dirpa} again shows superior performance for accuracy when paired with \gls{gl:ce}.
For F1-score, \gls{gl:ce} \gls{gl:dirpa} outperforms the baseline in the $5$- and $20$-shot settings.
When incorporated into \gls{gl:fl}, it, however, often falls short in terms of F1-scores, compared to its baseline.

\item[Estonia] 
When combined with \gls{gl:ce}, \gls{gl:dirpa} shows consistent superior performance compared to the \gls{gl:ce} baseline (and all other options) in terms of kappa and accuracy (\cref{tab:results_EE_lv6}).
The highest relative performance gain in kappa even reaches \num{53}\% ($1$-shot), with gains again persisting at \num{9}\% even in the $100$-shot case.
For $1$-shot, it surpasses the remaining methods in terms of F1-score.
For the combination of \gls{gl:fl} and prior augmentation, we can observe a similar pattern to the one in Belgium, where \gls{gl:dirpa} enables \gls{gl:fl} to often surpass the \gls{gl:ce} baseline in terms of kappa and accuracy.
Furthermore, for $5$-shot, it improves all three metrics compared to its baseline.
Moreover, \gls{gl:dirpa} yields noticeable gains on broader parent-level metrics (\cref{tab:results_EE_lv3}), consistently outperforming the non-\gls{gl:dirpa} baselines, except when paired with \gls{gl:fl} in the $1$-shot benchmark task.
The greatest improvement is observed for $1$-shot \gls{gl:ce}, where \gls{gl:dirpa} achieves \num{27}\% performance gain in accuracy and \num{31}\% in F1-score.

\item[Latvia] 
For Latvia (\cref{tab:results_LV_lv6}), incorporating prior augmentation with \gls{gl:fl} yields substantial gains in kappa and accuracy across all few-shot tasks.
Most notably, for $5$-shot, \gls{gl:dirpa} improves all three metrics compared to the \gls{gl:fl} baseline, showing increases of \num{41}\% (kappa), \num{44}\% (accuracy), and \num{9}\% (F1-score).
Although the improvements are less consistent when paired with \gls{gl:ce}, the method still provides a performance boost in terms of kappa and accuracy for $1$-, $10$-, and $20$-shot settings.
Similar results are observed for \gls{gl:hcat} level \num{3} in \cref{tab:results_LV_lv3}, where \gls{gl:fl} \gls{gl:dirpa} is the top performer in nearly every scenario, with the sole exception of the F1-score at \num{100}-shots.
Meanwhile, adding prior augmentation to \gls{gl:ce} improves both accuracy and F1-scores on parent-level classes for $1$-, $10$-, and $20$-shot settings.

\item[Lithuania]
For fine-grained classification (\cref{tab:results_LT_lv6}), incorporating \gls{gl:dirpa} yields considerable improvements in both Cohen's kappa and accuracy across all few-shot scenarios and loss functions.
When combined with \gls{gl:fl}, the method also enhances F1-scores across nearly all benchmarks, except for the $10$-shot scenario, where it exhibits a loss of around \num{7}\%.
For \gls{gl:ce}, this multi-metric improvement is observed for both $1$- and $20$-shot tasks.
Regarding \gls{gl:hcat} level \num{3} performance (\cref{tab:results_LT_lv3}), \gls{gl:dirpa} consistently outperforms its baselines across both evaluation metrics and all few-shot settings, regardless of the loss function.

\item[Portugal]
The results for Portugal (\cref{tab:results_PT}) are more nuanced.
However, \gls{gl:dirpa} still facilitates several key improvements.
For \gls{gl:hcat} level \num{6}, it achieves joint superiority across all three metrics for $10$- and $20$-shot when used with \gls{gl:ce} and for $10$-, $20$-, and even $100$-shot when adding it to \gls{gl:fl}.
Parent-level metrics also benefit substantially, particularly in the mid-range few-shot tasks ($k \in \{5, 10, 20\}$) for \gls{gl:ce} and for $k \in \{1, 10, 100\}$ for \gls{gl:fl}.

\item[Slovakia]
For fine-grained classification in Slovakia (\cref{tab:results_SK_lv6}), adding Dirichlet priors to \gls{gl:ce} leads to a marked improvement in kappa and accuracy for $k\geq5$, with $5$- and $10$-shot tasks showing superiority across all three metrics.
When paired with \gls{gl:fl}, \gls{gl:dirpa} outperforms the standard \gls{gl:fl} baseline for every few-shot scenario in terms of kappa and accuracy, while also achieving higher F1-scores for $5$-, $10$-, and $20$-shot.
On broader parent-level crop types (\cref{tab:results_SK_lv3}), adding Dirichlet prior augmentation consistently achieves higher accuracies and F1-scores, except for \gls{gl:ce} in the $10$-shot benchmark, where it matches its baseline.
 
\item[Slovenia]
Adding prior augmentation consistently improves both Cohen's kappa and accuracy for all few-shot scenarios, regardless of the loss function used (\cref{tab:results_SI_lv6}).
Furthermore, for $k \in \{1, 5, 20\}$, it either improves or matches ($20$-shot \gls{gl:ce}) the F1-scores.
For the $10$- and $100$-shot scenario, \gls{gl:dirpa} falls slightly behind the baselines, showing a marginally lower F1-score of \num{8}\% or \num{12}\% when incorporated with \gls{gl:ce} and \num{5}\% or \num{14}\% when added to \gls{gl:fl}.
This performance advantage is even more pronounced for parent-level classes (\cref{tab:results_SI_lv3}), where \gls{gl:dirpa} consistently achieves higher accuracies and F1-scores across all few-shot tasks and loss functions.

\item[Spain]
For few-shot classification in Spain (\cref{tab:results_ES}), incorporating prior augmentation with \gls{gl:ce} results in consistent---and most often substantial---improvements across all three metrics for both \gls{gl:hcat} levels.
The sole exception is $5$-shot F1, where the decline is negligible at just \num{2}\%.
In contrast, when paired with \gls{gl:fl}, \gls{gl:dirpa} shows less consistent gains, particularly regarding F1-scores.
However, for $1$-, $5$-, and $20$-shot, it substantially enhances both kappa and accuracy for \gls{gl:hcat} level \num{6} (\cf \cref{tab:results_ES_lv6}).
In the $20$-shot scenario, \gls{gl:dirpa} demonstrates superiority over its \gls{gl:fl} baseline across all three metrics.
Notably, in contrast to other countries, the combination of \gls{gl:fl} and \gls{gl:dirpa} appears less effective for parent-level classes (\cref{tab:results_ES_lv3}), where accuracy and F1-score improvements are confined to the $1$- and $20$-shot benchmarks.

\item[\gls{gl:ce} \gls{gl:ros}]
Regarding Cohen's kappa and accuracy, \gls{gl:ce} \gls{gl:dirpa} outperforms random oversampling in most cases, with the latter often falling below the \gls{gl:ce} baseline.
Random oversampling achieves its strongest results when measured in F1-score, outperforming other methods at \gls{gl:hcat} level \num{6} in Latvia for $1$- and $5$-shot (\cref{tab:results_LV_lv6}) and Lithuania for $10$- and $100$-shot (\cref{tab:results_LT_lv6}).
For $1$-shot in Slovakia it even shows superior performance across all metrics (\cref{tab:results_SK_lv6}).
Additional gains are observed at the parent level in Belgium for $1$- and $10$-shot (\cref{tab:results_BE_lv3}) and Portugal for $1$-shot (\cref{tab:results_PT_lv3}) in terms of both accuracy and F1-score.
However, these results remain limited to specific settings and do not generalize across all metrics or a broader range of scenarios.

\item[PRESTO]
For \gls{gl:ce} \gls{gl:presto}, evaluated on Belgium (\cref{tab:results_BE_lv6}), Estonia (\cref{tab:results_EE_lv6}), Slovenia (\cref{tab:results_SI_lv6}) and Spain (\cref{tab:results_ES_lv6}), incorporating prior augmentation consistently improves Cohen's kappa and accuracy for $k \leq 10$.
Slovenia and Spain retain these gains across all few-shot settings, as does Estonia up to $20$-shot. 
F1-scores at \gls{gl:hcat} level \num{6} decrease when adding \gls{gl:dirpa} across all countries and shot settings.
At the broader \gls{gl:hcat} level \num{3}, \gls{gl:dirpa} improves accuracy across most settings and countries, with Estonia (\cref{tab:results_ES_lv3}) showing stable gains in both accuracy and F1-score across all few-shot settings.

\end{description}

\section{Discussion}\label{sec:discussion}

\begin{figure}[t]
  \centering
  \subcaptionbox*{\label{fig:dir_a5_softmax}}[.6\linewidth]{\includegraphics[width=\linewidth, keepaspectratio]{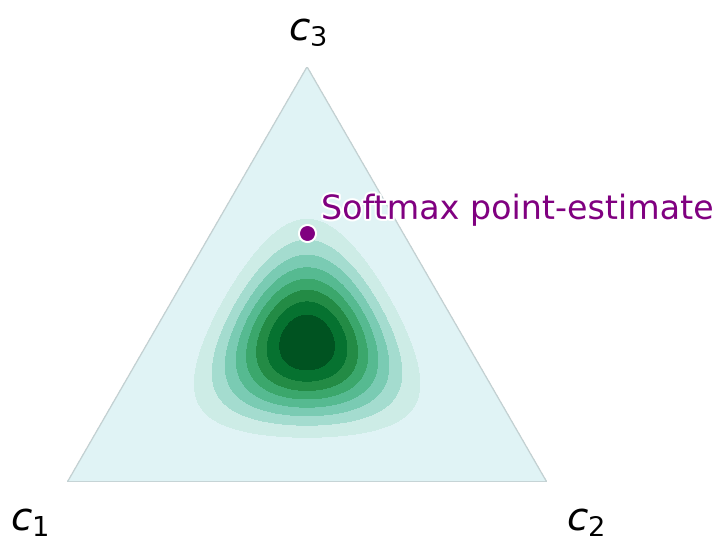}}
  \hspace{1em}
  \subcaptionbox*{\label{fig:dir_cbar_disc}}[.05\linewidth]{%
    \resizebox{!}{3.9cm}{\input{images/dirichlet/dirichlet_densitycolorbar}}
  }\hspace*{0.04\linewidth}
  \caption{Comparison of predictive output space on the $2$-simplex $\Delta^2$ for a three-class classification $(c_1, c_2, c_3)$.
  The Softmax function reduces feature information to a single coordinate on the simplex, producing a point estimate.
  In contrast, the \gls{gl:DirDist} maps features to a probability density over the simplex, thereby capturing epistemic uncertainty.}
  \label{fig:dirichlet_softmax}
\end{figure}

The empirical results in \cref{sec:results} across eight diverse countries show a consistent and significant performance gain over standard baselines, particularly in low-shot regimes ($k \leq 20$), as shown in \cref{fig:results_bubbles,fig:results_bubbles_lv3}.
This underscores the cross-dataset stability of the \gls{gl:dirpa} method, suggesting that the Dirichlet prior induces a form of universal regularization and improves model robustness.
The chance-corrected gains in Cohen's kappa across all countries confirm that these improvements represent a genuine increase in discriminative power, rather than being a result of frequentist exploitation.

In low-shot regimes, when training sets are inherently balanced by the $k$-shot sampling strategy, and data is extremely scarce, \gls{gl:dirpa} prevents feature hallucination.
While standard Softmax forces a hard point-estimate choice, \gls{gl:dirpa} maps the features to a density on the $(K-1)$-simplex, as visualized in \cref{fig:dirichlet_softmax}.
This allows the model to capture uncertainty arising from data scarcity, preventing overconfidence in poor features and leading to higher kappa scores.
As a result, \gls{gl:dirpa} maximizes system-wide reliability, achieving large gains in kappa and accuracy, \cf \cref{fig:results_bubbles}.
In contrast, \gls{gl:dirpa} outperforms random oversampling, which in several scenarios even falls below the \gls{gl:ce} baseline, suggesting that duplicating scarce samples does not resolve the underlying prior shift.

Conversely, in high-shot regimes, where the natural class imbalance begins to re-emerge in the training data (\cref{tab:dBEuroCropsML}), the prior augmentation shifts to preventing majority-class dominance.
Consequently, higher performance gains in class-specific metrics are observed for higher $k$ (equivalent to lower $D_\text{B}$), meaning bigger prior shifts.
Importantly, integrating \gls{gl:dirpa} does not introduce performance penalties in higher-shot environments.
Confirming our previous findings in \citet{Reuss26:DirPA}, all evaluated methods converge to a similar performance ceiling in high-shot scenarios.
Ultimately, the efficacy of \gls{gl:dirpa} is strongly correlated with the dataset's structural inequality.
Countries with extreme class imbalance (\cref{tab:gini}), like Belgium ($C_\text{G} = 0.939$), Slovenia ($C_\text{G} = 0.928$), and Estonia ($C_\text{G} = 0.919$), benefit most from this regularization, where countries with smaller Gini coefficients, such as Spain and Lithuania, often benefit less in terms of kappa, as visualized in \cref{fig:results_bubbles,fig:results_bubbles_lv3}.
Notably, these gains extend across architectures: \gls{gl:dirpa} consistently improves kappa and accuracy for \gls{gl:presto} in low-shot regimes, confirming that the prior augmentation is architecture-agnostic.
The considerable increases in Cohen's kappa combined with decreases in F1-scores for \gls{gl:presto} suggest that the \gls{gl:ce} \gls{gl:presto} baseline may have achieved relatively higher F1-scores through unstable or spurious predictions on individual classes rather than consistent discrimination, which \gls{gl:dirpa} appears to regularize.

While initial experiments \citep{Reuss26:DirPA} did not show any improvement in F1-scores, our results on the adapted, real-world dataset split, as outlined in \cref{sec:dataset_split}, often demonstrate \gls{gl:dirpa}'s superiority even in terms of F1-scores.
This indicates that for \gls{gl:dirpa} to effectively improve per-class discriminative power, the model must be exposed to training samples from all target classes.
By ensuring representation of all classes in the training set and maximizing overlap between splits, the method can correctly regularize the feature space based on the actual crop types rather than generalizing to unseen categories.
Nevertheless, most consistent improvements are observed in Cohen's kappa and overall accuracy.
As noted by \citet{Reuss26:MTLSSL}, this represents a necessary trade-off to achieve superior global performance.
The strong regularization required for overall stability can sometimes lead to over-smoothed boundaries for high-resource categories.
This minor localized loss is outweighed by the improved global classification reliability ($\kappa$), ensuring the model is more effective at handling the dataset's full class variability.
Overall, F1-scores remain constrained by the dataset's extreme class imbalance and high cardinality.
These lower values reflect the fundamental difficulty of maintaining high precision and recall across a large number of low-resource, long-tail classes, in particular in low-shot regimes.

A key finding is the superior performance of \gls{gl:dirpa} across broader parent classes, particularly in F1-scores.
This also underscores the improved system robustness achieved by incorporating Dirichlet priors.
Even in cases where the model fails to distinguish between fine-grained crop types---leading to lower fine-grained F1-scores---, it almost consistently better identifies the parent land cover category.
The higher kappa scores and overall accuracies at the fine-grained level, paired with superior parent-level F1-scores, suggest that the system's errors are often restricted to closely related sub-classes within the same correct parent category, rather than being total classification failures.
Importantly, this emerges without any explicit use of the dataset's label hierarchy during training, since \gls{gl:dirpa} operates purely at the most granular class level.
The evaluation across additional class granularities was conducted post hoc, as explained in \cref{sec:results}, and serves solely as an additional evaluation.
This indicates that \gls{gl:dirpa} additionally acts as a hierarchical stabilizer, not just a class-specific one, a promising but secondary finding that provides direction for future research.
In contrast to the performance gains for fine-grained classification, when evaluated on \gls{gl:hcat} level \num{3}, \gls{gl:dirpa} most often exhibits the highest performance gains in F1 with an increasing number of shots.
This disparity in F1-score gains across different hierarchical levels reveals that \gls{gl:dirpa}'s regularization mechanism adapts to data availability.
In very low-shot scenarios, it mitigates catastrophic misclassifications between unrelated land-cover types.
As $k$ increases and class features become better defined, it transitions from providing structural stability to refining class boundaries.
This enhances the separation between closely related crop types, thus leading to higher gains in fine-grained F1-scores.

Comparing \gls{gl:ce} \gls{gl:dirpa} with \gls{gl:fl} \gls{gl:dirpa} does not yield a universal performance advantage across all countries but rather highlights a distinct, country-level performance split.
The differences are mostly consistent across both label granularity levels and most pronounced on kappa and accuracy, while F1-scores show negligible differences between the two loss variants in almost all countries.
In countries with high prior mismatch, such as Belgium and Estonia (\cf Bhattacharyya distance, \cref{tab:dBEuroCropsML}), \gls{gl:ce} already performs comparably to or better than the \gls{gl:fl} at baseline and 
\gls{gl:ce} \gls{gl:dirpa} consistently achieves larger gains over its baseline than \gls{gl:fl} \gls{gl:dirpa} does over its own.
A possible explanation is that under strong prior mismatch, the prior adjustment interacts more effectively with \gls{gl:ce}, while the combination of \gls{gl:fl}'s focusing parameter $\gamma$ and the prior adjustment results in less pronounced improvements, suggesting that the two mechanisms partially overlap when applying corrections, even if both individually remain beneficial.
Slovenia shows a similar pattern in terms of gain magnitude, though the two baselines are nearly identical there, making it less conclusive.
Lithuania shows the opposite pattern.
\Gls{gl:fl} already outperforms the \gls{gl:ce} baseline, and \gls{gl:fl} \gls{gl:dirpa} additionally achieves larger gains over its baseline than \gls{gl:ce}.
Difficulty-weighting and prior adjustment are here genuinely complementing each other.
Latvia presents a particularly interesting case.
The two baselines perform nearly identical, yet \gls{gl:fl} \gls{gl:dirpa} achieves substantially larger gains than its \gls{gl:ce} counterpart, suggesting a strong and specific interaction between \glsxtrlong{gl:fl} and the \gls{gl:dirpa} adjustment that warrants further investigation in future work.
Slovakia and Portugal show small and largely symmetric gains for both variants with no clear winner, while Spain, despite having the lowest and constant prior mismatch (\cf \cref{tab:dBEuroCropsML}), shows larger relative gains for \gls{gl:ce} \gls{gl:dirpa}.
This is possibly because when training and test distributions are already closely aligned, the prior adjustment has less corrective work to do and the additional interaction between \gls{gl:fl}'s difficulty-weighting and the prior adjustment provides little further benefit over \gls{gl:ce} \gls{gl:dirpa}.

While the symmetric Dirichlet ($\boldsymbol{\alpha} = \alpha \cdot \boldsymbol{1}$) serves as our robust baseline, we also investigated an asymmetric variant \citep{Reuss26:DirPA}.
Our intuition was to amplify the signal for a selected focus class, which we randomly sample at each step $s$ and assign a different value to its concentration parameter $\alpha$.
However, empirical results on Estonia, Belgium, and Slovenia showed that this asymmetric sampling did not yield a performance gain over the simpler symmetric version.
This suggests that the original method already provides sufficient regularization and that the additional complexity of a class-specific focus does not offer further discriminative advantages.

\section{Conclusion}\label{sec:conclusion}

This study presents a geographically extended application of the \glsxtrlong{gl:dirpa} method, which aims to bridge the prior shift between balanced few-shot training sets and imbalanced test data.
By augmenting the training data with dynamically sampled Dirichlet pseudo-priors, \gls{gl:dirpa} encourages the model to learn feature representations that are robust across a variety of potential class distributions.
Instead of converging to a rigid point estimate based on a fixed prior, the model is regularized to maintain discriminative power across the $(K-1)$-simplex, capturing the uncertainty inherent in few-shot regimes.
Notably, as \gls{gl:dirpa} is already applied during the fine-tuning training stage, it results in a fixed pipeline requiring no further adjustment after deployment at inference time---a major practical advantage for non-technical users such as agriculturalists.

Our evaluation across eight \gls{gl:eu} countries and various model architectures demonstrates the method's cross-dataset and architectural robustness, yielding consistent improvements in Cohen's kappa and overall accuracy.
In particular for low-shot regimes, \gls{gl:dirpa} maximizes system-wide reliability.
Notably, the most substantial gains were observed in countries with extreme class imbalance (high Gini coefficients).
This confirms that \gls{gl:dirpa} effectively mitigates the structural inequalities inherent in real-world few-shot crop classification datasets.
Crucially, our experiments with a realistic, comprehensive dataset split demonstrate that \gls{gl:dirpa}'s ability to enhance F1-scores depends on the model being exposed to all target classes during training, rather than attempting to generalize to zero-shot categories \citep[\cf][]{Reuss26:DirPA}.
Given the highest gains in higher-shot scenarios, we found that \gls{gl:dirpa} maximizes class-specific performance when there is enough data to learn class boundaries.

Furthermore, while the method occasionally trades off fine-grained class-specific performance due to the over-smoothing of decision boundaries, it notably enhances hierarchical consistency, ensuring that parent-level land cover classes are correctly identified even when sub-class distinctions fail.
This superior performance provides a compelling argument for incorporating \gls{gl:dirpa} into hierarchical crop-type classification frameworks in future work.

Despite the effectiveness of the \gls{gl:dirpa} method, challenges remain.
For instance, its performance gain is highly dependent on identifying the optimal values for hyperparameters $\alpha$ and $\tau$, \cf \ref{sec:hyperparameters_dirpa}.
Hence, additional preliminary experiments aimed at dynamically learning the Dirichlet concentration parameter $\alpha$.
However, we were not yet able to achieve significant improvements over the fixed-parameter setup.
These initial results suggest that optimizing $\alpha$ is non-trivial and warrants further investigation.

\section*{CRediT authorship contribution statement}
\textbf{J{.} Reuss:} Conceptualization, Data curation, Formal analysis, Investigation, Methodology, Resources, Software, Validation, Visualization, Writing -- original draft, Writing -- review \& editing. \textbf{E{.} Gikalo:} Investigation, Visualization, Writing -- original draft, Writing -- review \& editing. \textbf{M{.} K{\"o}rner:} Funding acquisition, Project administration, Supervision, Writing -- review \& editing.

\section*{Acknowledgments}\label{sec:acknowledgements}
The project is funded by the German Federal Ministry for Economic Affairs and Energy based on a decision by the German Bundestag under the funding references 50EE2007B (J{.} Reuss, E{.} Gikalo, and M{.} K{\"o}rner) and 50EE2105 (M{.} K{\"o}rner).

\appendix
\section{Implementation and hardware details}
We implemented all algorithms within the \texttt{PyTorch} framework \citep{Paszke19:PyTorch} for machine learning and automatic differentiation.
For all experiments, we used the \emph{Adam} optimizer.

The experiments were distributed across four virtualized machines with different hardware specifications, as detailed in \cref{tab:hardware}.

\begin{table}
    \caption{
    Hardware specifications. 
    \label{tab:hardware}
    }
    \centering
    \scriptsize
    \begin{tabular}[t]{@{}Xlllrl@{}}
      \toprule
      \multicolumn{1}{@{}X}{Node(s)} & \multicolumn{1}{X@{}}{GPU} & \multicolumn{1}{X@{}}{VRAM} & \multicolumn{1}{X@{}}{CPU} & \multicolumn{1}{X@{}}{vCPUs} & \multicolumn{1}{X@{}}{Architecture}\\
      \cmidrule(r){1-1} \cmidrule(lr){2-2} \cmidrule(lr){3-3} \cmidrule(lr){4-4} \cmidrule(lr){5-5} \cmidrule(l){6-6}
      1--3 & NVIDIA RTX A6000 & \num{48}GB & Intel Xeon & \num{12} & Ampere \\
      4 & NVIDIA A100-SXM4 & \num{40}GB & AMD EPYC & \num{24} & Ampere \\
      \bottomrule
    \end{tabular}
\end{table}

\section{Hyperparameter tuning}
\label{appendix_hyperparametertuning}

\subsection{Pre-training}
\label{sec:hyperparametertuning_pretrain}
The performance on the validation set is used to select the best hyperparameters and the model to be used for fine-tuning. 
We employ a total of \num{5000} warm-up steps (corresponding to batches) and keep the top \num{75}th percentile of runs.
We use the best validation accuracy to determine the best pre-training epoch and model state, which is subsequently used  to initialize the model backbone weights before fine-tuning.
The exact hyperparameter values or value ranges are listed in \cref{tab:pretraining_hyperparameters}.

\begin{table}
    \caption{Pre-training hyperparameters.
    The hyperparameters utilized during pre-training, along with their corresponding valid values.}
    \centering
    \renewcommand{\arraystretch}{1.2}
    \begin{tabular}[t]{@{}Xr@{}}
      \toprule
      \multicolumn{1}{@{}X}{hyperparameter} & \multicolumn{1}{X@{}}{value / value range}\\
      \cmidrule(r){1-1} \cmidrule(l){2-2}
        training/tuning epochs & \num{100}  \\
        tuning trials per setup & \num{24}  \\
        tuning warmup steps (batches) & \num{5000}  \\
        early stopping patience (epochs) & \num{15} \\
        batch size $b$ & \num{256} \\
        learning rate $\beta$ & $\left[\num{e-4}, \num{e-2}\right]$  \\
        cosine annealing cycles & \num{1} \\
        cosine annealing cycles warmup epochs & \num{1}  \\
        label smoothing factor & \num{0.1} \\
      \bottomrule
    \end{tabular}
\label{tab:pretraining_hyperparameters}
\end{table}

\subsection{Fine-tuning}
\label{sec:hyperparametertuning_finetune}
For fine-tuning, we use only \num{300} warm-up steps while again keeping the \num{75}th percentile of runs.
Furthermore, we treat backbone freezing as a tunable hyperparameter.
Finally, the model state exhibiting the best validation kappa score is used to compute the final reported model performance metrics on the fine-tuning test dataset.
The exact hyperparameter values or value ranges are listed in \cref{tab:finetuning_generalhyperparameters,tab:dirpafinetuning_hyperparameters}.

\begin{table}
    \begin{threeparttable}
    \caption{General fine-tuning hyperparameters.
    The hyperparameters utilized during fine-tuning, along with their corresponding valid values.
    \label{tab:finetuning_generalhyperparameters}
    }
    \centering
    \renewcommand{\arraystretch}{1.2}
    \begin{tabularx}{\columnwidth}{@{}Lr@{}}
      \toprule
      \multicolumn{1}{@{}L}{hyperparameter} & \multicolumn{1}{L@{}}{value / value range}\\
      \cmidrule(r){1-1} \cmidrule(l){2-2}
        training/tuning epochs & \num{200}  \\
        tuning trials per setup\textsuperscript{\textdagger} & $\{32, 40, 72, 80\}$  \\
        tuning warmup steps (batches) & \num{300}  \\
        early stopping patience (epochs) & \num{20} \\
        batch size $b$ & \num{16} \\ 
        head learning rate $\beta_\text{head}$ & $\left[\num{e-6}, \num{e-2}\right]$  \\
        backbone learning rate $\beta_\text{backbone}$ & $ \left[\num{e-7}, \num{e-3}\right]$  \\
        backbone freezing (epochs) & $\{0, 2, 5, 8, 15\}$ \\
        \glsxtrlong{gl:fl} focusing parameter $\gamma$ & $\{1.0, 1.5, 2.0, 3.0\}$\\
      \bottomrule
    \end{tabularx}
    \begin{tablenotes}
      \footnotesize
      \item[\textdagger] Depending on the algorithm and, hence, tunable parameters, we use a different number of tuning trials. 
      We set them as follows: \num{32} (\gls{gl:ce}), \num{40} (\gls{gl:fl}), \num{72} (\gls{gl:ce} with \gls{gl:dirpa}), \num{80} (\gls{gl:fl} with \gls{gl:dirpa}).
    \end{tablenotes}
  \end{threeparttable}
\end{table}

\begin{table}
    \caption{\gls{gl:dirpa} fine-tuning hyperparameters.
    The \gls{gl:dirpa} hyperparameters utilized during fine-tuning, along with their corresponding valid values.
    \label{tab:dirpafinetuning_hyperparameters}
    }
    \centering
    \renewcommand{\arraystretch}{1.2}
    \begin{tabular}[t]{@{}Xr@{}}
      \toprule
      \multicolumn{1}{@{}X}{hyperparameter} & \multicolumn{1}{X@{}}{value / value range}\\
      \cmidrule(r){1-1} \cmidrule(l){2-2}
      \gls{gl:dirpa} concentration $\alpha$ & $\{0.01,0.1,0.5,1.0,2.0,3.0,5.0\}$\\
        \gls{gl:dirpa} scaling factor $\tau$ & $\{0.5,1.0,2.0,5.0,10.0\}$\\
      \bottomrule
    \end{tabular}
\end{table}

\subsubsection{DirPA hyperparameters analysis}
\label{sec:hyperparameters_dirpa}
\Cref{tab:ce_selected_alphatau,tab:fl_selected_alphatau} show the median values of the \gls{gl:dirpa} hyperparameters $\alpha$ and $\tau$ selected for the final training runs across countries and few-shot scenarios.
Although a comparison of these values does not reveal a consistent numerical pattern or trend, optimal values for $\alpha$ are rather low, while for $\tau$, \num{10} was most often selected, resulting in a strong prior adjustment.
Furthermore, an analysis of \gls{gl:dirpa} hyperparameter importances across all investigated countries and loss functions does not provide a clear trend.
In fact, \cref{tab:ce_importance_alphatau,tab:fl_importance_alphatau} show that the relative importance of $\alpha$ and $\tau$ varies severely across countries and data regimes. 
Thus, as optimal settings depend heavily on the specific dataset, we conclude that there is no universal recommendation for configuring these hyperparameters.
We therefore recommend using a predefined search space of possible values, as detailed in \cref{tab:dirpafinetuning_hyperparameters}, and tuning to find the optimal value.

\begin{table}[t]
    \caption{%
        Median of selected optimal values of \gls{gl:dirpa} hyperparameters with \glsxtrlong{gl:ce}, calculated based on finding the optimal hyperparameter values to maximize Cohen's kappa.
        The median is calculated over five Optuna studies with different random seeds.      
        \label{tab:ce_selected_alphatau}
    }
    \centering
    \begin{subtable}{\linewidth}
        \centering
        \caption{%
            Concentration parameter $\alpha$.
            \label{tab:ce_selected_alpha}
        }
        \begin{tabular}[t]{@{}Xrrrrr@{}}
      \toprule
        Country & \multicolumn{5}{X@{}}{benchmark task ($k$-shot)} \\
        \cmidrule(l){2-6}
        {}  & \multicolumn{1}{X}{1} & \multicolumn{1}{X}{5} & \multicolumn{1}{X}{10} & \multicolumn{1}{X}{20} & \multicolumn{1}{X}{100} \\
        \cmidrule(r){1-1} \cmidrule(lr){2-2} \cmidrule(lr){3-3} \cmidrule(lr){4-4} \cmidrule(lr){5-5} \cmidrule(l){6-6}
      Belgium &  \num{0.10} & \num{1.00} & \num{2.00} & \num{0.10} & \num{0.50} \\
      Estonia &  \num{0.50} & \num{0.10} & \num{0.10} & \num{0.01} & \num{1.00} \\
      Latvia &  \num{0.50} & \num{1.00} & \num{0.50} & \num{3.00} & \num{0.75} \\
      Lithuania & \num{0.75} & \num{1.00} & \num{1.00} & \num{0.75} & \num{3.00} \\
      Portugal &  \num{1.00} & \num{0.75} & \num{0.75} & \num{1.00} & \num{1.50} \\
      Slovakia &  \num{5.00} & \num{0.75} & \num{0.50} & \num{0.75} & \num{1.00} \\
      Slovenia &  \num{2.00} & \num{2.00} & \num{0.10} & \num{0.10} & \num{1.00} \\
      Spain &  \num{0.10} & \num{3.00} & \num{0.50} & \num{2.00} & \num{3.00} \\
        \bottomrule
    \end{tabular}
        \vspace{10pt}
    \end{subtable}%
    \hfill
    \begin{subtable}{\linewidth}
        \centering
        \caption{%
            Temperature parameter $\tau$.
            \label{tab:ce_selected_tau}
        }
        \begin{tabular}[t]{@{}Xrrrrr@{}}
      \toprule
        Country & \multicolumn{5}{X@{}}{benchmark task ($k$-shot)} \\
        \cmidrule(l){2-6}
        {}  & \multicolumn{1}{X}{1} & \multicolumn{1}{X}{5} & \multicolumn{1}{X}{10} & \multicolumn{1}{X}{20} & \multicolumn{1}{X}{100} \\
        \cmidrule(r){1-1} \cmidrule(lr){2-2} \cmidrule(lr){3-3} \cmidrule(lr){4-4} \cmidrule(lr){5-5} \cmidrule(l){6-6}
      Belgium &  \num{2.0} & \num{5.0} & \num{5.0} & \num{10.0} & \num{5.0} \\
      Estonia &  \num{5.0} & \num{10.0} & \num{5.0} & \num{5.0} & \num{10.0} \\
      Latvia & \num{5.0} & \num{10.0} & \num{10.0} & \num{5.0} & \num{2.0} \\
      Lithuania & \num{10.0} & \num{5.0} & \num{5.0} & \num{2.0} & \num{5.0} \\
      Portugal & \num{2.0} & \num{5.0} & \num{2.0} & \num{1.0} & \num{1.0} \\
      Slovakia & \num{10.0} & \num{10.0} & \num{5.0} & \num{10.0} & \num{10.0} \\
      Slovenia &  \num{10.0} & \num{10.0} & \num{10.0} & \num{10.0} & \num{10.0} \\
      Spain &  \num{10.0} & \num{2.0} & \num{2.0} & \num{5.0} & \num{5.0} \\
        \bottomrule
    \end{tabular}
    \end{subtable}%
\end{table}

\begin{table}
    \caption{%
        Median of selected optimal values of \gls{gl:dirpa} hyperparameters with \glsxtrlong{gl:fl}, calculated based on finding the optimal hyperparameter values to maximize Cohen's kappa.
        The median is calculated over five Optuna studies with different random seeds.  
        \label{tab:fl_selected_alphatau}
    }
    \centering
    \begin{subtable}{\linewidth}
        \centering
        \caption{%
            Concentration parameter $\alpha$.
            \label{tab:fl_selected_alpha}
        }
        \begin{tabular}[t]{@{}Xrrrrr@{}}
      \toprule
        Country & \multicolumn{5}{X@{}}{benchmark task ($k$-shot)} \\
        \cmidrule(l){2-6}
        {}  & \multicolumn{1}{X}{1} & \multicolumn{1}{X}{5} & \multicolumn{1}{X}{10} & \multicolumn{1}{X}{20} & \multicolumn{1}{X}{100} \\
        \cmidrule(r){1-1} \cmidrule(lr){2-2} \cmidrule(lr){3-3} \cmidrule(lr){4-4} \cmidrule(lr){5-5} \cmidrule(l){6-6}
      Belgium & \num{2.00} & \num{0.10} & \num{0.50} & \num{2.00} & \num{0.50} \\
      Estonia &  \num{0.10} & \num{0.10} & \num{0.01} & \num{0.10} & \num{0.10} \\
      Latvia &  \num{1.00} & \num{0.75} & \num{0.01} & \num{1.50} & \num{1.00} \\
      Lithuania & \num{2.00} & \num{0.10} & \num{0.01} & \num{0.10} & \num{0.50} \\
      Portugal &  \num{0.50} & \num{1.00} & \num{0.50} & \num{0.50} & \num{3.00} \\
      Slovakia & \num{3.00} & \num{0.75} & \num{0.75} & \num{1.50} & \num{1.00} \\
      Slovenia & \num{0.10} & \num{0.01} & \num{1.50} & \num{0.10} & \num{0.10} \\
      Spain & \num{0.50} & \num{1.00} & \num{0.50} & \num{0.50} & \num{0.50} \\
        \bottomrule
    \end{tabular}
        \vspace{10pt}
    \end{subtable}%
    \hfill
    \begin{subtable}{\linewidth}
        \centering
        \caption{%
            Temperature parameter $\tau$.
            \label{tab:fl_selected_tau}
        }
        \begin{tabular}[t]{@{}Xrrrrr@{}}
      \toprule
        Country & \multicolumn{5}{X@{}}{benchmark task ($k$-shot)} \\
        \cmidrule(l){2-6}
        {}  & \multicolumn{1}{X}{1} & \multicolumn{1}{X}{5} & \multicolumn{1}{X}{10} & \multicolumn{1}{X}{20} & \multicolumn{1}{X}{100} \\
        \cmidrule(r){1-1} \cmidrule(lr){2-2} \cmidrule(lr){3-3} \cmidrule(lr){4-4} \cmidrule(lr){5-5} \cmidrule(l){6-6}
      Belgium & \num{10.0} & \num{10.0} & \num{10.0} & \num{10.0} & \num{1.000} \\
      Estonia & \num{10.0} & \num{2.0} & \num{5.0} & \num{10.0} & \num{5.0} \\
      Latvia &  \num{10.0} & \num{1.0} & \num{5.0} & \num{5.0} & \num{2.0} \\
      Lithuania & \num{10.0} & \num{5.0} & \num{1.0} & \num{10.0} & \num{2.0} \\
      Portugal & \num{10.0} & \num{5.0} & \num{1.0} & \num{2.0} & \num{2.0} \\
      Slovakia & \num{5.0} & \num{10.0} & \num{10.0} & \num{10.0} & \num{10.0} \\
      Slovenia & \num{2.0} & \num{10.0} & \num{10.0} & \num{5.0} & \num{10.0} \\
      Spain & \num{5.0} & \num{10.0} & \num{1.0} & \num{1.0} & \num{2.0} \\
        \bottomrule
    \end{tabular}
    \end{subtable}%
\end{table}

\begin{table}
    \caption{%
        Relative importance of \gls{gl:dirpa} hyperparameters with \glsxtrlong{gl:ce}, calculated based on finding the optimal hyperparameter values to maximize Cohen's kappa.
        The values represent the median over five Optuna studies with different random seeds.     
        The relative importance is calculated based on a total of five tunable hyperparameters.
    }
    \label{tab:ce_importance_alphatau}
    \centering
    \begin{subtable}{\linewidth}
        \centering
        \caption{%
            Concentration parameter $\alpha$.
            \label{tab:ce_importance_alpha}
        }
        \begin{tabular}[t]{@{}Xrrrrr@{}}
      \toprule
        Country & \multicolumn{5}{X@{}}{benchmark task ($k$-shot)} \\
        \cmidrule(l){2-6}
        {}  & \multicolumn{1}{X}{1} & \multicolumn{1}{X}{5} & \multicolumn{1}{X}{10} & \multicolumn{1}{X}{20} & \multicolumn{1}{X}{100} \\
        \cmidrule(r){1-1} \cmidrule(lr){2-2} \cmidrule(lr){3-3} \cmidrule(lr){4-4} \cmidrule(lr){5-5} \cmidrule(l){6-6}
      Belgium &  \num{0.139} & \num{0.341} & \num{0.189} & \num{0.230} & \num{0.332} \\
      Estonia &  \num{0.273} & \num{0.426} & \num{0.437} & \num{0.589} & \num{0.163} \\
      Latvia &  \num{0.121} & \num{0.146} & \num{0.171} & \num{0.205} & \num{0.106} \\
      Lithuania & \num{0.086} & \num{0.135} & \num{0.086} & \num{0.069} & \num{0.058} \\
      Portugal &  \num{0.111} & \num{0.140} & \num{0.200} & \num{0.147} & \num{0.086} \\
      Slovakia &  \num{0.070} & \num{0.144} & \num{0.090} & \num{0.179} & \num{0.112} \\
      Slovenia &  \num{0.290} & \num{0.376} & \num{0.242} & \num{0.351} & \num{0.120} \\
      Spain &  \num{0.250} & \num{0.205} & \num{0.090} & \num{0.165} & \num{0.191} \\
        \bottomrule
    \end{tabular}
        \vspace{10pt}
    \end{subtable}%
    \hfill
    \begin{subtable}{\linewidth}
        \centering
        \caption{%
            Temperature parameter $\tau$.
            \label{tab:ce_importance_tau}
        }
        \begin{tabular}[t]{@{}Xrrrrr@{}}
      \toprule
        Country & \multicolumn{5}{X@{}}{benchmark task ($k$-shot)} \\
        \cmidrule(l){2-6}
        {}  & \multicolumn{1}{X}{1} & \multicolumn{1}{X}{5} & \multicolumn{1}{X}{10} & \multicolumn{1}{X}{20} & \multicolumn{1}{X}{100} \\
        \cmidrule(r){1-1} \cmidrule(lr){2-2} \cmidrule(lr){3-3} \cmidrule(lr){4-4} \cmidrule(lr){5-5} \cmidrule(l){6-6}
      Belgium &  \num{0.111} & \num{0.113} & \num{0.091} & \num{0.410} & \num{0.104} \\
      Estonia &  \num{0.057} & \num{0.325} & \num{0.165} & \num{0.123} & \num{0.413} \\
      Latvia &  \num{0.148} & \num{0.109} & \num{0.220} & \num{0.194} & \num{0.118} \\
      Lithuania & \num{0.156} & \num{0.226} & \num{0.130} & \num{0.137} & \num{0.141} \\
      Portugal &  \num{0.110} & \num{0.312} & \num{0.079} & \num{0.142} & \num{0.075} \\
      Slovakia &  \num{0.040} & \num{0.298} & \num{0.251} & \num{0.249} & \num{0.224} \\
      Slovenia &  \num{0.062} & \num{0.193} & \num{0.144} & \num{0.273} & \num{0.369} \\
      Spain &  \num{0.078} & \num{0.054} & \num{0.041} & \num{0.089} & \num{0.108} \\
        \bottomrule
    \end{tabular}
    \end{subtable}%
\end{table}

\begin{table}
    \caption{%
        Relative importance of \gls{gl:dirpa} hyperparameters with \glsxtrlong{gl:fl}, calculated based on finding the optimal hyperparameter values to maximize Cohen's kappa.
        The values represent the median over five Optuna studies with different random seeds.    
        Please note that an additional hyperparameter (\gls{gl:fl} focus parameter $\gamma$) was taken into account in these studies.
        Thus, the relative importance of $\alpha$ and $\tau$ is calculated against a broader set of parameters (six in total) and, therefore, not directly comparable to the ones in \cref{tab:ce_importance_alphatau}.
    }
    \label{tab:fl_importance_alphatau}
    \centering
    \begin{subtable}{\linewidth}
        \centering
        \caption{%
            Concentration parameter $\alpha$.
            \label{tab:fl_importance_alpha}
        }
        \begin{tabular}[t]{@{}Xrrrrr@{}}
      \toprule
        Country & \multicolumn{5}{X@{}}{benchmark task ($k$-shot)} \\
        \cmidrule(l){2-6}
        {}  & \multicolumn{1}{X}{1} & \multicolumn{1}{X}{5} & \multicolumn{1}{X}{10} & \multicolumn{1}{X}{20} & \multicolumn{1}{X}{100} \\
        \cmidrule(r){1-1} \cmidrule(lr){2-2} \cmidrule(lr){3-3} \cmidrule(lr){4-4} \cmidrule(lr){5-5} \cmidrule(l){6-6}
      Belgium & \num{0.114} & \num{0.209} & \num{0.130} & \num{0.344} & \num{0.228} \\
      Estonia &  \num{0.560} & \num{0.319} & \num{0.373} & \num{0.258} & \num{0.121} \\
      Latvia &  \num{0.161} & \num{0.245} & \num{0.213} & \num{0.240} & \num{0.390} \\
      Lithuania & \num{0.082} & \num{0.284} & \num{0.472} & \num{0.415} & \num{0.219} \\
      Portugal &  \num{0.234} & \num{0.106} & \num{0.229} & \num{0.176} & \num{0.312} \\
      Slovakia &  \num{0.046} & \num{0.141} & \num{0.111} & \num{0.094} & \num{0.084} \\
      Slovenia &  \num{0.236} & \num{0.360} & \num{0.277} & \num{0.566} & \num{0.383} \\
      Spain & \num{0.341} & \num{0.237} & \num{0.200} & \num{0.168} & \num{0.237} \\
        \bottomrule
    \end{tabular}
        \vspace{10pt}
    \end{subtable}%
    \hfill
    \begin{subtable}{\linewidth}
        \centering
        \caption{%
            Temperature parameter $\tau$.
            \label{tab:fl_importance_tau}
        }
        \begin{tabular}[t]{@{}Xrrrrr@{}}
      \toprule
        Country & \multicolumn{5}{X@{}}{benchmark task ($k$-shot)} \\
        \cmidrule(l){2-6}
        {}  & \multicolumn{1}{X}{1} & \multicolumn{1}{X}{5} & \multicolumn{1}{X}{10} & \multicolumn{1}{X}{20} & \multicolumn{1}{X}{100} \\
        \cmidrule(r){1-1} \cmidrule(lr){2-2} \cmidrule(lr){3-3} \cmidrule(lr){4-4} \cmidrule(lr){5-5} \cmidrule(l){6-6}
      Belgium & \num{0.146} & \num{0.254} & \num{0.083} & \num{0.178} & \num{0.171} \\
      Estonia &  \num{0.103} & \num{0.157} & \num{0.119} & \num{0.213} & \num{0.022} \\
      Latvia &  \num{0.131} & \num{0.243} & \num{0.163} & \num{0.113} & \num{0.095} \\
      Lithuania & \num{0.115} & \num{0.176} & \num{0.093} & \num{0.057} & \num{0.107} \\
      Portugal &  \num{0.107} & \num{0.126} & \num{0.048} & \num{0.134} & \num{0.123} \\
      Slovakia &  \num{0.118} & \num{0.207} & \num{0.132} & \num{0.398} & \num{0.334} \\
      Slovenia & \num{0.071} & \num{0.136} & \num{0.081} & \num{0.118} & \num{0.173} \\
      Spain & \num{0.118} & \num{0.034} & \num{0.056} & \num{0.076} & \num{0.059} \\
        \bottomrule
    \end{tabular}
    \end{subtable}%
\end{table}

\subsubsection{Runtime analysis}
\Cref{tab:runtimes_be,tab:runtimes_ee,tab:runtimes_lt,tab:runtimes_lv,tab:runtimes_pt,tab:runtimes_sk,tab:runtimes_si,tab:runtimes_es} show a comparison of runtimes with and without the \gls{gl:dirpa} method.
While these values are not strictly comparable, as the experiments were partially conducted in parallel on multiple machines with varying hardware specifications (\cf \cref{tab:hardware}), they still give an indication of the training time required.
Furthermore, the training time is influenced by early stopping.
This may also result in greater standard deviations.
Nevertheless, no clear trend suggests that applying Dirichlet priors substantially increases training time.
In fact, \gls{gl:dirpa} often requires less training time on average than its counterpart.

\begin{table}
    \caption{
    Fine-tuning runtimes for final training on \textbf{Belgium}.
    We present the average training time (in minutes) as mean $\pm$ standard deviation over five runs (seeds).
    The fastest run time per few-shot scenario is marked in \textbf{bold}.}
    \label{tab:runtimes_be}
    \centering
    \scriptsize
    \input{tables/runtimes/BE}
\end{table}

\begin{table}
    \caption{
    Fine-tuning runtimes for final training on \textbf{Estonia}.
    We present the average training time (in minutes) as mean $\pm$ standard deviation over five runs (seeds).
    The fastest run time per few-shot scenario is marked in \textbf{bold}.}
    \label{tab:runtimes_ee}
    \centering
    \scriptsize
    \input{tables/runtimes/EE}
\end{table}

\begin{table}[t]
    \caption{
    Fine-tuning runtimes for final training on \textbf{Lithuania}.
    We present the average training time (in minutes) as mean $\pm$ standard deviation over five runs (seeds).
    The fastest run time per few-shot scenario is marked in \textbf{bold}.}
    \label{tab:runtimes_lt}
    \centering
    \scriptsize
    \input{tables/runtimes/LT}
\end{table}

\begin{table}
    \caption{
    Fine-tuning runtimes for final training on \textbf{Latvia}.
    We present the average training time (in minutes) as mean $\pm$ standard deviation over five runs (seeds).
    The fastest run time per few-shot scenario is marked in \textbf{bold}.}
    \label{tab:runtimes_lv}
    \centering
    \scriptsize
    \input{tables/runtimes/LV}
\end{table}

\begin{table}
    \caption{
    Fine-tuning runtimes for final training on \textbf{Portugal}.
    We present the average training time (in minutes) as mean $\pm$ standard deviation over five runs (seeds).
    The fastest run time per few-shot scenario is marked in \textbf{bold}.}
    \label{tab:runtimes_pt}
    \centering
    \scriptsize
    \input{tables/runtimes/PT}

\end{table}

\begin{table}
    \caption{
    Fine-tuning runtimes for final training on \textbf{Slovakia}.
    We present the average training time (in minutes) as mean $\pm$ standard deviation over five runs (seeds).
    The fastest run time per few-shot scenario is marked in \textbf{bold}.}
    \label{tab:runtimes_sk}
    \centering
    \scriptsize
    \input{tables/runtimes/SK}
\end{table}

\begin{table}
    \caption{
    Fine-tuning runtimes for final training on \textbf{Slovenia}.
    We present the average training time (in minutes) as mean $\pm$ standard deviation over five runs (seeds).
    The fastest run time per few-shot scenario is marked in \textbf{bold}.}
    \label{tab:runtimes_si}
    \centering
    \scriptsize
    \input{tables/runtimes/SI}
\end{table}

\begin{table}
    \caption{
    Fine-tuning runtimes for final training on \textbf{Spain}.
    We present the average training time (in minutes) as mean $\pm$ standard deviation over five runs (seeds).
    The fastest run time per few-shot scenario is marked in \textbf{bold}.}
    \label{tab:runtimes_es}
    \centering
    \scriptsize
    \input{tables/runtimes/ES}
\end{table}

\clearpage
\bibliographystyle{elsarticle-harv}
\bibliography{references_natbib}

\end{document}

\endinput

%% file: acronyms.tex
\newacronym[user1=Reuss26:DirPA]{gl:dirpa}{DirPA}{\textbf{Dir}ichlet \textbf{P}rior \textbf{A}ugmentation}
\newacronym{gl:symdirpa}{symDirPA}{symmetric DirPA}
\newacronym{gl:asymdirpa}{asymDirPA}{asymmetric DirPA}
\newacronym{gl:fixeddirpa}{$\text{DirPA}_{\text{fix}}$}{fixed Dirichlet Prior Augmentation}
\newacronym{gl:adaptivedirpa}{$\text{DirPA}_{\text{adapt}}$}{adaptive Dirichlet Prior Augmentation}
\newacronym[user1={Tseng23:PRESTO}]{gl:presto}{PRESTO}{\textbf{P}retrained \textbf{Re}mote \textbf{S}ensing \textbf{T}ransf\textbf{o}rmer}

\newacronym{gl:un}{UN}{United Nations}
\newacronym{gl:sdg}{SDG}{Sustainable Development Goal}
\newacronym{gl:eo}{EO}{Earth observation}
\newacronym{gl:rs}{RS}{remote sensing}
\newacronym[user1={Hall95:ndsi}]{gl:ndsi}{NDSI}{Normalized Difference Snow Index}

\newacronym{gl:ai}{AI}{artificial intelligence}
\newacronym{gl:dl}{DL}{deep learning}
\newacronym{gl:ml}{ML}{machine learning}
\newacronym{gl:fsl}{FSL}{few-shot learning}
\newacronym{gl:mtl}{MTL}{meta-learning}
\newacronym{gl:ssl}{SSL}{self-supervised learning}
\newacronym{gl:nn}{NN}{neural network}
\newacronym{gl:sota}{SOTA}{state-of-the-art}

\newacronym[user1={Finn17:MAML}]{gl:maml}{MAML}{Model-agnostic Meta-learning}
\newacronym[user1={Raghu19:ANIL}]{gl:anil}{ANIL}{Almost No Inner Loop}
\newacronym[user1={Tseng22:TIML}]{gl:timl}{TIML}{Task-informed Meta-learning}
\newacronym[user1={Lipton18:BlackBox}]{gl:bbse}{BBSE}{Black Box Shift Estimation}

\newacronym{gl:vit}{ViT}{Vision Transformer}
\newacronym{gl:cv}{CV}{computer vision}
\newacronym{gl:mlp}{MLP}{Multi-layer Perceptron}
\newacronym[user1={Lin18:FocalLoss}]{gl:fl}{FL}{focal loss}
\newacronym{gl:ce}{CE}{cross-entropy loss}
\newacronym{gl:ros}{ROS}{random oversampling}

\newacronym{gl:eu}{EU}{European Union}
\newacronym{gl:esa}{ESA}{European Space Agency}
\newacronym{gl:eea}{EEA}{European Environment Agency}
\newacronym{gl:gee}{GEE}{Google Earth Engine}

\newacronym{gl:mse}{MSE}{mean squared error}
\newacronym{gl:tpe}{TPE}{Bayesian tree-Parzen estimator}
\newacronym{gl:sgd}{SGD}{stochastic gradient descent}

\newacronym{gl:cap}{CAP}{Common Agriculture Policy}
\newacronym{gl:nuts}{NUTS}{Nomenclature of Territorial Units for Statistics}
\newacronym{gl:nrw}{NRW}{North Rhine-Westfalia} 
\newacronym{gl:ls}{LS}{Lower Saxony} 

\newacronym[user1={Schneider2023:EuroCrops,Schneider2021:TinyEuroCrops}]{gl:hcat}{HCAT}{Hierarchical Crop and Agriculture Taxonomy}

\newglossaryentry{gl:DirDist}{%
    name={Dirichlet distribution},%
    description={}%
}


%% file: images/priorshift.tex

\begin{tikzpicture}

\pgfmathdeclarefunction{gauss}{3}{%
  \pgfmathparse{exp(-((#1-#2)^2)/(2*#3^2))}%
}

\begin{axis}[
    axis lines = left,
    width=0.95\linewidth,      
    height=5cm,
    xlabel = {Label $y$},
    ylabel = {Label probability $p(y)$},
    font=\footnotesize\sansmath\sffamily,
    domain=0:15, 
    xmax=15,      
    samples=200,  
    ytick=\empty,
    xtick=\empty,
    enlargelimits=upper,
    legend style={
        at={(0.98,0.98)}, 
        anchor=north east, 
        draw=none,
        fill=none,
        row sep=2pt,
        legend cell align={left}},
    clip=false
]

\addplot [
    fill=cyan, 
    fill opacity=0.4, 
    draw=cyan, 
    thick
] {gauss(x, 5, 1.5)} \closedcycle;
\addlegendentry{Train data}

\addplot [
    fill=purple, 
    fill opacity=0.5, 
    draw=purple, 
    thick
] {gauss(x, 1.2, 0.8) + 0.2*gauss(x, 6, 3)} \closedcycle;
\addlegendentry{Test data}

\draw [stealth-stealth, thick, dashed, gray] (axis cs:2, 1.05) -- (axis cs:4.5, 1.05) node[midway, above, text=black] {$\Delta$ Prior shift};

\end{axis}
\end{tikzpicture}

%% file: tables/at_de_classes_logbin.tex
\definecolor{lavender}{RGB}{234,234,242}
\definecolor{steelblue}{RGB}{76,114,176}
\definecolor{darkblue}{RGB}{34, 65, 120}
\definecolor{midblue}{RGB}{72, 121, 195}
\definecolor{lightblue}{RGB}{160, 195, 235}

\definecolor{peru}{RGB}{221,132,82}
\definecolor{mediumseagreen}{RGB}{85,168,104}

\pgfplotstableread[col sep=comma]{tables/de_ls_classes_per_logbin.csv}\tableDELS
\pgfplotstableread[col sep=comma]{tables/de_nrw_classes_per_logbin.csv}\tableDENERW
\pgfplotstableread[col sep=comma]{tables/austria_classes_per_logbin.csv}\tableAT

\pgfplotsset{
every tick label/.append style={font=\scriptsize\sansmath\sffamily},
every label/.append style={font=\footnotesize\sansmath\sffamily},
BarPlot/.style={
    axis background/.style={fill=lavender},
    axis line style={white},
    font=\sffamily\footnotesize,
    ybar,
    bar width=0.01\linewidth,
    width=\linewidth,
    height=4cm,
    ymin=0,
    ymax=21,
    ymajorgrids,
    ymajorticks=true,
    ytick style={color=white},
    y grid style={white},
    xtick={1,10,100,1000,10000,100000,1000000},
    log base x=10,
    xmin=0.7,
    xmax=2e5,
    xtick pos=left,
    xtick align=outside,
    legend cell align={left},
    legend style={font=\sffamily\scriptsize},
}
}

\begin{tikzpicture}
\begin{axis}[
    BarPlot,
    xmode=log,
    ylabel={\# Crop classes},
    title=\textbf{\scriptsize Germany (LS)},
    title style={at={(0.5,0.92)}, anchor=south},
]
\addplot[draw=darkgray, fill=darkblue, opacity=0.75]
    table [x=bins, y={DE LS}] {\tableDELS};
\end{axis}
\end{tikzpicture}

\begin{tikzpicture}
\begin{axis}[
    BarPlot,
    xmode=log,
    ylabel={\# Crop classes},
    title=\textbf{\scriptsize Germany (NRW)},
    title style={at={(0.5,0.92)}, anchor=south},
]
\addplot[draw=darkgray, fill=midblue, opacity=0.75]
    table [x=bins, y={DE NRW}] {\tableDENERW};
\end{axis}
\end{tikzpicture}

\begin{tikzpicture}
\begin{axis}[
    BarPlot,
    xmode=log,
    ylabel={\# Crop classes},
    xlabel={\# Parcels per crop class (log scale)},
    xlabel style={yshift=-0.1cm},
    title=\textbf{\scriptsize Austria},
    title style={at={(0.5,0.92)}, anchor=south},
]
\addplot[draw=darkgray, fill=lightblue, opacity=0.75]
    table [x=bins, y={AT}] {\tableAT};
\end{axis}
\end{tikzpicture}

%% file: images/classoverlap/AT_DE.tex
\begin{tikzpicture}
\definecolor{mediummidnightblue}{RGB}{72,121,195} 
\definecolor{lightseagreen}{RGB}{62,186,103} 
\scriptsize
\coordinate (AT) at  (0,0);
\coordinate (DE) at (1,0);
\coordinate (cap) at ($(AT)!0.5!(DE)$);

\begin{scope}[blend group=soft light]
    \fill[mediummidnightblue, opacity=0.45] (AT) circle[radius=1.5];
    \fill[mediummidnightblue, opacity=0.45] (DE) circle[radius=1.5];
\end{scope}

\draw (AT) circle[radius=1.5];
\draw (DE) circle[radius=1.5];

\node[font=\normalsize\sffamily] at ($(AT)+(-0.5,-1.5)$) [below,align=center]{Austria};
\node[font=\normalsize\sffamily] at ($(DE)+(0.5,-1.5)$) [below,align=center]{Germany};

\node[font=\normalsize\sffamily] at (AT) [left=0.75cm,align=center]{29}; 
\node[font=\normalsize\sffamily] at (DE) [right=0.75cm,align=center]{129}; 
\node[font=\normalsize\sffamily] at (cap) [align=center]{69}; 

\end{tikzpicture}


%% file: images/classoverlap/AT_DE_BE.tex
\begin{tikzpicture}
\definecolor{mediummidnightblue}{RGB}{72,121,195} 
\definecolor{lightseagreen}{RGB}{62,186,103} 
\scriptsize
\coordinate (AT) at (210:0.75);
\coordinate (DE) at (330:0.75);
\coordinate (BE) at (90:0.75);
\coordinate (capBEATDE) at (0,0);
\coordinate (capBEDE) at (30:1.15);
\coordinate (capBEAT) at (150:1.15);
\coordinate (capATDE) at (270:1.15);

\begin{scope}[blend group=soft light]
    \fill[lightseagreen, opacity=0.45] (BE) circle[radius=1.5];
    \fill[mediummidnightblue, opacity=0.45] (AT) circle[radius=1.5];
    \fill[mediummidnightblue, opacity=0.45] (DE) circle[radius=1.5];
\end{scope}

\draw (AT) circle[radius=1.5];
\draw (DE) circle[radius=1.5];
\draw (BE) circle[radius=1.5];

\node[font=\normalsize\sffamily] at ($(AT)+(-0.5,-1.5)$) [below,align=center]{Austria};
\node[font=\normalsize\sffamily] at ($(BE)+(0,+1.5)$) [above,align=center]{Belgium};
\node[font=\normalsize\sffamily] at ($(DE)+(0.5,-1.5)$) [below,align=center]{Germany};

\node[font=\normalsize\sffamily] at (210:1.65) [align=center]{20}; 
\node[font=\normalsize\sffamily] at (330:1.65) [align=center]{90}; 
\node[font=\normalsize\sffamily] at (90:1.65) [align=center]{38}; 

\node[font=\normalsize\sffamily] at (capBEAT) [align=center]{9}; 
\node[font=\normalsize\sffamily] at (capBEDE) [align=center]{38}; 
\node[font=\normalsize\sffamily] at (capATDE) [align=center]{19}; 

\node[font=\normalsize\sffamily] at (capBEATDE) [align=center]{50}; 

\end{tikzpicture}


%% file: images/classoverlap/AT_DE_EE.tex
\begin{tikzpicture}
\definecolor{mediummidnightblue}{RGB}{72,121,195} 
\definecolor{lightseagreen}{RGB}{62,186,103} 
\scriptsize
\coordinate (AT) at (210:0.75);
\coordinate (DE) at (330:0.75);
\coordinate (EE) at (90:0.75);
\coordinate (capEEATDE) at (0,0);
\coordinate (capEEDE) at (30:1.15);
\coordinate (capEEAT) at (150:1.15);
\coordinate (capATDE) at (270:1.15);

\begin{scope}[blend group=soft light]
    \fill[lightseagreen, opacity=0.45] (EE) circle[radius=1.5];
    \fill[mediummidnightblue, opacity=0.45] (AT) circle[radius=1.5];
    \fill[mediummidnightblue, opacity=0.45] (DE) circle[radius=1.5];
\end{scope}

\draw (AT) circle[radius=1.5];
\draw (DE) circle[radius=1.5];
\draw (EE) circle[radius=1.5];

\node[font=\normalsize\sffamily] at ($(AT)+(-0.5,-1.5)$) [below,align=center]{Austria};
\node[font=\normalsize\sffamily] at ($(EE)+(0,+1.5)$) [above,align=center]{Estonia};
\node[font=\normalsize\sffamily] at ($(DE)+(0.5,-1.5)$) [below,align=center]{Germany};

\node[font=\normalsize\sffamily] at (210:1.65) [align=center]{25}; 
\node[font=\normalsize\sffamily] at (330:1.65) [align=center]{89}; 
\node[font=\normalsize\sffamily] at (90:1.65) [align=center]{35}; 

\node[font=\normalsize\sffamily] at (capEEAT) [align=center]{4}; 
\node[font=\normalsize\sffamily] at (capEEDE) [align=center]{40}; 
\node[font=\normalsize\sffamily] at (capATDE) [align=center]{21}; 

\node[font=\normalsize\sffamily] at (capEEATDE) [align=center]{48}; 

\end{tikzpicture}


%% file: images/classoverlap/AT_DE_LV.tex
\begin{tikzpicture}
\definecolor{mediummidnightblue}{RGB}{72,121,195} 
\definecolor{lightseagreen}{RGB}{62,186,103} 
\scriptsize
\coordinate (AT) at (210:0.75);
\coordinate (DE) at (330:0.75);
\coordinate (LV) at (90:0.75);
\coordinate (capLVATDE) at (0,0);
\coordinate (capLVDE) at (30:1.15);
\coordinate (capLVAT) at (150:1.15);
\coordinate (capATDE) at (270:1.15);

\begin{scope}[blend group=soft light]
    \fill[lightseagreen, opacity=0.45] (LV) circle[radius=1.5];
    \fill[mediummidnightblue, opacity=0.45] (AT) circle[radius=1.5];
    \fill[mediummidnightblue, opacity=0.45] (DE) circle[radius=1.5];
\end{scope}

\draw (AT) circle[radius=1.5];
\draw (DE) circle[radius=1.5];
\draw (LV) circle[radius=1.5];

\node[font=\normalsize\sffamily] at ($(AT)+(-0.5,-1.5)$) [below,align=center]{Austria};
\node[font=\normalsize\sffamily] at ($(LV)+(0,+1.5)$) [above,align=center]{Latvia};
\node[font=\normalsize\sffamily] at ($(DE)+(0.5,-1.5)$) [below,align=center]{Germany};

\node[font=\normalsize\sffamily] at (210:1.65) [align=center]{24}; 
\node[font=\normalsize\sffamily] at (330:1.65) [align=center]{99}; 
\node[font=\normalsize\sffamily] at (90:1.65) [align=center]{29}; 

\node[font=\normalsize\sffamily] at (capLVAT) [align=center]{5}; 
\node[font=\normalsize\sffamily] at (capLVDE) [align=center]{30}; 
\node[font=\normalsize\sffamily] at (capATDE) [align=center]{25}; 

\node[font=\normalsize\sffamily] at (capLVATDE) [align=center]{44}; 

\end{tikzpicture}


%% file: images/classoverlap/AT_DE_LT.tex
\begin{tikzpicture}
\definecolor{mediummidnightblue}{RGB}{72,121,195} 
\definecolor{lightseagreen}{RGB}{62,186,103} 
\scriptsize
\coordinate (AT) at (210:0.75);
\coordinate (DE) at (330:0.75);
\coordinate (LT) at (90:0.75);
\coordinate (capLTATDE) at (0,0);
\coordinate (capLTDE) at (30:1.15);
\coordinate (capLTAT) at (150:1.15);
\coordinate (capATDE) at (270:1.15);

\begin{scope}[blend group=soft light]
    \fill[lightseagreen, opacity=0.45] (LT) circle[radius=1.5];
    \fill[mediummidnightblue, opacity=0.45] (AT) circle[radius=1.5];
    \fill[mediummidnightblue, opacity=0.45] (DE) circle[radius=1.5];
\end{scope}

\draw (AT) circle[radius=1.5];
\draw (DE) circle[radius=1.5];
\draw (LT) circle[radius=1.5];

\node[font=\normalsize\sffamily] at ($(AT)+(-0.5,-1.5)$) [below,align=center]{Austria};
\node[font=\normalsize\sffamily] at ($(LT)+(0,+1.5)$) [above,align=center]{Lithuania};
\node[font=\normalsize\sffamily] at ($(DE)+(0.5,-1.5)$) [below,align=center]{Germany};

\node[font=\normalsize\sffamily] at (210:1.65) [align=center]{28}; 
\node[font=\normalsize\sffamily] at (330:1.65) [align=center]{124}; 
\node[font=\normalsize\sffamily] at (90:1.65) [align=center]{4}; 

\node[font=\normalsize\sffamily] at (capLTAT) [align=center]{1}; 
\node[font=\normalsize\sffamily] at (capLTDE) [align=center]{5}; 
\node[font=\normalsize\sffamily] at (capATDE) [align=center]{57}; 

\node[font=\normalsize\sffamily] at (capLTATDE) [align=center]{12}; 

\end{tikzpicture}


%% file: images/classoverlap/AT_DE_PT.tex
\begin{tikzpicture}
\definecolor{mediummidnightblue}{RGB}{72,121,195} 
\definecolor{lightseagreen}{RGB}{62,186,103} 
\scriptsize
\coordinate (AT) at (210:0.75);
\coordinate (DE) at (330:0.75);
\coordinate (PT) at (90:0.75);
\coordinate (capPTATDE) at (0,0);
\coordinate (capPTDE) at (30:1.15);
\coordinate (capPTAT) at (150:1.15);
\coordinate (capATDE) at (270:1.15);

\begin{scope}[blend group=soft light]
    \fill[lightseagreen, opacity=0.45] (PT) circle[radius=1.5];
    \fill[mediummidnightblue, opacity=0.45] (AT) circle[radius=1.5];
    \fill[mediummidnightblue, opacity=0.45] (DE) circle[radius=1.5];
\end{scope}

\draw (AT) circle[radius=1.5];
\draw (DE) circle[radius=1.5];
\draw (PT) circle[radius=1.5];

\node[font=\normalsize\sffamily] at ($(AT)+(-0.5,-1.5)$) [below,align=center]{Austria};
\node[font=\normalsize\sffamily] at ($(PT)+(0,+1.5)$) [above,align=center]{Portugal};
\node[font=\normalsize\sffamily] at ($(DE)+(0.5,-1.5)$) [below,align=center]{Germany};

\node[font=\normalsize\sffamily] at (210:1.65) [align=center]{21}; 
\node[font=\normalsize\sffamily] at (330:1.65) [align=center]{114}; 
\node[font=\normalsize\sffamily] at (90:1.65) [align=center]{31}; 

\node[font=\normalsize\sffamily] at (capPTAT) [align=center]{8}; 
\node[font=\normalsize\sffamily] at (capPTDE) [align=center]{15}; 
\node[font=\normalsize\sffamily] at (capATDE) [align=center]{38}; 

\node[font=\normalsize\sffamily] at (capPTATDE) [align=center]{30}; 

\end{tikzpicture}


%% file: images/classoverlap/AT_DE_SK.tex
\begin{tikzpicture}
\definecolor{mediummidnightblue}{RGB}{72,121,195} 
\definecolor{lightseagreen}{RGB}{62,186,103} 
\scriptsize
\coordinate (AT) at (210:0.75);
\coordinate (DE) at (330:0.75);
\coordinate (SK) at (90:0.75);
\coordinate (capSKATDE) at (0,0);
\coordinate (capSKDE) at (30:1.15);
\coordinate (capSKAT) at (150:1.15);
\coordinate (capATDE) at (270:1.15);

\begin{scope}[blend group=soft light]
    \fill[lightseagreen, opacity=0.45] (SK) circle[radius=1.5];
    \fill[mediummidnightblue, opacity=0.45] (AT) circle[radius=1.5];
    \fill[mediummidnightblue, opacity=0.45] (DE) circle[radius=1.5];
\end{scope}

\draw (AT) circle[radius=1.5];
\draw (DE) circle[radius=1.5];
\draw (SK) circle[radius=1.5];

\node[font=\normalsize\sffamily] at ($(AT)+(-0.5,-1.5)$) [below,align=center]{Austria};
\node[font=\normalsize\sffamily] at ($(SK)+(0,+1.5)$) [above,align=center]{Slovakia};
\node[font=\normalsize\sffamily] at ($(DE)+(0.5,-1.5)$) [below,align=center]{Germany};

\node[font=\normalsize\sffamily] at (210:1.65) [align=center]{18}; 
\node[font=\normalsize\sffamily] at (330:1.65) [align=center]{100}; 
\node[font=\normalsize\sffamily] at (90:1.65) [align=center]{33}; 

\node[font=\normalsize\sffamily] at (capSKAT) [align=center]{11}; 
\node[font=\normalsize\sffamily] at (capSKDE) [align=center]{29}; 
\node[font=\normalsize\sffamily] at (capATDE) [align=center]{21}; 

\node[font=\normalsize\sffamily] at (capSKATDE) [align=center]{48}; 

\end{tikzpicture}


%% file: images/classoverlap/AT_DE_SI.tex
\begin{tikzpicture}
\definecolor{mediummidnightblue}{RGB}{72,121,195} 
\definecolor{lightseagreen}{RGB}{62,186,103} 
\scriptsize
\coordinate (AT) at (210:0.75);
\coordinate (DE) at (330:0.75);
\coordinate (SI) at (90:0.75);
\coordinate (capSIATDE) at (0,0);
\coordinate (capSIDE) at (30:1.15);
\coordinate (capSIAT) at (150:1.15);
\coordinate (capATDE) at (270:1.15);

\begin{scope}[blend group=soft light]
    \fill[lightseagreen, opacity=0.45] (SI) circle[radius=1.5];
    \fill[mediummidnightblue, opacity=0.45] (AT) circle[radius=1.5];
    \fill[mediummidnightblue, opacity=0.45] (DE) circle[radius=1.5];
\end{scope}

\draw (AT) circle[radius=1.5];
\draw (DE) circle[radius=1.5];
\draw (SI) circle[radius=1.5];

\node[font=\normalsize\sffamily] at ($(AT)+(-0.5,-1.5)$) [below,align=center]{Austria};
\node[font=\normalsize\sffamily] at ($(SI)+(0,+1.5)$) [above,align=center]{Slovenia};
\node[font=\normalsize\sffamily] at ($(DE)+(0.5,-1.5)$) [below,align=center]{Germany};

\node[font=\normalsize\sffamily] at (210:1.65) [align=center]{17}; 
\node[font=\normalsize\sffamily] at (330:1.65) [align=center]{104}; 
\node[font=\normalsize\sffamily] at (90:1.65) [align=center]{28}; 

\node[font=\normalsize\sffamily] at (capSIAT) [align=center]{12}; 
\node[font=\normalsize\sffamily] at (capSIDE) [align=center]{25}; 
\node[font=\normalsize\sffamily] at (capATDE) [align=center]{22}; 

\node[font=\normalsize\sffamily] at (capSIATDE) [align=center]{47}; 

\end{tikzpicture}


%% file: images/classoverlap/AT_DE_ES.tex
\begin{tikzpicture}
\definecolor{mediummidnightblue}{RGB}{72,121,195} 
\definecolor{lightseagreen}{RGB}{62,186,103} 
\scriptsize
\coordinate (AT) at (210:0.75);
\coordinate (DE) at (330:0.75);
\coordinate (ES) at (90:0.75);
\coordinate (capESATDE) at (0,0);
\coordinate (capESDE) at (30:1.15);
\coordinate (capESAT) at (150:1.15);
\coordinate (capATDE) at (270:1.15);

\begin{scope}[blend group=soft light]
    \fill[lightseagreen, opacity=0.45] (ES) circle[radius=1.5];
    \fill[mediummidnightblue, opacity=0.45] (AT) circle[radius=1.5];
    \fill[mediummidnightblue, opacity=0.45] (DE) circle[radius=1.5];
\end{scope}

\draw (AT) circle[radius=1.5];
\draw (DE) circle[radius=1.5];
\draw (ES) circle[radius=1.5];

\node[font=\normalsize\sffamily] at ($(AT)+(-0.5,-1.5)$) [below,align=center]{Austria};
\node[font=\normalsize\sffamily] at ($(ES)+(0,+1.5)$) [above,align=center]{Spain};
\node[font=\normalsize\sffamily] at ($(DE)+(0.5,-1.5)$) [below,align=center]{Germany};

\node[font=\normalsize\sffamily] at (210:1.65) [align=center]{28}; 
\node[font=\normalsize\sffamily] at (330:1.65) [align=center]{129}; 
\node[font=\normalsize\sffamily] at (90:1.65) [align=center]{2}; 

\node[font=\normalsize\sffamily] at (capESAT) [align=center]{1}; 
\node[font=\normalsize\sffamily] at (capESDE) [align=center]{0}; 
\node[font=\normalsize\sffamily] at (capATDE) [align=center]{63}; 

\node[font=\normalsize\sffamily] at (capESATDE) [align=center]{6}; 

\end{tikzpicture}


%% file: images/biogeographical_regions_legend.tex
\definecolor{alpineColor}{RGB}{175, 89, 168}     
\definecolor{anatolianColor}{RGB}{219, 160, 184}       
\definecolor{arcticColor}{RGB}{142, 212, 191}      
\definecolor{atlanticColor}{RGB}{85, 165, 222}    
\definecolor{blackSeaColor}{RGB}{60, 38, 73}   
\definecolor{borealColor}{RGB}{63, 93, 159}    
\definecolor{continentalColor}{RGB}{211, 233, 177}    
\definecolor{macaronesiaColor}{RGB}{157, 114, 237}    
\definecolor{mediterraneanColor}{RGB}{237, 217, 62}    
\definecolor{pannonianColor}{RGB}{194, 108, 80}    
\definecolor{steppicColor}{RGB}{217 199 171}    

\begin{tikzpicture}[
    square/.style={rectangle,draw,fill,minimum size=0.3cm}
]
    \node[square,draw=none,fill=alpineColor] at (-3.5, 0,0) {};
    \node[square,draw=none,fill=anatolianColor] at (-3.5,-0.5) {};
    \node[square,draw=none,fill=arcticColor] at (-3.5,-1.0) {};
    
    \node[anchor=west,font=\sffamily\normalsize] at (-3.2,0) {Alpine};
    \node[anchor=west,font=\sffamily\normalsize] at (-3.2,-0.5) {Anatolian};
    \node[anchor=west,font=\sffamily\normalsize] at (-3.2,-1.0) {Arctic};
    
    \node[square,draw=none,fill=atlanticColor] at (-0.5,0) {};
    \node[square,draw=none,fill=blackSeaColor] at (-0.5,-0.5) {};
    \node[square,draw=none,fill=borealColor] at (-0.5,-1.0) {};
    
    \node[anchor=west,font=\sffamily\normalsize] at (-0.2,0) {Atlantic};
    \node[anchor=west,font=\sffamily\normalsize] at (-0.2,-0.5) {Black Sea};
    \node[anchor=west,font=\sffamily\normalsize] at (-0.2,-1.0) {Boreal};      

    \node[square,draw=none,fill=continentalColor] at (2.5,0) {};
    \node[square,draw=none,fill=pannonianColor] at (2.5,-0.5) {};
    \node[square,draw=none,fill=mediterraneanColor] at (2.5,-1.0) {};
    
    \node[anchor=west,font=\sffamily\normalsize] at (2.8,0) {Continental};
    \node[anchor=west,font=\sffamily\normalsize] at (2.8,-0.5) {Pannonian};
    \node[anchor=west,font=\sffamily\normalsize] at (2.8,-1.0) {Mediterranean};    

    \node[square,draw=none,fill=steppicColor] at (5.5,0) {};
    
    \node[anchor=west,font=\sffamily\normalsize] at (5.8,0) {Steppic};
\end{tikzpicture}

%% file: images/dirichlet/dirichlet_densitycolorbar.tex
\begin{tikzpicture}

\node (img1) at (0,0) {\includegraphics[height=3.5cm, width=0.3cm]{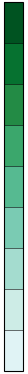}};

\node[anchor=south] at ($(img1.north) + (0, 0.05)$) (upperlimittest) {\large high density};

\node[anchor=north] at ($(img1.south) + (0, -0.05)$) (lowerlimittest) {\large low density};

\end{tikzpicture}

%% file: images/dirichlet/dirichlet_bars.tex
\pgfplotsset{
    dirichlet_grouped/.style={
        ybar=0pt,                
        bar width=0.18,          
        axis lines=left,
        width=\linewidth,              
        height=5.5cm,
        ymin=0, ymax=1.1,
        xmin=0.5, xmax=5.5,      
        enlarge x limits={abs=0.4}, 
        clip=false,              
        xtick={1,2,3,4,5},
        xticklabels={$c_1$, $c_2$, $c_3$, $c_4$, $c_5$},
        ytick={0, 0.2, 0.4, 0.6, 0.8, 1.0},
        tick label style={font=\footnotesize\sansmath\sffamily},
        label style={font=\footnotesize\sansmath\sffamily},
        tick style={thin, black!60},
        axis line style={black!80, thick},
        legend style={
            font=\footnotesize\sansmath\sffamily, 
            at={(1.0,0.9)},      
            anchor=north east,         
            draw=none,          
            fill=none,            
            legend columns=2,
            /tikz/every even column/.append style={column sep=15pt} 
        },
        ymajorgrids=true,
        grid style={dashed, gray!30},
        title style={
            at={(0.5,1)}, 
            anchor=south, 
            font=\footnotesize\sansmath\bfseries\sffamily, 
            yshift=-10pt
        }
    }
}

\begin{tikzpicture}[font=\sffamily]

\pgfmathsetmacro{\unif}{0.2}
\def\rowsep{5.25cm} 
\begin{axis}[
    dirichlet_grouped,
    at={(0,0)},
    ylabel={Pseudo-prior $\tilde{\pi}_c$},
    title={Sparse Dirichlet prior ($\alpha = 0.5$)}
]
    \addplot [fill=orange!90, draw=black!80, thin] coordinates {(1,0.117)(2,0.548)(3,0.020)(4,0.003)(5,0.312)};
    \addplot [fill=orange!65, draw=black!80, thin] coordinates {(1,0.000)(2,0.778)(3,0.023)(4,0.066)(5,0.133)};
    \addplot [fill=orange!45, draw=black!80, thin] coordinates {(1,0.353)(2,0.077)(3,0.187)(4,0.036)(5,0.347)};
    \addplot [fill=orange!25, draw=black!80, thin] coordinates {(1,0.001)(2,0.840)(3,0.023)(4,0.132)(5,0.004)};
        
    \legend{Iteration 1, Iteration 2, Iteration 3, Iteration 4}
    
    \draw[gray!80!black, thick, dashdotted] (axis cs:0.5,\unif) -- (axis cs:5.5,\unif) 
        node[right, font=\footnotesize\sffamily, color=gray!80!black, xshift=2pt] {Uniform};
\end{axis}

\begin{axis}[
    dirichlet_grouped,
    at={(0,-\rowsep)},
    ylabel={Pseudo-prior $\tilde{\pi}_c$},
    xlabel={Class label $y$},
    title={Dense Dirichlet prior ($\alpha = 5.0$)}
]
    \addplot [fill=teal!90, draw=black!80, thin] coordinates {(1,0.289)(2,0.108)(3,0.102)(4,0.221)(5,0.280)};
    \addplot [fill=teal!65, draw=black!80, thin] coordinates {(1,0.205)(2,0.180)(3,0.164)(4,0.152)(5,0.299)};
    \addplot [fill=teal!45, draw=black!80, thin] coordinates {(1,0.208)(2,0.149)(3,0.129)(4,0.235)(5,0.279)};
    \addplot [fill=teal!25, draw=black!80, thin] coordinates {(1,0.316)(2,0.139)(3,0.183)(4,0.168)(5,0.194)};

    \legend{Iteration 1, Iteration 2, Iteration 3, Iteration 4}
    
    \draw[gray!80!black, thick, dashdotted] (axis cs:0.5,\unif) -- (axis cs:5.5,\unif)
        node[right, font=\footnotesize\sffamily, color=gray!80!black, xshift=2pt] {Uniform};
\end{axis}

\end{tikzpicture}

%% file: tables/results_revision/BE.tex
\robustify\bfseries
\sisetup{detect-all=true,uncertainty-mode=separate,table-align-uncertainty=true,round-mode=uncertainty,round-precision=3}

\begin{tabular}{llS[table-format=1.3(3)]S[table-format=1.3(3)]S[table-format=1.3(3)]S[table-format=1.3(3)]S[table-format=1.3(3)]}
\toprule
& Algorithm/Loss & \multicolumn{5}{l@{}}{Benchmark task ($k$-shot)} \\
\cmidrule(l){3-7}
{} & {} & {1} & {5} & {10} & {20} & {100} \\
\cmidrule(lr){1-2} \cmidrule(lr){3-3} \cmidrule(lr){4-4} \cmidrule(lr){5-5} \cmidrule(lr){6-6} \cmidrule(l){7-7}

\multirow[c]{7}{*}{\rotatebox[origin=c]{90}{Kappa}} & \gls{gl:ce} & 0.170 +- 0.011 & 0.278 +- 0.029 & 0.403 +- 0.040 & 0.462 +- 0.015 & 0.545 +- 0.012 \\
 & \gls{gl:fl} & 0.156 +- 0.020 & \bfseries 0.268 +- 0.020 & 0.378 +- 0.030 & 0.435 +- 0.023 & 0.508 +- 0.025 \\
 & \gls{gl:ce} \gls{gl:ros} & 0.170 +- 0.011 & 0.263 +- 0.036 & 0.418 +- 0.043 & 0.407 +- 0.032 & 0.482 +- 0.018 \\
 & \gls{gl:ce} \gls{gl:dirpa} & \colorblue \bfseries 0.221 +- 0.010 & \colorblue \bfseries 0.371 +- 0.039 & \colorblue \bfseries 0.492 +- 0.052 & \colorblue \bfseries 0.530 +- 0.040 & \colorblue \bfseries 0.597 +- 0.027 \\
 & \gls{gl:fl} \gls{gl:dirpa} & \bfseries 0.176 +- 0.036 & 0.248 +- 0.053 & \bfseries 0.405 +- 0.028 & \bfseries 0.439 +- 0.063 & \bfseries 0.536 +- 0.022 \\
 \arrayrulecolor{gray} 
 \cmidrule(lr){2-7}
 \arrayrulecolor{black}
& \gls{gl:ce} \gls{gl:presto} & 0.046 +- 0.016 & 0.106 +- 0.014 & 0.205 +- 0.041 & \colorteal \bfseries 0.273 +- 0.016 & \colorteal \bfseries 0.379 +- 0.034 \\
 & \gls{gl:ce} \gls{gl:presto} \gls{gl:dirpa} & \colorteal \bfseries 0.054 +- 0.035 & \colorteal \bfseries 0.159 +- 0.045 & \colorteal \bfseries 0.233 +- 0.087 & 0.265 +- 0.032 & 0.359 +- 0.041 \\
\midrule
\multirow[c]{7}{*}{\rotatebox[origin=c]{90}{Accuracy}} & \gls{gl:ce} & 0.185 +- 0.013 & 0.333 +- 0.053 & 0.480 +- 0.058 & 0.546 +- 0.019 & 0.627 +- 0.014 \\
 & \gls{gl:fl} & 0.170 +- 0.022 & \bfseries 0.325 +- 0.041 & 0.462 +- 0.041 & 0.513 +- 0.027 & 0.594 +- 0.023 \\
 & \gls{gl:ce} \gls{gl:ros} & 0.185 +- 0.013 & 0.305 +- 0.050 & 0.516 +- 0.064 & 0.483 +- 0.038 & 0.560 +- 0.017 \\
 & \gls{gl:ce} \gls{gl:dirpa} & \colorblue \bfseries 0.254 +- 0.040 & \colorblue \bfseries 0.469 +- 0.044 & \colorblue \bfseries 0.589 +- 0.061 & \colorblue \bfseries 0.625 +- 0.049 & \colorblue \bfseries 0.687 +- 0.025 \\
 & \gls{gl:fl} \gls{gl:dirpa} & \bfseries 0.194 +- 0.039 & 0.282 +- 0.075 & \bfseries 0.478 +- 0.038 & \bfseries 0.520 +- 0.064 & \bfseries 0.619 +- 0.022 \\
 \arrayrulecolor{gray} 
 \cmidrule(lr){2-7}
 \arrayrulecolor{black}
& \gls{gl:ce} \gls{gl:presto} & 0.055 +- 0.020 & 0.151 +- 0.063 & 0.315 +- 0.095 & 0.354 +- 0.018 & \colorteal \bfseries 0.464 +- 0.041 \\
 & \gls{gl:ce} \gls{gl:presto} \gls{gl:dirpa} & \colorteal \bfseries 0.111 +- 0.099 & \colorteal \bfseries 0.361 +- 0.155 & \colorteal \bfseries 0.355 +- 0.155 & \colorteal \bfseries 0.372 +- 0.035 & 0.445 +- 0.052 \\
\midrule
\multirow[c]{7}{*}{\rotatebox[origin=c]{90}{F1-score}} & \gls{gl:ce} & 0.084 +- 0.007 & 0.124 +- 0.018 & \colorblue \bfseries 0.149 +- 0.010 & \bfseries 0.179 +- 0.017 & \colorblue \bfseries 0.236 +- 0.010 \\
 & \gls{gl:fl} & \bfseries 0.085 +- 0.007 & \bfseries 0.121 +- 0.017 & 0.140 +- 0.016 & \colorblue \bfseries 0.182 +- 0.010 & 0.221 +- 0.024 \\
 & \gls{gl:ce} \gls{gl:ros} & 0.084 +- 0.007 & 0.117 +- 0.020 & 0.128 +- 0.025 & 0.171 +- 0.010 & 0.220 +- 0.006 \\
 & \gls{gl:ce} \gls{gl:dirpa} & \colorblue \bfseries 0.087 +- 0.006 & \colorblue \bfseries 0.128 +- 0.015 & 0.144 +- 0.017 & 0.161 +- 0.025 & 0.215 +- 0.029 \\
 & \gls{gl:fl} \gls{gl:dirpa} & 0.081 +- 0.007 & 0.121 +- 0.023 & \bfseries 0.146 +- 0.012 & 0.175 +- 0.018 & \bfseries 0.232 +- 0.019 \\
 \arrayrulecolor{gray} 
 \cmidrule(lr){2-7}
 \arrayrulecolor{black}
& \gls{gl:ce} \gls{gl:presto} & \colorteal \bfseries 0.015 +- 0.006 & \colorteal \bfseries 0.036 +- 0.014 & \colorteal \bfseries 0.048 +- 0.018 & \colorteal \bfseries 0.083 +- 0.009 & \colorteal \bfseries 0.132 +- 0.018 \\
 & \gls{gl:ce} \gls{gl:presto} \gls{gl:dirpa} & 0.010 +- 0.005 & 0.015 +- 0.016 & 0.038 +- 0.021 & 0.068 +- 0.008 & 0.123 +- 0.014 \\
\bottomrule
\end{tabular}

%% file: tables/results_revision/parentlv3/BE.tex
\robustify\bfseries
\sisetup{detect-all=true,uncertainty-mode=separate,table-align-uncertainty=true,round-mode=uncertainty,round-precision=3}

\begin{tabular}{llS[table-format=1.3(3)]S[table-format=1.3(3)]S[table-format=1.3(3)]S[table-format=1.3(3)]S[table-format=1.3(3)]}
\toprule
& Algorithm/Loss & \multicolumn{5}{l@{}}{Benchmark task ($k$-shot)} \\
\cmidrule(l){3-7}
{} & {} & {1} & {5} & {10} & {20} & {100} \\
\cmidrule(lr){1-2} \cmidrule(lr){3-3} \cmidrule(lr){4-4} \cmidrule(lr){5-5} \cmidrule(lr){6-6} \cmidrule(l){7-7}

\multirow[c]{7}{*}{\rotatebox[origin=c]{90}{Accuracy}} & \gls{gl:ce} & 0.397 +- 0.015 & 0.488 +- 0.072 & 0.609 +- 0.058 & 0.664 +- 0.021 & 0.720 +- 0.015 \\
 & \gls{gl:fl} & 0.390 +- 0.004 & \bfseries 0.492 +- 0.060 & \bfseries 0.613 +- 0.041 & 0.631 +- 0.024 & 0.700 +- 0.018 \\
 & \gls{gl:ce} \gls{gl:ros} & 0.397 +- 0.015 & 0.455 +- 0.038 & 0.675 +- 0.074 & 0.615 +- 0.036 & 0.665 +- 0.009 \\
 & \gls{gl:ce} \gls{gl:dirpa} & \colorblue \bfseries 0.422 +- 0.074 & \colorblue \bfseries 0.618 +- 0.047 & \colorblue \bfseries 0.710 +- 0.052 & \colorblue \bfseries 0.738 +- 0.050 & \colorblue \bfseries 0.785 +- 0.023 \\
 & \gls{gl:fl} \gls{gl:dirpa} & \bfseries 0.392 +- 0.007 & 0.436 +- 0.074 & 0.592 +- 0.041 & \bfseries 0.648 +- 0.041 & \bfseries 0.720 +- 0.020 \\
 \arrayrulecolor{gray} 
 \cmidrule(lr){2-7}
 \arrayrulecolor{black}
& \gls{gl:ce} \gls{gl:presto} & 0.350 +- 0.021 & 0.421 +- 0.080 & 0.574 +- 0.129 & 0.568 +- 0.025 & \colorteal \bfseries 0.620 +- 0.042 \\
 & \gls{gl:ce} \gls{gl:presto} \gls{gl:dirpa} & \colorteal \bfseries 0.417 +- 0.073 & \colorteal \bfseries 0.583 +- 0.127 & \colorteal \bfseries 0.602 +- 0.169 & \colorteal \bfseries 0.625 +- 0.018 & 0.618 +- 0.040 \\
\midrule
\multirow[c]{7}{*}{\rotatebox[origin=c]{90}{F1-score}} & \gls{gl:ce} & \colorblue \bfseries 0.167 +- 0.018 & 0.202 +- 0.034 & 0.280 +- 0.006 & 0.341 +- 0.025 & \colorblue \bfseries 0.458 +- 0.037 \\
 & \gls{gl:fl} & \bfseries 0.157 +- 0.012 & \bfseries 0.205 +- 0.038 & \bfseries 0.278 +- 0.015 & \bfseries 0.335 +- 0.048 & 0.403 +- 0.070 \\
 & \gls{gl:ce} \gls{gl:ros} & \colorblue \bfseries 0.167 +- 0.018 & 0.184 +- 0.027 & \colorblue \bfseries 0.306 +- 0.025 & 0.311 +- 0.012 & 0.370 +- 0.006 \\
 & \gls{gl:ce} \gls{gl:dirpa} & 0.161 +- 0.038 & \colorblue \bfseries 0.242 +- 0.039 & 0.297 +- 0.022 & \colorblue \bfseries 0.351 +- 0.018 & 0.422 +- 0.036 \\
 & \gls{gl:fl} \gls{gl:dirpa} & 0.155 +- 0.015 & 0.173 +- 0.055 & 0.274 +- 0.014 & 0.327 +- 0.033 & \bfseries 0.435 +- 0.049 \\
 \arrayrulecolor{gray} 
 \cmidrule(lr){2-7}
 \arrayrulecolor{black}
& \gls{gl:ce} \gls{gl:presto} & 0.111 +- 0.017 & 0.144 +- 0.020 & \colorteal \bfseries 0.242 +- 0.106 & 0.273 +- 0.023 & \colorteal \bfseries 0.326 +- 0.036 \\
 & \gls{gl:ce} \gls{gl:presto} \gls{gl:dirpa} & \colorteal \bfseries 0.129 +- 0.019 & \colorteal \bfseries 0.157 +- 0.024 & 0.211 +- 0.031 & \colorteal \bfseries 0.299 +- 0.023 & 0.323 +- 0.014 \\
\bottomrule
\end{tabular}

%% file: tables/results_revision/EE.tex
\robustify\bfseries
\sisetup{detect-all=true,uncertainty-mode=separate,table-align-uncertainty=true,round-mode=uncertainty,round-precision=3}


\begin{tabular}{llS[table-format=1.3(3)]S[table-format=1.3(3)]S[table-format=1.3(3)]S[table-format=1.3(3)]S[table-format=1.3(3)]}
\toprule
& Algorithm/Loss & \multicolumn{5}{l@{}}{Benchmark task ($k$-shot)} \\
\cmidrule(l){3-7}
{} & {} & {1} & {5} & {10} & {20} & {100} \\
\cmidrule(lr){1-2} \cmidrule(lr){3-3} \cmidrule(lr){4-4} \cmidrule(lr){5-5} \cmidrule(lr){6-6} \cmidrule(l){7-7}

\multirow[c]{7}{*}{\rotatebox[origin=c]{90}{Kappa}} & \gls{gl:ce} & 0.200 +- 0.035 & 0.275 +- 0.013 & 0.334 +- 0.017 & 0.383 +- 0.016 & 0.487 +- 0.018 \\
 & \gls{gl:fl} & \bfseries 0.198 +- 0.029 & 0.241 +- 0.021 & 0.308 +- 0.028 & 0.372 +- 0.035 & 0.473 +- 0.026 \\
 & \gls{gl:ce} \gls{gl:ros} & 0.200 +- 0.035 & 0.235 +- 0.019 & 0.292 +- 0.030 & 0.348 +- 0.033 & 0.427 +- 0.020 \\
 & \gls{gl:ce} \gls{gl:dirpa} & \colorblue \bfseries 0.306 +- 0.060 & \colorblue \bfseries 0.401 +- 0.037 & \colorblue \bfseries 0.448 +- 0.032 & \colorblue \bfseries 0.479 +- 0.025 & \colorblue \bfseries 0.532 +- 0.008 \\
 & \gls{gl:fl} \gls{gl:dirpa} & 0.188 +- 0.059 & \bfseries 0.295 +- 0.035 & \bfseries 0.349 +- 0.089 & \bfseries 0.379 +- 0.040 & \bfseries 0.484 +- 0.029 \\
 \arrayrulecolor{gray} 
 \cmidrule(lr){2-7}
 \arrayrulecolor{black}
& \gls{gl:ce} \gls{gl:presto} & 0.052 +- 0.016 & 0.109 +- 0.024 & 0.139 +- 0.031 & 0.160 +- 0.016 & \colorteal \bfseries 0.282 +- 0.033 \\
 & \gls{gl:ce} \gls{gl:presto} \gls{gl:dirpa} & \colorteal \bfseries 0.098 +- 0.036 & \colorteal \bfseries 0.144 +- 0.050 & \colorteal \bfseries 0.181 +- 0.049 & \colorteal \bfseries 0.182 +- 0.052 & 0.276 +- 0.030 \\
\midrule
\multirow[c]{7}{*}{\rotatebox[origin=c]{90}{Accuracy}} & \gls{gl:ce} & 0.268 +- 0.061 & 0.350 +- 0.023 & 0.420 +- 0.030 & 0.479 +- 0.034 & 0.575 +- 0.020 \\
 & \gls{gl:fl} & \bfseries 0.268 +- 0.049 & 0.311 +- 0.034 & 0.384 +- 0.040 & 0.465 +- 0.053 & 0.565 +- 0.029 \\
 & \gls{gl:ce} \gls{gl:ros} & 0.268 +- 0.061 & 0.291 +- 0.041 & 0.361 +- 0.045 & 0.429 +- 0.047 & 0.512 +- 0.022 \\
 & \gls{gl:ce} \gls{gl:dirpa} & \colorblue \bfseries 0.436 +- 0.090 & \colorblue \bfseries 0.529 +- 0.044 & \colorblue \bfseries 0.570 +- 0.047 & \colorblue \bfseries 0.601 +- 0.025 & \colorblue \bfseries 0.640 +- 0.026 \\
 & \gls{gl:fl} \gls{gl:dirpa} & 0.252 +- 0.111 & \bfseries 0.380 +- 0.060 & \bfseries 0.441 +- 0.137 & \bfseries 0.483 +- 0.071 & \bfseries 0.580 +- 0.035 \\
 \arrayrulecolor{gray} 
 \cmidrule(lr){2-7}
 \arrayrulecolor{black}
& \gls{gl:ce} \gls{gl:presto} & 0.095 +- 0.062 & 0.197 +- 0.080 & 0.222 +- 0.058 & 0.247 +- 0.039 & \colorteal \bfseries 0.399 +- 0.068 \\
 & \gls{gl:ce} \gls{gl:presto} \gls{gl:dirpa} & \colorteal \bfseries 0.283 +- 0.163 & \colorteal \bfseries 0.295 +- 0.122 & \colorteal \bfseries 0.343 +- 0.116 & \colorteal \bfseries 0.298 +- 0.125 & 0.390 +- 0.040 \\
\midrule
\multirow[c]{7}{*}{\rotatebox[origin=c]{90}{F1-score}} & \gls{gl:ce} & 0.049 +- 0.005 & \colorblue \bfseries 0.094 +- 0.005 & \colorblue \bfseries 0.115 +- 0.005 & \colorblue \bfseries 0.117 +- 0.024 & \colorblue \bfseries 0.174 +- 0.009 \\
 & \gls{gl:fl} & 0.048 +- 0.004 & 0.086 +- 0.009 & \bfseries 0.110 +- 0.013 & \bfseries 0.112 +- 0.013 & \bfseries 0.155 +- 0.018 \\
 & \gls{gl:ce} \gls{gl:ros} & 0.049 +- 0.005 & 0.083 +- 0.007 & 0.095 +- 0.007 & 0.115 +- 0.011 & 0.154 +- 0.004 \\
 & \gls{gl:ce} \gls{gl:dirpa} & \colorblue \bfseries 0.050 +- 0.007 & 0.078 +- 0.019 & 0.101 +- 0.016 & 0.095 +- 0.004 & 0.127 +- 0.029 \\
 & \gls{gl:fl} \gls{gl:dirpa} & \bfseries 0.049 +- 0.003 & \bfseries 0.089 +- 0.014 & 0.094 +- 0.015 & 0.099 +- 0.027 & 0.140 +- 0.036 \\
 \arrayrulecolor{gray} 
 \cmidrule(lr){2-7}
 \arrayrulecolor{black}
& \gls{gl:ce} \gls{gl:presto} & \colorteal \bfseries 0.011 +- 0.004 & \colorteal \bfseries 0.020 +- 0.008 & \colorteal \bfseries 0.026 +- 0.003 & \colorteal \bfseries 0.038 +- 0.017 & \colorteal \bfseries 0.064 +- 0.016 \\
 & \gls{gl:ce} \gls{gl:presto} \gls{gl:dirpa} & 0.009 +- 0.003 & 0.018 +- 0.007 & 0.019 +- 0.007 & 0.030 +- 0.006 & 0.064 +- 0.020 \\
\bottomrule
\end{tabular}

%% file: tables/results_revision/parentlv3/EE.tex
\robustify\bfseries
\sisetup{detect-all=true,uncertainty-mode=separate,table-align-uncertainty=true,round-mode=uncertainty,round-precision=3}

\begin{tabular}{llS[table-format=1.3(3)]S[table-format=1.3(3)]S[table-format=1.3(3)]S[table-format=1.3(3)]S[table-format=1.3(3)]}
\toprule
& Algorithm/Loss & \multicolumn{5}{l@{}}{Benchmark task ($k$-shot)} \\
\cmidrule(l){3-7}
{} & {} & {1} & {5} & {10} & {20} & {100} \\
\cmidrule(lr){1-2} \cmidrule(lr){3-3} \cmidrule(lr){4-4} \cmidrule(lr){5-5} \cmidrule(lr){6-6} \cmidrule(l){7-7}

\multirow[c]{7}{*}{\rotatebox[origin=c]{90}{Accuracy}} & \gls{gl:ce} & 0.577 +- 0.058 & 0.623 +- 0.029 & 0.669 +- 0.042 & 0.706 +- 0.039 & 0.760 +- 0.018 \\
 & \gls{gl:fl} & \bfseries 0.579 +- 0.054 & 0.599 +- 0.045 & 0.637 +- 0.036 & 0.693 +- 0.049 & 0.757 +- 0.028 \\
 & \gls{gl:ce} \gls{gl:ros} & 0.577 +- 0.058 & 0.571 +- 0.053 & 0.616 +- 0.042 & 0.667 +- 0.046 & 0.711 +- 0.024 \\
 & \gls{gl:ce} \gls{gl:dirpa} & \colorblue \bfseries 0.730 +- 0.066 & \colorblue \bfseries 0.757 +- 0.041 & \colorblue \bfseries 0.785 +- 0.034 & \colorblue \bfseries 0.812 +- 0.032 & \colorblue \bfseries 0.816 +- 0.021 \\
 & \gls{gl:fl} \gls{gl:dirpa} & 0.567 +- 0.115 & \bfseries 0.643 +- 0.047 & \bfseries 0.676 +- 0.127 & \bfseries 0.717 +- 0.079 & \bfseries 0.776 +- 0.039 \\
 \arrayrulecolor{gray} 
 \cmidrule(lr){2-7}
 \arrayrulecolor{black}
& \gls{gl:ce} \gls{gl:presto} & 0.458 +- 0.050 & 0.550 +- 0.087 & 0.577 +- 0.036 & 0.581 +- 0.046 & 0.685 +- 0.077 \\
 & \gls{gl:ce} \gls{gl:presto} \gls{gl:dirpa} & \colorteal \bfseries 0.579 +- 0.085 & \colorteal \bfseries 0.612 +- 0.076 & \colorteal \bfseries 0.668 +- 0.096 & \colorteal \bfseries 0.605 +- 0.100 & \colorteal \bfseries 0.698 +- 0.045 \\
\midrule
\multirow[c]{7}{*}{\rotatebox[origin=c]{90}{F1-score}} & \gls{gl:ce} & 0.209 +- 0.026 & 0.233 +- 0.014 & 0.261 +- 0.017 & 0.272 +- 0.014 & 0.307 +- 0.007 \\
 & \gls{gl:fl} & \bfseries 0.213 +- 0.027 & 0.219 +- 0.022 & 0.243 +- 0.013 & 0.266 +- 0.021 & 0.303 +- 0.006 \\
 & \gls{gl:ce} \gls{gl:ros} & 0.209 +- 0.026 & 0.202 +- 0.028 & 0.230 +- 0.023 & 0.254 +- 0.021 & 0.285 +- 0.009 \\
 & \gls{gl:ce} \gls{gl:dirpa} & \colorblue \bfseries 0.273 +- 0.022 & \colorblue \bfseries 0.287 +- 0.012 & \colorblue \bfseries 0.299 +- 0.014 & \colorblue \bfseries 0.311 +- 0.011 & \colorblue \bfseries 0.313 +- 0.016 \\
 & \gls{gl:fl} \gls{gl:dirpa} & 0.197 +- 0.061 & \bfseries 0.242 +- 0.021 & \bfseries 0.253 +- 0.063 & \bfseries 0.271 +- 0.028 & \bfseries 0.308 +- 0.011 \\
 \arrayrulecolor{gray} 
 \cmidrule(lr){2-7}
 \arrayrulecolor{black}
& \gls{gl:ce} \gls{gl:presto} & 0.152 +- 0.034 & 0.195 +- 0.025 & 0.203 +- 0.020 & 0.213 +- 0.021 & 0.255 +- 0.028 \\
 & \gls{gl:ce} \gls{gl:presto} \gls{gl:dirpa} & \colorteal \bfseries 0.194 +- 0.036 & \colorteal \bfseries 0.220 +- 0.016 & \colorteal \bfseries 0.234 +- 0.027 & \colorteal \bfseries 0.217 +- 0.046 & \colorteal \bfseries 0.256 +- 0.015 \\
\bottomrule
\end{tabular}

%% file: tables/results_revision/LV.tex
\robustify\bfseries
\sisetup{detect-all=true,uncertainty-mode=separate,table-align-uncertainty=true,round-mode=uncertainty,round-precision=3}



\begin{tabular}{llS[table-format=1.3(3)]S[table-format=1.3(3)]S[table-format=1.3(3)]S[table-format=1.3(3)]S[table-format=1.3(3)]}
\toprule
& Algorithm/Loss & \multicolumn{5}{l@{}}{Benchmark task ($k$-shot)} \\
\cmidrule(l){3-7}
{} & {} & {1} & {5} & {10} & {20} & {100} \\
\cmidrule(lr){1-2} \cmidrule(lr){3-3} \cmidrule(lr){4-4} \cmidrule(lr){5-5} \cmidrule(lr){6-6} \cmidrule(l){7-7}

\multirow[c]{5}{*}{\rotatebox[origin=c]{90}{Kappa}} & \gls{gl:ce} & 0.141 +- 0.012 & 0.243 +- 0.031 & 0.251 +- 0.023 & 0.290 +- 0.024 & \bfseries 0.417 +- 0.025 \\
 & \gls{gl:fl} & 0.155 +- 0.022 & 0.244 +- 0.059 & 0.243 +- 0.058 & 0.306 +- 0.023 & 0.397 +- 0.025 \\
 & \gls{gl:ce} \gls{gl:ros} & 0.141 +- 0.012 & 0.203 +- 0.026 & 0.239 +- 0.040 & 0.299 +- 0.038 & 0.367 +- 0.015 \\
 & \gls{gl:ce} \gls{gl:dirpa} & \bfseries 0.143 +- 0.025 & \bfseries 0.243 +- 0.017 & \bfseries 0.263 +- 0.057 & \bfseries 0.306 +- 0.041 & 0.411 +- 0.053 \\
 & \gls{gl:fl} \gls{gl:dirpa} & \colorblue \bfseries 0.204 +- 0.055 & \colorblue \bfseries 0.345 +- 0.077 & \colorblue \bfseries 0.339 +- 0.057 & \colorblue \bfseries 0.384 +- 0.061 & \colorblue \bfseries 0.424 +- 0.029 \\
\midrule
\multirow[c]{5}{*}{\rotatebox[origin=c]{90}{Accuracy}} & \gls{gl:ce} & 0.187 +- 0.031 & \bfseries 0.305 +- 0.052 & 0.312 +- 0.035 & 0.337 +- 0.029 & 0.485 +- 0.028 \\
 & \gls{gl:fl} & 0.214 +- 0.039 & 0.307 +- 0.090 & 0.304 +- 0.088 & 0.367 +- 0.034 & 0.464 +- 0.028 \\
 & \gls{gl:ce} \gls{gl:ros} & 0.187 +- 0.031 & 0.254 +- 0.033 & 0.291 +- 0.058 & 0.357 +- 0.054 & 0.429 +- 0.017 \\
 & \gls{gl:ce} \gls{gl:dirpa} & \bfseries 0.197 +- 0.043 & 0.304 +- 0.027 & \bfseries 0.333 +- 0.092 & \bfseries 0.372 +- 0.067 & \bfseries 0.490 +- 0.071 \\
 & \gls{gl:fl} \gls{gl:dirpa} & \colorblue \bfseries 0.290 +- 0.087 & \colorblue \bfseries 0.442 +- 0.104 & \colorblue \bfseries 0.442 +- 0.075 & \colorblue \bfseries 0.477 +- 0.074 & \colorblue \bfseries 0.503 +- 0.032 \\
\midrule
\multirow[c]{5}{*}{\rotatebox[origin=c]{90}{F1-score}} & \gls{gl:ce} & \colorblue \bfseries 0.048 +- 0.004 & 0.089 +- 0.008 & \colorblue \bfseries 0.100 +- 0.007 & \colorblue \bfseries 0.130 +- 0.004 & \colorblue \bfseries 0.168 +- 0.005 \\
 & \gls{gl:fl} & \bfseries 0.047 +- 0.006 & 0.081 +- 0.004 & \bfseries 0.097 +- 0.009 & \bfseries 0.125 +- 0.023 & \bfseries 0.166 +- 0.007 \\
 & \gls{gl:ce} \gls{gl:ros} &  \colorblue \bfseries 0.048 +- 0.004 & \colorblue \bfseries 0.094 +- 0.003 & 0.099 +- 0.010 & 0.121 +- 0.009 & 0.152 +- 0.011 \\
 & \gls{gl:ce} \gls{gl:dirpa} & 0.047 +- 0.005 & 0.092 +- 0.011 & 0.087 +- 0.014 & 0.116 +- 0.022 & 0.137 +- 0.031 \\
 & \gls{gl:fl} \gls{gl:dirpa} & 0.042 +- 0.009 & \bfseries 0.088 +- 0.009 & 0.088 +- 0.015 & 0.107 +- 0.012 & 0.145 +- 0.015 \\
\bottomrule
\end{tabular}

%% file: tables/results_revision/parentlv3/LV.tex
\robustify\bfseries
\sisetup{detect-all=true,uncertainty-mode=separate,table-align-uncertainty=true,round-mode=uncertainty,round-precision=3}

\begin{tabular}{llS[table-format=1.3(3)]S[table-format=1.3(3)]S[table-format=1.3(3)]S[table-format=1.3(3)]S[table-format=1.3(3)]}
\toprule
& Algorithm/Loss & \multicolumn{5}{l@{}}{Benchmark task ($k$-shot)} \\
\cmidrule(l){3-7}
{} & {} & {1} & {5} & {10} & {20} & {100} \\
\cmidrule(lr){1-2} \cmidrule(lr){3-3} \cmidrule(lr){4-4} \cmidrule(lr){5-5} \cmidrule(lr){6-6} \cmidrule(l){7-7}

\multirow[c]{5}{*}{\rotatebox[origin=c]{90}{Accuracy}} & \gls{gl:ce} & 0.559 +- 0.037 & \bfseries 0.634 +- 0.045 & 0.637 +- 0.032 & 0.615 +- 0.028 & \bfseries 0.731 +- 0.020 \\
 & \gls{gl:fl} & 0.579 +- 0.023 & 0.627 +- 0.090 & 0.613 +- 0.072 & 0.652 +- 0.037 & 0.715 +- 0.019 \\
 & \gls{gl:ce} \gls{gl:ros} & 0.559 +- 0.037 & 0.585 +- 0.029 & 0.610 +- 0.048 & 0.651 +- 0.050 & 0.686 +- 0.012 \\
 & \gls{gl:ce} \gls{gl:dirpa} & \bfseries 0.568 +- 0.053 & 0.623 +- 0.033 & \bfseries 0.656 +- 0.073 & \bfseries 0.653 +- 0.068 & 0.731 +- 0.058 \\
 & \gls{gl:fl} \gls{gl:dirpa} & \colorblue \bfseries 0.626 +- 0.065 & \colorblue \bfseries 0.719 +- 0.082 & \colorblue \bfseries 0.726 +- 0.053 & \colorblue \bfseries 0.751 +- 0.047 & \colorblue \bfseries 0.751 +- 0.025 \\
\midrule
\multirow[c]{5}{*}{\rotatebox[origin=c]{90}{F1-score}} & \gls{gl:ce} & 0.238 +- 0.040 & \bfseries 0.296 +- 0.033 & 0.301 +- 0.021 & 0.283 +- 0.019 & \colorblue \bfseries 0.377 +- 0.010 \\
 & \gls{gl:fl} & 0.256 +- 0.024 & 0.295 +- 0.067 & 0.276 +- 0.051 & 0.311 +- 0.020 & 0.366 +- 0.017 \\
 & \gls{gl:ce} \gls{gl:ros} & 0.238 +- 0.040 & 0.277 +- 0.025 & 0.283 +- 0.035 & 0.304 +- 0.031 & 0.338 +- 0.006 \\
 & \gls{gl:ce} \gls{gl:dirpa} & \bfseries 0.258 +- 0.031 & 0.293 +- 0.012 & \bfseries 0.302 +- 0.050 & \bfseries 0.306 +- 0.038 & 0.369 +- 0.021 \\
 & \gls{gl:fl} \gls{gl:dirpa} & \colorblue \bfseries 0.277 +- 0.044 & \colorblue \bfseries 0.352 +- 0.056 & \colorblue \bfseries 0.351 +- 0.037 & \colorblue \bfseries 0.367 +- 0.032 & \bfseries 0.377 +- 0.020 \\
\bottomrule
\end{tabular}

%% file: tables/results_revision/LT.tex
\robustify\bfseries
\sisetup{detect-all=true,uncertainty-mode=separate,table-align-uncertainty=true,round-mode=uncertainty,round-precision=2}



\begin{tabular}{llS[table-format=1.3(3)]S[table-format=1.3(3)]S[table-format=1.3(3)]S[table-format=1.3(3)]S[table-format=1.3(3)]}
\toprule
& Algorithm/Loss & \multicolumn{5}{l@{}}{Benchmark task ($k$-shot)} \\
\cmidrule(l){3-7}
{} & {} & {1} & {5} & {10} & {20} & {100} \\
\cmidrule(lr){1-2} \cmidrule(lr){3-3} \cmidrule(lr){4-4} \cmidrule(lr){5-5} \cmidrule(lr){6-6} \cmidrule(l){7-7}

\multirow[c]{5}{*}{\rotatebox[origin=c]{90}{Kappa}} & \gls{gl:ce} & 0.337 +- 0.042 & 0.441 +- 0.021 & 0.470 +- 0.019 & 0.538 +- 0.022 & 0.606 +- 0.007 \\
 & \gls{gl:fl} & 0.375 +- 0.042 & 0.456 +- 0.032 & 0.546 +- 0.046 & 0.577 +- 0.018 & 0.641 +- 0.019 \\
 & \gls{gl:ce} \gls{gl:ros} & 0.337 +- 0.042 & 0.435 +- 0.024 & 0.491 +- 0.021 & 0.550 +- 0.033 & 0.605 +- 0.021 \\
 & \gls{gl:ce} \gls{gl:dirpa} & \bfseries 0.347 +- 0.036 & \bfseries 0.510 +- 0.044 & \bfseries 0.495 +- 0.041 & \bfseries 0.576 +- 0.073 & \bfseries 0.637 +- 0.034 \\
 & \gls{gl:fl} \gls{gl:dirpa} & \colorblue \bfseries 0.445 +- 0.018 & \colorblue \bfseries 0.575 +- 0.020 & \colorblue \bfseries 0.637 +- 0.018 & \colorblue \bfseries 0.640 +- 0.032 & \colorblue \bfseries 0.675 +- 0.018 \\
\midrule
\multirow[c]{5}{*}{\rotatebox[origin=c]{90}{Accuracy}} & \gls{gl:ce} & 0.431 +- 0.061 & 0.525 +- 0.020 & 0.552 +- 0.029 & 0.614 +- 0.026 & 0.682 +- 0.005 \\
 & \gls{gl:fl} & 0.479 +- 0.048 & 0.548 +- 0.043 & 0.632 +- 0.054 & 0.662 +- 0.026 & 0.719 +- 0.020 \\
 & \gls{gl:ce} \gls{gl:ros} & 0.431 +- 0.061 & 0.521 +- 0.024 & 0.572 +- 0.023 & 0.627 +- 0.035 & 0.678 +- 0.020 \\
 & \gls{gl:ce} \gls{gl:dirpa} & \bfseries 0.436 +- 0.040 & \bfseries 0.606 +- 0.042 & \bfseries 0.579 +- 0.041 & \bfseries 0.653 +- 0.074 & \bfseries 0.710 +- 0.033 \\
 & \gls{gl:fl} \gls{gl:dirpa} & \colorblue \bfseries 0.565 +- 0.028 & \colorblue \bfseries 0.668 +- 0.020 & \colorblue \bfseries 0.730 +- 0.009 & \colorblue \bfseries 0.724 +- 0.030 & \colorblue \bfseries 0.747 +- 0.018 \\
\midrule
\multirow[c]{5}{*}{\rotatebox[origin=c]{90}{F1-score}} & \gls{gl:ce} & 0.233 +- 0.036 & \bfseries 0.331 +- 0.019 & 0.351 +- 0.012 & 0.398 +- 0.014 & 0.432 +- 0.011 \\
 & \gls{gl:fl} & 0.246 +- 0.041 & 0.329 +- 0.022 & \colorblue \bfseries 0.360 +- 0.015 & 0.375 +- 0.039 & 0.431 +- 0.021 \\
 & \gls{gl:ce} \gls{gl:ros} & 0.233 +- 0.036 & 0.321 +- 0.019 & \bfseries 0.356 +- 0.010 & 0.396 +- 0.004 & \bfseries 0.434 +- 0.017 \\
 & \gls{gl:ce} \gls{gl:dirpa} & \bfseries 0.248 +- 0.028 & 0.328 +- 0.025 & 0.345 +- 0.024 & \colorblue \bfseries 0.401 +- 0.023 & 0.427 +- 0.025 \\
 & \gls{gl:fl} \gls{gl:dirpa} & \colorblue \bfseries 0.251 +- 0.031 & \colorblue \bfseries 0.337 +- 0.020 & 0.335 +- 0.023 & \bfseries 0.384 +- 0.015 & \colorblue \bfseries 0.437 +- 0.021 \\
\bottomrule
\end{tabular}

%% file: tables/results_revision/parentlv3/LT.tex
\robustify\bfseries
\sisetup{detect-all=true,uncertainty-mode=separate,table-align-uncertainty=true,round-mode=uncertainty,round-precision=2}



\begin{tabular}{llS[table-format=1.3(3)]S[table-format=1.3(3)]S[table-format=1.3(3)]S[table-format=1.3(3)]S[table-format=1.3(3)]}
\toprule
& Algorithm/Loss & \multicolumn{5}{l@{}}{Benchmark task ($k$-shot)} \\
\cmidrule(l){3-7}
{} & {} & {1} & {5} & {10} & {20} & {100} \\
\cmidrule(lr){1-2} \cmidrule(lr){3-3} \cmidrule(lr){4-4} \cmidrule(lr){5-5} \cmidrule(lr){6-6} \cmidrule(l){7-7}

\multirow[c]{5}{*}{\rotatebox[origin=c]{90}{Accuracy}} & \gls{gl:ce} & 0.685 +- 0.067 & 0.720 +- 0.021 & 0.719 +- 0.032 & 0.748 +- 0.035 & 0.809 +- 0.007 \\
 & \gls{gl:fl} & 0.736 +- 0.039 & 0.757 +- 0.046 & 0.794 +- 0.065 & 0.808 +- 0.045 & 0.847 +- 0.028 \\
 & \gls{gl:ce} \gls{gl:ros} & 0.685 +- 0.067 & 0.726 +- 0.027 & 0.740 +- 0.024 & 0.758 +- 0.033 & 0.797 +- 0.017 \\
 & \gls{gl:ce} \gls{gl:dirpa} & \bfseries 0.687 +- 0.030 & \bfseries 0.810 +- 0.025 & \bfseries 0.743 +- 0.041 & \bfseries 0.783 +- 0.072 & \bfseries 0.831 +- 0.032 \\
 & \gls{gl:fl} \gls{gl:dirpa} & \colorblue \bfseries 0.798 +- 0.031 & \colorblue \bfseries 0.837 +- 0.018 & \colorblue \bfseries 0.873 +- 0.013 & \colorblue \bfseries 0.858 +- 0.025 & \colorblue \bfseries 0.860 +- 0.019 \\
\midrule
\multirow[c]{5}{*}{\rotatebox[origin=c]{90}{F1-score}} & \gls{gl:ce} & 0.264 +- 0.034 & 0.330 +- 0.009 & 0.324 +- 0.018 & 0.372 +- 0.029 & 0.431 +- 0.010 \\
 & \gls{gl:fl} & 0.295 +- 0.019 & 0.345 +- 0.026 & 0.367 +- 0.038 & 0.393 +- 0.015 & 0.453 +- 0.014 \\
 & \gls{gl:ce} \gls{gl:ros} & 0.264 +- 0.034 & 0.330 +- 0.015 & 0.329 +- 0.012 & 0.390 +- 0.030 & 0.416 +- 0.021 \\
 & \gls{gl:ce} \gls{gl:dirpa} & \bfseries 0.274 +- 0.014 & \bfseries 0.364 +- 0.012 & \bfseries 0.331 +- 0.029 & \bfseries 0.396 +- 0.040 & \bfseries 0.434 +- 0.010 \\
 & \gls{gl:fl} \gls{gl:dirpa} & \colorblue \bfseries 0.313 +- 0.010 & \colorblue \bfseries 0.377 +- 0.018 & \colorblue \bfseries 0.368 +- 0.039 & \colorblue \bfseries 0.421 +- 0.028 & \colorblue \bfseries 0.454 +- 0.012 \\
\bottomrule
\end{tabular}

%% file: tables/results_revision/PT.tex
\robustify\bfseries
\sisetup{detect-all=true,uncertainty-mode=separate,table-align-uncertainty=true,round-mode=uncertainty,round-precision=3}

\begin{tabular}{llS[table-format=1.3(3)]S[table-format=1.3(3)]S[table-format=1.3(3)]S[table-format=1.3(3)]S[table-format=1.3(3)]}
\toprule
& Algorithm/Loss & \multicolumn{5}{l@{}}{Benchmark task ($k$-shot)} \\
\cmidrule(l){3-7}
{} & {} & {1} & {5} & {10} & {20} & {100} \\
\cmidrule(lr){1-2} \cmidrule(lr){3-3} \cmidrule(lr){4-4} \cmidrule(lr){5-5} \cmidrule(lr){6-6} \cmidrule(l){7-7}

\multirow[c]{5}{*}{\rotatebox[origin=c]{90}{Kappa}} & \gls{gl:ce} & \bfseries 0.049 +- 0.003 & 0.092 +- 0.008 & 0.126 +- 0.006 & 0.165 +- 0.011 & \colorblue \bfseries 0.286 +- 0.017 \\
 & \gls{gl:fl} & 0.043 +- 0.008 & \bfseries 0.083 +- 0.005 & 0.130 +- 0.009 & 0.168 +- 0.018 & 0.262 +- 0.018 \\
 & \gls{gl:ce} \gls{gl:ros} & \bfseries 0.049 +- 0.003 & 0.082 +- 0.008 & 0.128 +- 0.010 & 0.170 +- 0.008 & 0.247 +- 0.010 \\
 & \gls{gl:ce} \gls{gl:dirpa} & 0.049 +- 0.006 & \colorblue \bfseries 0.097 +- 0.007 & \bfseries 0.130 +- 0.010 & \bfseries 0.176 +- 0.011 & 0.282 +- 0.012 \\
 & \gls{gl:fl} \gls{gl:dirpa} & \colorblue \bfseries 0.055 +- 0.012 & 0.082 +- 0.002 & \colorblue \bfseries 0.137 +- 0.010 & \colorblue \bfseries 0.181 +- 0.004 & \bfseries 0.280 +- 0.017 \\
\midrule
\multirow[c]{5}{*}{\rotatebox[origin=c]{90}{Accuracy}} & \gls{gl:ce} & 0.070 +- 0.009 & 0.125 +- 0.015 & 0.162 +- 0.010 & 0.206 +- 0.009 & \colorblue \bfseries 0.343 +- 0.021 \\
 & \gls{gl:fl} & 0.063 +- 0.013 & \bfseries 0.114 +- 0.008 & 0.168 +- 0.011 & 0.217 +- 0.033 & 0.322 +- 0.017 \\
 & \gls{gl:ce} \gls{gl:ros} & 0.070 +- 0.009 & 0.114 +- 0.016 & 0.167 +- 0.013 & 0.210 +- 0.011 & 0.305 +- 0.014 \\
 & \gls{gl:ce} \gls{gl:dirpa} & \bfseries 0.074 +- 0.011 & \colorblue \bfseries 0.137 +- 0.019 & \bfseries 0.170 +- 0.014 & \bfseries 0.223 +- 0.015 & 0.342 +- 0.017 \\
 & \gls{gl:fl} \gls{gl:dirpa} & \colorblue \bfseries 0.085 +- 0.024 & 0.112 +- 0.007 & \colorblue \bfseries 0.182 +- 0.019 & \colorblue \bfseries 0.233 +- 0.020 & \bfseries 0.339 +- 0.024 \\
\midrule
\multirow[c]{5}{*}{\rotatebox[origin=c]{90}{F1-score}} & \gls{gl:ce} & \colorblue \bfseries 0.032 +- 0.006 & \colorblue \bfseries 0.056 +- 0.005 & 0.070 +- 0.003 & 0.089 +- 0.008 & \bfseries 0.132 +- 0.009 \\
 & \gls{gl:fl} & \bfseries 0.032 +- 0.005 & \bfseries 0.052 +- 0.008 & 0.071 +- 0.004 & 0.083 +- 0.010 & 0.130 +- 0.015 \\
 & \gls{gl:ce} \gls{gl:ros} & \colorblue \bfseries 0.032 +- 0.006 & 0.052 +- 0.007 & 0.067 +- 0.007 & 0.088 +- 0.002 & 0.111 +- 0.009 \\
 & \gls{gl:ce} \gls{gl:dirpa} & 0.029 +- 0.009 & 0.052 +- 0.003 & \colorblue \bfseries 0.073 +- 0.004 & \colorblue \bfseries 0.093 +- 0.004 & 0.131 +- 0.011 \\
 & \gls{gl:fl} \gls{gl:dirpa} & 0.028 +- 0.007 & 0.052 +- 0.008 & \bfseries 0.072 +- 0.011 & \bfseries 0.093 +- 0.008 & \colorblue \bfseries 0.139 +- 0.009 \\
\bottomrule
\end{tabular}

%% file: tables/results_revision/parentlv3/PT.tex
\robustify\bfseries
\sisetup{detect-all=true,uncertainty-mode=separate,table-align-uncertainty=true,round-mode=uncertainty,round-precision=3}

\begin{tabular}{llS[table-format=1.3(3)]S[table-format=1.3(3)]S[table-format=1.3(3)]S[table-format=1.3(3)]S[table-format=1.3(3)]}
\toprule
& Algorithm/Loss & \multicolumn{5}{l@{}}{Benchmark task ($k$-shot)} \\
\cmidrule(l){3-7}
{} & {} & {1} & {5} & {10} & {20} & {100} \\
\cmidrule(lr){1-2} \cmidrule(lr){3-3} \cmidrule(lr){4-4} \cmidrule(lr){5-5} \cmidrule(lr){6-6} \cmidrule(l){7-7}

\multirow[c]{5}{*}{\rotatebox[origin=c]{90}{Accuracy}} & \gls{gl:ce} & \colorblue \bfseries 0.313 +- 0.005 & 0.343 +- 0.012 & 0.375 +- 0.012 & 0.420 +- 0.026 & 0.520 +- 0.015 \\
 & \gls{gl:fl} & 0.307 +- 0.017 & \bfseries 0.333 +- 0.012 & 0.387 +- 0.013 & \bfseries 0.432 +- 0.045 & 0.502 +- 0.022 \\
 & \gls{gl:ce} \gls{gl:ros} & \colorblue \bfseries 0.313 +- 0.005 & 0.337 +- 0.020 & 0.389 +- 0.016 & 0.430 +- 0.012 & 0.501 +- 0.016 \\
 & \gls{gl:ce} \gls{gl:dirpa} & 0.308 +- 0.020 & \colorblue \bfseries 0.361 +- 0.011 & \colorblue \bfseries 0.394 +- 0.011 & \colorblue \bfseries 0.436 +- 0.019 & \colorblue \bfseries 0.522 +- 0.016 \\
 & \gls{gl:fl} \gls{gl:dirpa} & \bfseries 0.312 +- 0.033 & 0.328 +- 0.021 & \bfseries 0.390 +- 0.021 & 0.429 +- 0.025 & \bfseries 0.511 +- 0.020 \\
\midrule
\multirow[c]{5}{*}{\rotatebox[origin=c]{90}{F1-score}} & \gls{gl:ce} & \colorblue \bfseries 0.245 +- 0.010 & 0.265 +- 0.008 & 0.297 +- 0.013 & 0.327 +- 0.025 & \colorblue \bfseries 0.440 +- 0.021 \\
 & \gls{gl:fl} & 0.234 +- 0.014 & 0.253 +- 0.012 & 0.310 +- 0.005 & \bfseries 0.348 +- 0.049 & 0.418 +- 0.023 \\
 & \gls{gl:ce} \gls{gl:ros} & \colorblue \bfseries 0.245 +- 0.010 & 0.254 +- 0.019 & 0.309 +- 0.009 & 0.337 +- 0.012 & 0.417 +- 0.019 \\
 & \gls{gl:ce} \gls{gl:dirpa} & 0.229 +- 0.024 & \colorblue \bfseries 0.279 +- 0.013 & \bfseries 0.314 +- 0.015 & \colorblue \bfseries 0.359 +- 0.012 & 0.432 +- 0.013 \\
 & \gls{gl:fl} \gls{gl:dirpa} & \bfseries 0.238 +- 0.034 & \bfseries 0.254 +- 0.007 & \colorblue \bfseries 0.319 +- 0.015 & 0.347 +- 0.023 & \bfseries 0.419 +- 0.018 \\
\bottomrule
\end{tabular}

%% file: tables/results_revision/SK.tex
\robustify\bfseries
\sisetup{detect-all=true,uncertainty-mode=separate,table-align-uncertainty=true,round-mode=uncertainty,round-precision=3}



\begin{tabular}{llS[table-format=1.3(3)]S[table-format=1.3(3)]S[table-format=1.3(3)]S[table-format=1.3(3)]S[table-format=1.3(3)]}
\toprule
& Algorithm/Loss & \multicolumn{5}{l@{}}{Benchmark task ($k$-shot)} \\
\cmidrule(l){3-7}
{} & {} & {1} & {5} & {10} & {20} & {100} \\
\cmidrule(lr){1-2} \cmidrule(lr){3-3} \cmidrule(lr){4-4} \cmidrule(lr){5-5} \cmidrule(lr){6-6} \cmidrule(l){7-7}

\multirow[c]{5}{*}{\rotatebox[origin=c]{90}{Kappa}} & \gls{gl:ce} & \colorblue \bfseries 0.185 +- 0.009 & 0.443 +- 0.021 & 0.455 +- 0.034 & 0.477 +- 0.027 & 0.565 +- 0.021 \\
 & \gls{gl:fl} & 0.175 +- 0.008 & 0.418 +- 0.036 & 0.456 +- 0.021 & 0.458 +- 0.036 & 0.556 +- 0.015 \\
 & \gls{gl:ce} \gls{gl:ros} & \colorblue \bfseries 0.185 +- 0.009 & 0.388 +- 0.020 & 0.398 +- 0.038 & 0.426 +- 0.026 & 0.491 +- 0.027 \\
 & \gls{gl:ce} \gls{gl:dirpa} & 0.182 +- 0.009 & \colorblue \bfseries 0.465 +- 0.031 & \bfseries 0.462 +- 0.038 & \colorblue \bfseries 0.527 +- 0.034 & \colorblue \bfseries 0.579 +- 0.010 \\
 & \gls{gl:fl} \gls{gl:dirpa} & \bfseries 0.177 +- 0.019 & \bfseries 0.449 +- 0.046 & \colorblue \bfseries 0.464 +- 0.036 & \bfseries 0.510 +- 0.024 & \bfseries 0.578 +- 0.025 \\
\midrule
\multirow[c]{5}{*}{\rotatebox[origin=c]{90}{Accuracy}} & \gls{gl:ce} & \colorblue \bfseries 0.209 +- 0.012 & 0.520 +- 0.025 & 0.526 +- 0.038 & 0.558 +- 0.037 & 0.641 +- 0.025 \\
 & \gls{gl:fl} & 0.197 +- 0.009 & 0.492 +- 0.045 & 0.535 +- 0.031 & 0.533 +- 0.041 & 0.632 +- 0.016 \\
 & \gls{gl:ce} \gls{gl:ros} & \colorblue \bfseries 0.209 +- 0.012 & 0.456 +- 0.023 & 0.465 +- 0.047 & 0.496 +- 0.034 & 0.563 +- 0.027 \\
 & \gls{gl:ce} \gls{gl:dirpa} & 0.209 +- 0.013 & \colorblue \bfseries 0.550 +- 0.044 & \bfseries 0.533 +- 0.044 & \colorblue \bfseries 0.611 +- 0.036 & \colorblue \bfseries 0.657 +- 0.011 \\
 & \gls{gl:fl} \gls{gl:dirpa} & \bfseries 0.201 +- 0.024 & \bfseries 0.525 +- 0.062 & \colorblue \bfseries 0.538 +- 0.040 & \bfseries 0.591 +- 0.027 & \bfseries 0.654 +- 0.029 \\
\midrule
\multirow[c]{5}{*}{\rotatebox[origin=c]{90}{F1-score}} & \gls{gl:ce} & \colorblue \bfseries 0.105 +- 0.003 & 0.159 +- 0.015 & 0.182 +- 0.009 & \bfseries 0.188 +- 0.020 & \colorblue \bfseries 0.233 +- 0.021 \\
 & \gls{gl:fl} & \bfseries 0.102 +- 0.004 & 0.154 +- 0.015 & 0.171 +- 0.014 & 0.188 +- 0.005 & \bfseries 0.221 +- 0.019 \\
 & \gls{gl:ce} \gls{gl:ros} & \colorblue \bfseries 0.105 +- 0.003 & 0.154 +- 0.009 & 0.159 +- 0.012 & 0.170 +- 0.008 & 0.213 +- 0.009 \\
 & \gls{gl:ce} \gls{gl:dirpa} & 0.090 +- 0.023 & \bfseries 0.160 +- 0.014 & \colorblue \bfseries 0.191 +- 0.008 & 0.188 +- 0.016 & 0.181 +- 0.021 \\
 & \gls{gl:fl} \gls{gl:dirpa} & 0.087 +- 0.024 & \colorblue \bfseries 0.163 +- 0.012 & \bfseries 0.184 +- 0.016 & \colorblue \bfseries 0.195 +- 0.011 & 0.200 +- 0.035 \\
\bottomrule
\end{tabular}

%% file: tables/results_revision/parentlv3/SK.tex
\robustify\bfseries
\sisetup{detect-all=true,uncertainty-mode=separate,table-align-uncertainty=true,round-mode=uncertainty,round-precision=3}



\begin{tabular}{llS[table-format=1.3(3)]S[table-format=1.3(3)]S[table-format=1.3(3)]S[table-format=1.3(3)]S[table-format=1.3(3)]}
\toprule
& Algorithm/Loss & \multicolumn{5}{l@{}}{Benchmark task ($k$-shot)} \\
\cmidrule(l){3-7}
{} & {} & {1} & {5} & {10} & {20} & {100} \\
\cmidrule(lr){1-2} \cmidrule(lr){3-3} \cmidrule(lr){4-4} \cmidrule(lr){5-5} \cmidrule(lr){6-6} \cmidrule(l){7-7}

\multirow[c]{5}{*}{\rotatebox[origin=c]{90}{Accuracy}} & \gls{gl:ce} & 0.500 +- 0.030 & 0.754 +- 0.029 & 0.746 +- 0.028 & 0.783 +- 0.050 & 0.833 +- 0.032 \\
 & \gls{gl:fl} & 0.496 +- 0.023 & 0.734 +- 0.041 & \colorblue \bfseries 0.773 +- 0.055 & 0.755 +- 0.038 & 0.826 +- 0.016 \\
 & \gls{gl:ce} \gls{gl:ros} & 0.500 +- 0.030 & 0.706 +- 0.020 & 0.709 +- 0.043 & 0.726 +- 0.038 & 0.771 +- 0.020 \\
 & \gls{gl:ce} \gls{gl:dirpa} & \colorblue \bfseries 0.519 +- 0.021 & \colorblue \bfseries 0.785 +- 0.046 & \bfseries 0.752 +- 0.032 & \colorblue \bfseries 0.820 +- 0.025 & \colorblue \bfseries 0.847 +- 0.012 \\
 & \gls{gl:fl} \gls{gl:dirpa} & \bfseries 0.510 +- 0.029 & \bfseries 0.759 +- 0.066 & 0.768 +- 0.035 & \bfseries 0.806 +- 0.028 & \bfseries 0.839 +- 0.030 \\
\midrule
\multirow[c]{5}{*}{\rotatebox[origin=c]{90}{F1-score}} & \gls{gl:ce} & 0.215 +- 0.007 & 0.358 +- 0.015 & \bfseries 0.340 +- 0.021 & 0.361 +- 0.037 & 0.423 +- 0.005 \\
 & \gls{gl:fl} & 0.220 +- 0.002 & 0.335 +- 0.027 & 0.352 +- 0.039 & 0.346 +- 0.018 & 0.411 +- 0.014 \\
 & \gls{gl:ce} \gls{gl:ros} & 0.215 +- 0.007 & 0.330 +- 0.009 & 0.313 +- 0.021 & 0.331 +- 0.022 & 0.373 +- 0.012 \\
 & \gls{gl:ce} \gls{gl:dirpa} & \colorblue \bfseries 0.233 +- 0.014 & \colorblue \bfseries 0.372 +- 0.041 & 0.340 +- 0.022 & \colorblue \bfseries 0.389 +- 0.035 & \colorblue \bfseries 0.428 +- 0.031 \\
 & \gls{gl:fl} \gls{gl:dirpa} & \bfseries 0.221 +- 0.013 & \bfseries 0.367 +- 0.053 & \colorblue \bfseries 0.361 +- 0.015 & \bfseries 0.385 +- 0.027 & \bfseries 0.422 +- 0.035 \\
\bottomrule
\end{tabular}

%% file: tables/results_revision/SI.tex
\robustify\bfseries
\sisetup{detect-all=true,uncertainty-mode=separate,table-align-uncertainty=true,round-mode=uncertainty,round-precision=3}

\begin{tabular}{llS[table-format=1.3(3)]S[table-format=1.3(3)]S[table-format=1.3(3)]S[table-format=1.3(3)]S[table-format=1.3(3)]}
\toprule
& Algorithm/Loss & \multicolumn{5}{l@{}}{Benchmark task ($k$-shot)} \\
\cmidrule(l){3-7}
{} & {} & {1} & {5} & {10} & {20} & {100} \\
\cmidrule(lr){1-2} \cmidrule(lr){3-3} \cmidrule(lr){4-4} \cmidrule(lr){5-5} \cmidrule(lr){6-6} \cmidrule(l){7-7}

\multirow[c]{7}{*}{\rotatebox[origin=c]{90}{Kappa}} & \gls{gl:ce} & 0.216 +- 0.032 & 0.198 +- 0.025 & 0.253 +- 0.012 & 0.288 +- 0.020 & 0.354 +- 0.018 \\
 & \gls{gl:fl} & 0.231 +- 0.030 & 0.212 +- 0.007 & 0.276 +- 0.030 & 0.294 +- 0.005 & 0.347 +- 0.017 \\
 & \gls{gl:ce} \gls{gl:ros} & 0.216 +- 0.032 & 0.180 +- 0.022 & 0.235 +- 0.012 & 0.269 +- 0.017 & 0.293 +- 0.012 \\
 & \gls{gl:ce} \gls{gl:dirpa} & \bfseries 0.287 +- 0.032 & \bfseries 0.314 +- 0.050 & \colorblue \bfseries 0.345 +- 0.025 & \colorblue \bfseries 0.388 +- 0.038 & \colorblue \bfseries 0.412 +- 0.014 \\
 & \gls{gl:fl} \gls{gl:dirpa} & \colorblue \bfseries 0.300 +- 0.030 & \colorblue \bfseries 0.318 +- 0.037 & \bfseries 0.327 +- 0.037 & \bfseries 0.381 +- 0.044 & \bfseries 0.405 +- 0.022 \\
 \arrayrulecolor{gray} 
 \cmidrule(lr){2-7}
 \arrayrulecolor{black}
& \gls{gl:ce} \gls{gl:presto} & 0.039 +- 0.008 & 0.083 +- 0.024 & 0.095 +- 0.021 & 0.127 +- 0.017 & 0.230 +- 0.027 \\
 & \gls{gl:ce} \gls{gl:presto} \gls{gl:dirpa} & \colorteal \bfseries 0.058 +- 0.035 & \colorteal \bfseries 0.093 +- 0.038 & \colorteal \bfseries 0.099 +- 0.036 & \colorteal \bfseries 0.145 +- 0.035 & \colorteal \bfseries 0.244 +- 0.030 \\
\midrule
\multirow[c]{7}{*}{\rotatebox[origin=c]{90}{Accuracy}} & \gls{gl:ce} & 0.305 +- 0.061 & 0.270 +- 0.041 & 0.338 +- 0.019 & 0.382 +- 0.028 & 0.456 +- 0.024 \\
 & \gls{gl:fl} & 0.348 +- 0.040 & 0.308 +- 0.012 & 0.378 +- 0.048 & 0.406 +- 0.021 & 0.444 +- 0.022 \\
 & \gls{gl:ce} \gls{gl:ros} & 0.305 +- 0.061 & 0.247 +- 0.034 & 0.318 +- 0.018 & 0.359 +- 0.030 & 0.377 +- 0.016 \\
 & \gls{gl:ce} \gls{gl:dirpa} & \bfseries 0.418 +- 0.051 & \bfseries 0.449 +- 0.072 & \colorblue \bfseries 0.491 +- 0.033 & \colorblue \bfseries 0.538 +- 0.057 & \colorblue \bfseries 0.549 +- 0.031 \\
 & \gls{gl:fl} \gls{gl:dirpa} & \colorblue \bfseries 0.438 +- 0.048 & \colorblue \bfseries 0.454 +- 0.051 & \bfseries 0.450 +- 0.053 & \bfseries 0.522 +- 0.066 & \bfseries 0.523 +- 0.031 \\
 \arrayrulecolor{gray} 
 \cmidrule(lr){2-7}
 \arrayrulecolor{black}
& \gls{gl:ce} \gls{gl:presto} & 0.062 +- 0.025 & 0.166 +- 0.065 & 0.166 +- 0.064 & 0.213 +- 0.059 & 0.346 +- 0.056 \\
 & \gls{gl:ce} \gls{gl:presto} \gls{gl:dirpa} & \colorteal \bfseries 0.125 +- 0.119 & \colorteal \bfseries 0.216 +- 0.134 & \colorteal \bfseries 0.230 +- 0.125 & \colorteal \bfseries 0.264 +- 0.105 & \colorteal \bfseries 0.386 +- 0.061 \\
\midrule
\multirow[c]{7}{*}{\rotatebox[origin=c]{90}{F1-score}} & \gls{gl:ce} & 0.071 +- 0.015 & 0.083 +- 0.006 & \bfseries 0.100 +- 0.005 & \colorblue \bfseries 0.106 +- 0.004 & \bfseries 0.129 +- 0.004 \\
 & \gls{gl:fl} & 0.068 +- 0.010 & 0.084 +- 0.002 & \colorblue \bfseries 0.104 +- 0.003 & 0.099 +- 0.011 & \colorblue \bfseries 0.142 +- 0.004 \\
 & \gls{gl:ce} \gls{gl:ros} & 0.071 +- 0.015 & 0.085 +- 0.002 & 0.095 +- 0.006 & 0.101 +- 0.005 & 0.117 +- 0.005 \\
 & \gls{gl:ce} \gls{gl:dirpa} & \bfseries 0.074 +- 0.014 & \bfseries 0.087 +- 0.008 & 0.092 +- 0.005 & 0.106 +- 0.006 & 0.114 +- 0.020 \\
 & \gls{gl:fl} \gls{gl:dirpa} & \colorblue \bfseries 0.076 +- 0.004 & \colorblue \bfseries 0.089 +- 0.007 & 0.099 +- 0.003 & \bfseries 0.101 +- 0.006 & 0.122 +- 0.010 \\
 \arrayrulecolor{gray} 
 \cmidrule(lr){2-7}
 \arrayrulecolor{black}
& \gls{gl:ce} \gls{gl:presto} & \colorteal \bfseries 0.014 +- 0.003 & \colorteal \bfseries 0.018 +- 0.005 & \colorteal \bfseries 0.027 +- 0.006 & \colorteal \bfseries 0.039 +- 0.006 & \colorteal \bfseries 0.065 +- 0.009 \\
 & \gls{gl:ce} \gls{gl:presto} \gls{gl:dirpa} & 0.012 +- 0.002 & 0.013 +- 0.005 & 0.014 +- 0.006 & 0.033 +- 0.008 & 0.058 +- 0.009 \\
\bottomrule
\end{tabular}

%% file: tables/results_revision/parentlv3/SI.tex
\robustify\bfseries
\sisetup{detect-all=true,uncertainty-mode=separate,table-align-uncertainty=true,round-mode=uncertainty,round-precision=3}

\begin{tabular}{llS[table-format=1.3(3)]S[table-format=1.3(3)]S[table-format=1.3(3)]S[table-format=1.3(3)]S[table-format=1.3(3)]}
\toprule
& Algorithm/Loss & \multicolumn{5}{l@{}}{Benchmark task ($k$-shot)} \\
\cmidrule(l){3-7}
{} & {} & {1} & {5} & {10} & {20} & {100} \\
\cmidrule(lr){1-2} \cmidrule(lr){3-3} \cmidrule(lr){4-4} \cmidrule(lr){5-5} \cmidrule(lr){6-6} \cmidrule(l){7-7}

\multirow[c]{7}{*}{\rotatebox[origin=c]{90}{Accuracy}} & \gls{gl:ce} & 0.540 +- 0.053 & 0.511 +- 0.041 & 0.564 +- 0.018 & 0.596 +- 0.022 & 0.656 +- 0.022 \\
 & \gls{gl:fl} & 0.578 +- 0.031 & 0.556 +- 0.021 & 0.610 +- 0.046 & 0.622 +- 0.029 & 0.644 +- 0.015 \\
 & \gls{gl:ce} \gls{gl:ros} & 0.540 +- 0.053 & 0.499 +- 0.030 & 0.553 +- 0.013 & 0.582 +- 0.030 & 0.593 +- 0.027 \\
 & \gls{gl:ce} \gls{gl:dirpa} & \bfseries 0.633 +- 0.036 & \colorblue \bfseries 0.674 +- 0.070 & \colorblue \bfseries 0.711 +- 0.030 & \colorblue \bfseries 0.727 +- 0.053 & \colorblue \bfseries 0.728 +- 0.034 \\
 & \gls{gl:fl} \gls{gl:dirpa} & \colorblue \bfseries 0.640 +- 0.043 & \bfseries 0.669 +- 0.061 & \bfseries 0.662 +- 0.050 & \bfseries 0.716 +- 0.056 & \bfseries 0.697 +- 0.028 \\
\arrayrulecolor{gray} 
 \cmidrule(lr){2-7}
 \arrayrulecolor{black}
& \gls{gl:ce} \gls{gl:presto} & 0.339 +- 0.017 & 0.454 +- 0.067 & 0.443 +- 0.058 & 0.479 +- 0.044 & 0.587 +- 0.054 \\
 & \gls{gl:ce} \gls{gl:presto} \gls{gl:dirpa} & \colorteal \bfseries 0.355 +- 0.034 & \colorteal \bfseries 0.474 +- 0.149 & \colorteal \bfseries 0.510 +- 0.110 & \colorteal \bfseries 0.526 +- 0.068 & \colorteal \bfseries 0.622 +- 0.050 \\
\midrule
\multirow[c]{7}{*}{\rotatebox[origin=c]{90}{F1-score}} & \gls{gl:ce} & 0.257 +- 0.030 & 0.245 +- 0.023 & 0.279 +- 0.006 & 0.287 +- 0.011 & 0.317 +- 0.009 \\
 & \gls{gl:fl} & 0.278 +- 0.011 & 0.268 +- 0.006 & 0.298 +- 0.017 & 0.299 +- 0.016 & 0.313 +- 0.008 \\
 & \gls{gl:ce} \gls{gl:ros} & 0.257 +- 0.030 & 0.237 +- 0.014 & 0.268 +- 0.009 & 0.280 +- 0.014 & 0.289 +- 0.010 \\
 & \gls{gl:ce} \gls{gl:dirpa} & \colorblue \bfseries 0.295 +- 0.016 & \colorblue \bfseries 0.315 +- 0.027 & \colorblue \bfseries 0.336 +- 0.007 & \colorblue \bfseries 0.340 +- 0.019 & \colorblue \bfseries 0.343 +- 0.013 \\
 & \gls{gl:fl} \gls{gl:dirpa} & \bfseries 0.295 +- 0.014 & \bfseries 0.314 +- 0.023 & \bfseries 0.316 +- 0.018 & \bfseries 0.333 +- 0.021 & \bfseries 0.334 +- 0.012 \\
 \arrayrulecolor{gray} 
 \cmidrule(lr){2-7}
 \arrayrulecolor{black}
& \gls{gl:ce} \gls{gl:presto} & 0.140 +- 0.010 & \colorteal \bfseries 0.205 +- 0.025 & 0.206 +- 0.025 & 0.227 +- 0.019 & 0.281 +- 0.014 \\
 & \gls{gl:ce} \gls{gl:presto} \gls{gl:dirpa} & \colorteal \bfseries 0.144 +- 0.020 & 0.204 +- 0.042 & \colorteal \bfseries 0.211 +- 0.018 & \colorteal \bfseries 0.243 +- 0.022 & \colorteal \bfseries 0.293 +- 0.015 \\
\bottomrule
\end{tabular}

%% file: tables/results_revision/ES.tex
\robustify\bfseries
\sisetup{detect-all=true,uncertainty-mode=separate,table-align-uncertainty=true,round-mode=uncertainty,round-precision=2}



\begin{tabular}{llS[table-format=1.3(3)]S[table-format=1.3(3)]S[table-format=1.3(3)]S[table-format=1.3(3)]S[table-format=1.3(3)]}
\toprule
& Algorithm/Loss & \multicolumn{5}{l@{}}{Benchmark task ($k$-shot)} \\
\cmidrule(l){3-7}
{} & {} & {1} & {5} & {10} & {20} & {100} \\
\cmidrule(lr){1-2} \cmidrule(lr){3-3} \cmidrule(lr){4-4} \cmidrule(lr){5-5} \cmidrule(lr){6-6} \cmidrule(l){7-7}

\multirow[c]{7}{*}{\rotatebox[origin=c]{90}{Kappa}} & \gls{gl:ce} & 0.271 +- 0.011 & 0.325 +- 0.017 & 0.341 +- 0.020 & 0.400 +- 0.023 & 0.460 +- 0.013 \\
 & \gls{gl:fl} & 0.284 +- 0.035 & 0.344 +- 0.026 & \bfseries 0.348 +- 0.029 & 0.405 +- 0.007 & \colorblue \bfseries 0.466 +- 0.026 \\
 & \gls{gl:ce} \gls{gl:ros} & 0.271 +- 0.011 & 0.321 +- 0.016 & 0.345 +- 0.023 & 0.402 +- 0.018 & 0.450 +- 0.013 \\
 & \gls{gl:ce} \gls{gl:dirpa} & \colorblue \bfseries 0.341 +- 0.014 & \colorblue \bfseries 0.356 +- 0.020 & \colorblue \bfseries 0.354 +- 0.012 & \colorblue \bfseries 0.422 +- 0.009 & \bfseries 0.462 +- 0.004 \\
 & \gls{gl:fl} \gls{gl:dirpa} & \bfseries 0.331 +- 0.030 & \bfseries 0.349 +- 0.027 & 0.333 +- 0.030 & \bfseries 0.410 +- 0.018 & 0.453 +- 0.030 \\
 \arrayrulecolor{gray} 
 \cmidrule(lr){2-7}
 \arrayrulecolor{black}
& \gls{gl:ce} \gls{gl:presto} & 0.209 +- 0.039 & 0.296 +- 0.008 & 0.286 +- 0.010 & 0.374 +- 0.015 & 0.421 +- 0.010 \\
 & \gls{gl:ce} \gls{gl:presto} \gls{gl:dirpa} & \colorteal \bfseries 0.240 +- 0.066 & \colorteal \bfseries 0.314 +- 0.035 & \colorteal \bfseries 0.308 +- 0.014 & \colorteal \bfseries 0.381 +- 0.013 & \colorteal \bfseries 0.438 +- 0.028 \\
\midrule
\multirow[c]{7}{*}{\rotatebox[origin=c]{90}{Accuracy}} & \gls{gl:ce} & 0.403 +- 0.013 & 0.472 +- 0.023 & 0.493 +- 0.028 & 0.522 +- 0.026 & 0.583 +- 0.016 \\
 & \gls{gl:fl} & 0.424 +- 0.047 & 0.490 +- 0.033 & \bfseries 0.498 +- 0.049 & 0.532 +- 0.011 & \colorblue \bfseries 0.592 +- 0.035 \\
 & \gls{gl:ce} \gls{gl:ros} & 0.403 +- 0.013 & 0.458 +- 0.018 & 0.490 +- 0.044 & 0.532 +- 0.015 & 0.573 +- 0.016 \\
 & \gls{gl:ce} \gls{gl:dirpa} & \colorblue \bfseries 0.525 +- 0.021 & \bfseries 0.507 +- 0.029 & \colorblue \bfseries 0.499 +- 0.020 & \colorblue \bfseries 0.551 +- 0.014 & \bfseries 0.584 +- 0.005 \\
 & \gls{gl:fl} \gls{gl:dirpa} & \bfseries 0.516 +- 0.036 & \colorblue \bfseries 0.510 +- 0.023 & 0.469 +- 0.034 & \bfseries 0.542 +- 0.017 & 0.578 +- 0.033 \\
 \arrayrulecolor{gray} 
 \cmidrule(lr){2-7}
 \arrayrulecolor{black}
& \gls{gl:ce} \gls{gl:presto} & 0.391 +- 0.042 & 0.426 +- 0.010 & 0.419 +- 0.009 & 0.522 +- 0.020 & 0.552 +- 0.012 \\
 & \gls{gl:ce} \gls{gl:presto} \gls{gl:dirpa} & \colorteal \bfseries 0.438 +- 0.058 & \colorteal \bfseries 0.454 +- 0.059 & \colorteal \bfseries 0.475 +- 0.030 & \colorteal \bfseries 0.524 +- 0.019 & \colorteal \bfseries 0.578 +- 0.032 \\
\midrule
\multirow[c]{7}{*}{\rotatebox[origin=c]{90}{F1-score}} & \gls{gl:ce} & 0.247 +- 0.008 & \bfseries 0.292 +- 0.014 & 0.301 +- 0.014 & 0.340 +- 0.016 & 0.398 +- 0.007 \\
 & \gls{gl:fl} & \bfseries 0.254 +- 0.012 & \colorblue \bfseries 0.294 +- 0.021 & 0.299 +- 0.015 & 0.337 +- 0.009 & \bfseries 0.395 +- 0.018 \\
 & \gls{gl:ce} \gls{gl:ros} & 0.247 +- 0.008 & 0.286 +- 0.014 & 0.290 +- 0.020 & 0.340 +- 0.005 & 0.381 +- 0.014 \\
 & \gls{gl:ce} \gls{gl:dirpa} & \colorblue \bfseries 0.271 +- 0.012 & 0.289 +- 0.008 & \bfseries 0.305 +- 0.013 & \colorblue \bfseries 0.353 +- 0.004 & \colorblue \bfseries 0.399 +- 0.007 \\
 & \gls{gl:fl} \gls{gl:dirpa} & 0.252 +- 0.027 & 0.277 +- 0.041 & \colorblue \bfseries 0.310 +- 0.020 & \bfseries 0.340 +- 0.014 & 0.392 +- 0.019 \\
 \arrayrulecolor{gray} 
 \cmidrule(lr){2-7}
 \arrayrulecolor{black}
& \gls{gl:ce} \gls{gl:presto} & \colorteal \bfseries 0.214 +- 0.017 & \colorteal \bfseries 0.289 +- 0.008 & \colorteal \bfseries 0.273 +- 0.024 & \colorteal \bfseries 0.320 +- 0.016 & \colorteal \bfseries 0.362 +- 0.017 \\
 & \gls{gl:ce} \gls{gl:presto} \gls{gl:dirpa} & 0.190 +- 0.036 & 0.269 +- 0.035 & 0.258 +- 0.042 & 0.317 +- 0.010 & 0.357 +- 0.020 \\
\bottomrule
\end{tabular}

%% file: tables/results_revision/parentlv3/ES.tex
\robustify\bfseries
\sisetup{detect-all=true,uncertainty-mode=separate,table-align-uncertainty=true,round-mode=uncertainty,round-precision=2}



\begin{tabular}{llS[table-format=1.3(3)]S[table-format=1.3(3)]S[table-format=1.3(3)]S[table-format=1.3(3)]S[table-format=1.3(3)]}
\toprule
& Algorithm/Loss & \multicolumn{5}{l@{}}{Benchmark task ($k$-shot)} \\
\cmidrule(l){3-7}
{} & {} & {1} & {5} & {10} & {20} & {100} \\
\cmidrule(lr){1-2} \cmidrule(lr){3-3} \cmidrule(lr){4-4} \cmidrule(lr){5-5} \cmidrule(lr){6-6} \cmidrule(l){7-7}

\multirow[c]{7}{*}{\rotatebox[origin=c]{90}{Accuracy}} & \gls{gl:ce} & 0.456 +- 0.014 & 0.518 +- 0.012 & 0.528 +- 0.014 & 0.566 +- 0.021 & 0.618 +- 0.013 \\
 & \gls{gl:fl} & 0.476 +- 0.036 & 0.530 +- 0.032 & \bfseries 0.532 +- 0.031 & 0.577 +- 0.005 & \colorblue \bfseries 0.627 +- 0.025 \\
 & \gls{gl:ce} \gls{gl:ros} & 0.456 +- 0.014 & 0.509 +- 0.022 & 0.530 +- 0.026 & 0.575 +- 0.019 & 0.604 +- 0.015 \\
 & \gls{gl:ce} \gls{gl:dirpa} & \colorblue \bfseries 0.555 +- 0.020 & \colorblue \bfseries 0.539 +- 0.019 & \colorblue \bfseries 0.541 +- 0.015 & \colorblue \bfseries 0.596 +- 0.007 & \bfseries 0.626 +- 0.007 \\
 & \gls{gl:fl} \gls{gl:dirpa} & \bfseries 0.537 +- 0.028 & \bfseries 0.538 +- 0.030 & 0.515 +- 0.018 & \bfseries 0.589 +- 0.018 & 0.612 +- 0.028 \\
 \arrayrulecolor{gray} 
 \cmidrule(lr){2-7}
 \arrayrulecolor{black}
& \gls{gl:ce} \gls{gl:presto} & 0.434 +- 0.031 & 0.481 +- 0.009 & 0.463 +- 0.009 & \colorteal \bfseries 0.559 +- 0.013 & 0.599 +- 0.017 \\
 & \gls{gl:ce} \gls{gl:presto} \gls{gl:dirpa} & \colorteal \bfseries 0.457 +- 0.052 & \colorteal \bfseries 0.488 +- 0.044 & \colorteal \bfseries 0.497 +- 0.025 & 0.557 +- 0.014 & \colorteal \bfseries 0.606 +- 0.025 \\
\midrule
\multirow[c]{7}{*}{\rotatebox[origin=c]{90}{F1-score}} & \gls{gl:ce} & 0.375 +- 0.009 & 0.417 +- 0.012 & 0.422 +- 0.012 & 0.462 +- 0.016 & 0.504 +- 0.007 \\
 & \gls{gl:fl} & 0.389 +- 0.026 & \bfseries 0.419 +- 0.021 & \bfseries 0.427 +- 0.017 & 0.471 +- 0.003 & \bfseries 0.510 +- 0.016 \\
 & \gls{gl:ce} \gls{gl:ros} & 0.375 +- 0.009 & 0.405 +- 0.012 & 0.414 +- 0.008 & 0.466 +- 0.015 & 0.493 +- 0.011 \\
 & \gls{gl:ce} \gls{gl:dirpa} & \colorblue \bfseries 0.423 +- 0.010 & \colorblue \bfseries 0.427 +- 0.011 & \colorblue \bfseries 0.434 +- 0.010 & \colorblue \bfseries 0.478 +- 0.002 & \colorblue \bfseries 0.513 +- 0.007 \\
 & \gls{gl:fl} \gls{gl:dirpa} & \bfseries 0.404 +- 0.019 & 0.416 +- 0.026 & 0.414 +- 0.013 & \bfseries 0.474 +- 0.012 & 0.500 +- 0.020 \\
 \arrayrulecolor{gray} 
 \cmidrule(lr){2-7}
 \arrayrulecolor{black}
& \gls{gl:ce} \gls{gl:presto} & \colorteal \bfseries 0.343 +- 0.003 & \colorteal \bfseries 0.396 +- 0.009 & 0.378 +- 0.012 & 0.450 +- 0.010 & 0.491 +- 0.012 \\
 & \gls{gl:ce} \gls{gl:presto} \gls{gl:dirpa} & 0.327 +- 0.054 & 0.389 +- 0.037 & \colorteal \bfseries 0.397 +- 0.018 & \colorteal \bfseries 0.453 +- 0.007 & \colorteal \bfseries 0.499 +- 0.011 \\
\bottomrule
\end{tabular}

%% file: images/bubble_plots/newformat/bubble_ce_ck.tex
\begin{tikzpicture}
\begin{axis}[
    width=\linewidth,
    height=5.5cm,
    font=\footnotesize\sansmath\sffamily,
    tick label style={font=\footnotesize\sansmath\sffamily},
    xlabel={Gini coefficient ($C_\text{G}$)},
    ylabel={\textbf{Cohen's kappa}\\Bhattacharyya distance ($D_\text{B}$)},
    ylabel style={align=center},
    grid=both,
    grid style={dashed, gray!30},
    xmin=0.6, xmax=0.96,
    ymin=0.2, ymax=1.0,
    legend pos=outer north east,
    legend cell align=left,
    yticklabel style={
        font=\footnotesize\sansmath\sffamily,
        anchor=east,
        yshift=0.5ex 
    },
]

\addplot[forget plot,only marks,mark=*,color=colorEE,fill=colorEE,fill opacity=0.2,mark size=10.64pt] coordinates {(0.919,0.857)};
\addplot[forget plot,only marks,mark=*,color=colorEE,fill=colorEE,fill opacity=0.2,mark size=12.53pt] coordinates {(0.919,0.740)};
\addplot[forget plot,only marks,mark=*,color=colorEE,fill=colorEE,fill opacity=0.2,mark size=11.41pt] coordinates {(0.919,0.691)};
\addplot[forget plot,only marks,mark=*,color=colorEE,fill=colorEE,fill opacity=0.2,mark size=9.58pt] coordinates {(0.919,0.629)};
\addplot[forget plot,only marks,mark=*,color=colorEE,fill=colorEE,fill opacity=0.2,mark size=4.51pt] coordinates {(0.919,0.450)};

\addplot[forget plot,only marks,mark=*,color=colorLV,fill=colorLV,fill opacity=0.2,mark size=0.20pt] coordinates {(0.910,0.736)};
\addplot[forget plot,only marks,mark=*,color=colorLV,fill=colorLV,fill opacity=0.2,mark size=0.08pt] coordinates {(0.910,0.716)};
\addplot[forget plot,only marks,mark=*,color=colorLV,fill=colorLV,fill opacity=0.2,mark size=1.19pt] coordinates {(0.910,0.688)};
\addplot[forget plot,only marks,mark=*,color=colorLV,fill=colorLV,fill opacity=0.2,mark size=1.64pt] coordinates {(0.910,0.655)};
\addplot[forget plot,only marks,mark=x,color=colorLV,fill=colorLV,fill opacity=0.2,mark size=0.57pt] coordinates {(0.910,0.539)};

\addplot[forget plot,only marks,mark=*,color=colorLT,fill=colorLT,fill opacity=0.2,mark size=0.99pt] coordinates {(0.761,0.365)};
\addplot[forget plot,only marks,mark=*,color=colorLT,fill=colorLT,fill opacity=0.2,mark size=6.95pt] coordinates {(0.761,0.365)};
\addplot[forget plot,only marks,mark=*,color=colorLT,fill=colorLT,fill opacity=0.2,mark size=2.50pt] coordinates {(0.761,0.365)};
\addplot[forget plot,only marks,mark=*,color=colorLT,fill=colorLT,fill opacity=0.2,mark size=3.86pt] coordinates {(0.761,0.365)};
\addplot[forget plot,only marks,mark=*,color=colorLT,fill=colorLT,fill opacity=0.2,mark size=3.04pt] coordinates {(0.761,0.365)};

\addplot[forget plot,only marks,mark=*,color=colorPT,fill=colorPT,fill opacity=0.2,mark size=0.05pt] coordinates {(0.881,0.705)};
\addplot[forget plot,only marks,mark=*,color=colorPT,fill=colorPT,fill opacity=0.2,mark size=0.52pt] coordinates {(0.881,0.588)};
\addplot[forget plot,only marks,mark=*,color=colorPT,fill=colorPT,fill opacity=0.2,mark size=0.44pt] coordinates {(0.881,0.552)};
\addplot[forget plot,only marks,mark=*,color=colorPT,fill=colorPT,fill opacity=0.2,mark size=1.11pt] coordinates {(0.881,0.495)};
\addplot[forget plot,only marks,mark=x,color=colorPT,fill=colorPT,fill opacity=0.2,mark size=0.39pt] coordinates {(0.881,0.370)};

\addplot[forget plot,only marks,mark=x,color=colorSK,fill=colorSK,fill opacity=0.2,mark size=0.28pt] coordinates {(0.904,0.773)};
\addplot[forget plot,only marks,mark=*,color=colorSK,fill=colorSK,fill opacity=0.2,mark size=2.20pt] coordinates {(0.904,0.680)};
\addplot[forget plot,only marks,mark=*,color=colorSK,fill=colorSK,fill opacity=0.2,mark size=0.74pt] coordinates {(0.904,0.645)};
\addplot[forget plot,only marks,mark=*,color=colorSK,fill=colorSK,fill opacity=0.2,mark size=5.05pt] coordinates {(0.904,0.605)};
\addplot[forget plot,only marks,mark=*,color=colorSK,fill=colorSK,fill opacity=0.2,mark size=1.40pt] coordinates {(0.904,0.462)};

\addplot[forget plot,only marks,mark=*,color=colorSI,fill=colorSI,fill opacity=0.2,mark size=7.13pt] coordinates {(0.928,0.866)};
\addplot[forget plot,only marks,mark=*,color=colorSI,fill=colorSI,fill opacity=0.2,mark size=11.64pt] coordinates {(0.928,0.804)};
\addplot[forget plot,only marks,mark=*,color=colorSI,fill=colorSI,fill opacity=0.2,mark size=9.18pt] coordinates {(0.928,0.782)};
\addplot[forget plot,only marks,mark=*,color=colorSI,fill=colorSI,fill opacity=0.2,mark size=9.97pt] coordinates {(0.928,0.750)};
\addplot[forget plot,only marks,mark=*,color=colorSI,fill=colorSI,fill opacity=0.2,mark size=5.76pt] coordinates {(0.928,0.621)};

\addplot[forget plot,only marks,mark=*,color=colorES,fill=colorES,fill opacity=0.2,mark size=7.06pt] coordinates {(0.634,0.213)};
\addplot[forget plot,only marks,mark=*,color=colorES,fill=colorES,fill opacity=0.2,mark size=3.10pt] coordinates {(0.634,0.213)};
\addplot[forget plot,only marks,mark=*,color=colorES,fill=colorES,fill opacity=0.2,mark size=1.35pt] coordinates {(0.634,0.213)};
\addplot[forget plot,only marks,mark=*,color=colorES,fill=colorES,fill opacity=0.2,mark size=2.17pt] coordinates {(0.634,0.213)};
\addplot[forget plot,only marks,mark=*,color=colorES,fill=colorES,fill opacity=0.2,mark size=0.22pt] coordinates {(0.634,0.213)};

\addplot[forget plot,only marks,mark=*,color=colorBE,fill=colorBE,fill opacity=0.2,mark size=5.13pt] coordinates {(0.939,0.880)};
\addplot[forget plot,only marks,mark=*,color=colorBE,fill=colorBE,fill opacity=0.2,mark size=9.22pt] coordinates {(0.939,0.850)};
\addplot[forget plot,only marks,mark=*,color=colorBE,fill=colorBE,fill opacity=0.2,mark size=8.88pt] coordinates {(0.939,0.827)};
\addplot[forget plot,only marks,mark=*,color=colorBE,fill=colorBE,fill opacity=0.2,mark size=6.86pt] coordinates {(0.939,0.786)};
\addplot[forget plot,only marks,mark=*,color=colorBE,fill=colorBE,fill opacity=0.2,mark size=5.15pt] coordinates {(0.939,0.637)};


\end{axis}
\end{tikzpicture}

%% file: images/bubble_plots/newformat/bubble_ce_acc.tex
\begin{tikzpicture}
\begin{axis}[
    width=\linewidth,
    height=5.5cm,
    font=\footnotesize\sansmath\sffamily,
    tick label style={font=\footnotesize\sansmath\sffamily},
    xlabel={Gini coefficient ($C_\text{G}$)},
    ylabel={\textbf{Accuracy}\\Bhattacharyya distance ($D_\text{B}$)},
    ylabel style={align=center},
    grid=both,
    grid style={dashed, gray!30},
    xmin=0.6, xmax=0.96,
    ymin=0.2, ymax=1.0,
    legend pos=outer north east,
    legend cell align=left,
    yticklabel style={
        font=\footnotesize\sansmath\sffamily,
        anchor=east,
        yshift=0.5ex 
    },
]

\addplot[forget plot,only marks,mark=*,color=colorEE,fill=colorEE,fill opacity=0.2,mark size=16.85pt] coordinates {(0.919,0.857)};
\addplot[forget plot,only marks,mark=*,color=colorEE,fill=colorEE,fill opacity=0.2,mark size=17.94pt] coordinates {(0.919,0.740)};
\addplot[forget plot,only marks,mark=*,color=colorEE,fill=colorEE,fill opacity=0.2,mark size=15.01pt] coordinates {(0.919,0.691)};
\addplot[forget plot,only marks,mark=*,color=colorEE,fill=colorEE,fill opacity=0.2,mark size=12.18pt] coordinates {(0.919,0.629)};
\addplot[forget plot,only marks,mark=*,color=colorEE,fill=colorEE,fill opacity=0.2,mark size=6.52pt] coordinates {(0.919,0.450)};

\addplot[only marks,mark=*,color=colorSI,fill=colorSI,fill opacity=0.2,mark size=11.26pt] coordinates {(0.928,0.866)};
\addplot[only marks,mark=*,color=colorSI,fill=colorSI,fill opacity=0.2,mark size=17.90pt] coordinates {(0.928,0.804)};
\addplot[only marks,mark=*,color=colorSI,fill=colorSI,fill opacity=0.2,mark size=15.34pt] coordinates {(0.928,0.782)};
\addplot[only marks,mark=*,color=colorSI,fill=colorSI,fill opacity=0.2,mark size=15.56pt] coordinates {(0.928,0.750)};
\addplot[only marks,mark=*,color=colorSI,fill=colorSI,fill opacity=0.2,mark size=9.31pt] coordinates {(0.928,0.621)};

\addplot[forget plot,only marks,mark=*,color=colorLT,fill=colorLT,fill opacity=0.2,mark size=0.48pt] coordinates {(0.761,0.365)};
\addplot[forget plot,only marks,mark=*,color=colorLT,fill=colorLT,fill opacity=0.2,mark size=8.11pt] coordinates {(0.761,0.365)};
\addplot[forget plot,only marks,mark=*,color=colorLT,fill=colorLT,fill opacity=0.2,mark size=2.68pt] coordinates {(0.761,0.365)};
\addplot[forget plot,only marks,mark=*,color=colorLT,fill=colorLT,fill opacity=0.2,mark size=3.94pt] coordinates {(0.761,0.365)};
\addplot[forget plot,only marks,mark=*,color=colorLT,fill=colorLT,fill opacity=0.2,mark size=2.82pt] coordinates {(0.761,0.365)};

\addplot[forget plot,only marks,mark=*,color=colorPT,fill=colorPT,fill opacity=0.2,mark size=0.35pt] coordinates {(0.881,0.705)};
\addplot[forget plot,only marks,mark=*,color=colorPT,fill=colorPT,fill opacity=0.2,mark size=1.21pt] coordinates {(0.881,0.588)};
\addplot[forget plot,only marks,mark=*,color=colorPT,fill=colorPT,fill opacity=0.2,mark size=0.86pt] coordinates {(0.881,0.552)};
\addplot[forget plot,only marks,mark=*,color=colorPT,fill=colorPT,fill opacity=0.2,mark size=1.71pt] coordinates {(0.881,0.495)};
\addplot[forget plot,only marks,mark=x,color=colorPT,fill=colorPT,fill opacity=0.2,mark size=0.06pt] coordinates {(0.881,0.370)};

\addplot[forget plot,only marks,mark=x,color=colorSK,fill=colorSK,fill opacity=0.2,mark size=0.03pt] coordinates {(0.904,0.773)};
\addplot[forget plot,only marks,mark=*,color=colorSK,fill=colorSK,fill opacity=0.2,mark size=3.04pt] coordinates {(0.904,0.680)};
\addplot[forget plot,only marks,mark=*,color=colorSK,fill=colorSK,fill opacity=0.2,mark size=0.65pt] coordinates {(0.904,0.645)};
\addplot[forget plot,only marks,mark=*,color=colorSK,fill=colorSK,fill opacity=0.2,mark size=5.35pt] coordinates {(0.904,0.605)};
\addplot[forget plot,only marks,mark=*,color=colorSK,fill=colorSK,fill opacity=0.2,mark size=1.53pt] coordinates {(0.904,0.462)};

\addplot[forget plot,only marks,mark=*,color=colorLV,fill=colorLV,fill opacity=0.2,mark size=0.95pt] coordinates {(0.910,0.736)};
\addplot[forget plot,only marks,mark=x,color=colorLV,fill=colorLV,fill opacity=0.2,mark size=0.11pt] coordinates {(0.910,0.716)};
\addplot[forget plot,only marks,mark=*,color=colorLV,fill=colorLV,fill opacity=0.2,mark size=2.06pt] coordinates {(0.910,0.688)};
\addplot[forget plot,only marks,mark=*,color=colorLV,fill=colorLV,fill opacity=0.2,mark size=3.53pt] coordinates {(0.910,0.655)};
\addplot[forget plot,only marks,mark=*,color=colorLV,fill=colorLV,fill opacity=0.2,mark size=0.47pt] coordinates {(0.910,0.539)};

\addplot[forget plot,only marks,mark=*,color=colorBE,fill=colorBE,fill opacity=0.2,mark size=6.84pt] coordinates {(0.939,0.880)};
\addplot[forget plot,only marks,mark=*,color=colorBE,fill=colorBE,fill opacity=0.2,mark size=13.60pt] coordinates {(0.939,0.850)};
\addplot[forget plot,only marks,mark=*,color=colorBE,fill=colorBE,fill opacity=0.2,mark size=10.89pt] coordinates {(0.939,0.827)};
\addplot[forget plot,only marks,mark=*,color=colorBE,fill=colorBE,fill opacity=0.2,mark size=7.89pt] coordinates {(0.939,0.786)};
\addplot[forget plot,only marks,mark=*,color=colorBE,fill=colorBE,fill opacity=0.2,mark size=5.95pt] coordinates {(0.939,0.637)};

\addplot[forget plot,only marks,mark=*,color=colorES,fill=colorES,fill opacity=0.2,mark size=12.24pt] coordinates {(0.634,0.213)};
\addplot[forget plot,only marks,mark=*,color=colorES,fill=colorES,fill opacity=0.2,mark size=3.51pt] coordinates {(0.634,0.213)};
\addplot[forget plot,only marks,mark=*,color=colorES,fill=colorES,fill opacity=0.2,mark size=0.61pt] coordinates {(0.634,0.213)};
\addplot[forget plot,only marks,mark=*,color=colorES,fill=colorES,fill opacity=0.2,mark size=2.95pt] coordinates {(0.634,0.213)};
\addplot[forget plot,only marks,mark=*,color=colorES,fill=colorES,fill opacity=0.2,mark size=0.06pt] coordinates {(0.634,0.213)};

\end{axis}
\end{tikzpicture}

%% file: images/bubble_plots/newformat/bubble_ce_f1.tex
\begin{tikzpicture}
\begin{axis}[
    width=\linewidth,
    height=5.5cm,
    font=\footnotesize\sansmath\sffamily,
    tick label style={font=\footnotesize\sansmath\sffamily},
    xlabel={Gini coefficient ($C_\text{G}$)},
    ylabel={\textbf{F1-score}\\Bhattacharyya distance ($D_\text{B}$)},
    ylabel style={align=center},
    grid=both,
    grid style={dashed, gray!30},
    xmin=0.6, xmax=0.96,
    ymin=0.2, ymax=1.0,
    legend pos=outer north east,
    legend cell align=left,
    yticklabel style={
        font=\footnotesize\sansmath\sffamily,
        anchor=east,
        yshift=0.5ex 
    },
]

\addplot[forget plot,only marks,mark=*,color=colorBE,fill=colorBE,fill opacity=0.2,mark size=0.30pt] coordinates {(0.939,0.880)};
\addplot[forget plot,only marks,mark=*,color=colorBE,fill=colorBE,fill opacity=0.2,mark size=0.36pt] coordinates {(0.939,0.850)};
\addplot[forget plot,only marks,mark=x,color=colorBE,fill=colorBE,fill opacity=0.2,mark size=0.53pt] coordinates {(0.939,0.827)};
\addplot[forget plot,only marks,mark=x,color=colorBE,fill=colorBE,fill opacity=0.2,mark size=1.82pt] coordinates {(0.939,0.786)};
\addplot[forget plot,only marks,mark=x,color=colorBE,fill=colorBE,fill opacity=0.2,mark size=2.10pt] coordinates {(0.939,0.637)};

\addplot[forget plot,only marks,mark=*,color=colorEE,fill=colorEE,fill opacity=0.2,mark size=0.10pt] coordinates {(0.919,0.857)};
\addplot[forget plot,only marks,mark=x,color=colorEE,fill=colorEE,fill opacity=0.2,mark size=1.63pt] coordinates {(0.919,0.740)};
\addplot[forget plot,only marks,mark=x,color=colorEE,fill=colorEE,fill opacity=0.2,mark size=1.42pt] coordinates {(0.919,0.691)};
\addplot[forget plot,only marks,mark=x,color=colorEE,fill=colorEE,fill opacity=0.2,mark size=2.21pt] coordinates {(0.919,0.629)};
\addplot[forget plot,only marks,mark=x,color=colorEE,fill=colorEE,fill opacity=0.2,mark size=4.68pt] coordinates {(0.919,0.450)};

\addplot[forget plot,only marks,mark=x,color=colorLV,fill=colorLV,fill opacity=0.2,mark size=0.11pt] coordinates {(0.910,0.736)};
\addplot[forget plot,only marks,mark=*,color=colorLV,fill=colorLV,fill opacity=0.2,mark size=0.27pt] coordinates {(0.910,0.716)};
\addplot[forget plot,only marks,mark=x,color=colorLV,fill=colorLV,fill opacity=0.2,mark size=1.30pt] coordinates {(0.910,0.688)};
\addplot[forget plot,only marks,mark=x,color=colorLV,fill=colorLV,fill opacity=0.2,mark size=1.39pt] coordinates {(0.910,0.655)};
\addplot[forget plot,only marks,mark=x,color=colorLV,fill=colorLV,fill opacity=0.2,mark size=3.05pt] coordinates {(0.910,0.539)};

\addplot[forget plot,only marks,mark=*,color=colorLT,fill=colorLT,fill opacity=0.2,mark size=1.54pt] coordinates {(0.761,0.365)};
\addplot[forget plot,only marks,mark=x,color=colorLT,fill=colorLT,fill opacity=0.2,mark size=0.34pt] coordinates {(0.761,0.365)};
\addplot[forget plot,only marks,mark=x,color=colorLT,fill=colorLT,fill opacity=0.2,mark size=0.63pt] coordinates {(0.761,0.365)};
\addplot[forget plot,only marks,mark=*,color=colorLT,fill=colorLT,fill opacity=0.2,mark size=0.37pt] coordinates {(0.761,0.365)};
\addplot[forget plot,only marks,mark=x,color=colorLT,fill=colorLT,fill opacity=0.2,mark size=0.56pt] coordinates {(0.761,0.365)};

\addplot[forget plot,only marks,mark=x,color=colorPT,fill=colorPT,fill opacity=0.2,mark size=0.36pt] coordinates {(0.881,0.705)};
\addplot[forget plot,only marks,mark=x,color=colorPT,fill=colorPT,fill opacity=0.2,mark size=0.40pt] coordinates {(0.881,0.588)};
\addplot[forget plot,only marks,mark=*,color=colorPT,fill=colorPT,fill opacity=0.2,mark size=0.30pt] coordinates {(0.881,0.552)};
\addplot[forget plot,only marks,mark=*,color=colorPT,fill=colorPT,fill opacity=0.2,mark size=0.47pt] coordinates {(0.881,0.495)};
\addplot[forget plot,only marks,mark=x,color=colorPT,fill=colorPT,fill opacity=0.2,mark size=0.12pt] coordinates {(0.881,0.370)};

\addplot[forget plot,only marks,mark=x,color=colorSK,fill=colorSK,fill opacity=0.2,mark size=1.45pt] coordinates {(0.904,0.773)};
\addplot[forget plot,only marks,mark=*,color=colorSK,fill=colorSK,fill opacity=0.2,mark size=0.07pt] coordinates {(0.904,0.680)};
\addplot[forget plot,only marks,mark=*,color=colorSK,fill=colorSK,fill opacity=0.2,mark size=0.94pt] coordinates {(0.904,0.645)};
\addplot[forget plot,only marks,mark=*,color=colorSK,fill=colorSK,fill opacity=0.2,mark size=0.00pt] coordinates {(0.904,0.605)};
\addplot[forget plot,only marks,mark=x,color=colorSK,fill=colorSK,fill opacity=0.2,mark size=5.19pt] coordinates {(0.904,0.462)};

\addplot[forget plot,only marks,mark=*,color=colorSI,fill=colorSI,fill opacity=0.2,mark size=0.35pt] coordinates {(0.928,0.866)};
\addplot[forget plot,only marks,mark=*,color=colorSI,fill=colorSI,fill opacity=0.2,mark size=0.39pt] coordinates {(0.928,0.804)};
\addplot[forget plot,only marks,mark=x,color=colorSI,fill=colorSI,fill opacity=0.2,mark size=0.80pt] coordinates {(0.928,0.782)};
\addplot[forget plot,only marks,mark=x,color=colorSI,fill=colorSI,fill opacity=0.2,mark size=0.02pt] coordinates {(0.928,0.750)};
\addplot[forget plot,only marks,mark=x,color=colorSI,fill=colorSI,fill opacity=0.2,mark size=1.55pt] coordinates {(0.928,0.621)};

\addplot[forget plot,only marks,mark=*,color=colorES,fill=colorES,fill opacity=0.2,mark size=2.40pt] coordinates {(0.634,0.213)};
\addplot[forget plot,only marks,mark=x,color=colorES,fill=colorES,fill opacity=0.2,mark size=0.27pt] coordinates {(0.634,0.213)};
\addplot[forget plot,only marks,mark=*,color=colorES,fill=colorES,fill opacity=0.2,mark size=0.43pt] coordinates {(0.634,0.213)};
\addplot[forget plot,only marks,mark=*,color=colorES,fill=colorES,fill opacity=0.2,mark size=1.25pt] coordinates {(0.634,0.213)};
\addplot[forget plot,only marks,mark=*,color=colorES,fill=colorES,fill opacity=0.2,mark size=0.09pt] coordinates {(0.634,0.213)};


\end{axis}
\end{tikzpicture}

%% file: images/bubble_plots/newformat/bubble_fl_ck.tex
\begin{tikzpicture}
\begin{axis}[
    width=\linewidth,
    height=5.5cm,
    font=\footnotesize\sansmath\sffamily,
    tick label style={font=\footnotesize\sansmath\sffamily},
    xlabel={Gini coefficient ($C_\text{G}$)},
    ylabel={\phantom{\textbf{Cohen's kappa}}\\Bhattacharyya distance ($D_\text{B}$)},
    ylabel style={align=center},
    grid=both,
    grid style={dashed, gray!30},
    xmin=0.6, xmax=0.96,
    ymin=0.2, ymax=1.0,
    legend pos=outer north east,
    legend cell align=left,
    yticklabel style={
        font=\footnotesize\sansmath\sffamily,
        anchor=east,
        yshift=0.5ex 
    },
]

\addplot[forget plot,only marks,mark=*,color=colorLV,fill=colorLV,fill opacity=0.2,mark size=4.89pt] coordinates {(0.910,0.736)};
\addplot[forget plot,only marks,mark=*,color=colorLV,fill=colorLV,fill opacity=0.2,mark size=10.10pt] coordinates {(0.910,0.716)};
\addplot[forget plot,only marks,mark=*,color=colorLV,fill=colorLV,fill opacity=0.2,mark size=9.53pt] coordinates {(0.910,0.688)};
\addplot[forget plot,only marks,mark=*,color=colorLV,fill=colorLV,fill opacity=0.2,mark size=7.82pt] coordinates {(0.910,0.655)};
\addplot[forget plot,only marks,mark=*,color=colorLV,fill=colorLV,fill opacity=0.2,mark size=2.74pt] coordinates {(0.910,0.539)};

\addplot[forget plot,only marks,mark=*,color=colorLT,fill=colorLT,fill opacity=0.2,mark size=7.01pt] coordinates {(0.761,0.365)};
\addplot[forget plot,only marks,mark=*,color=colorLT,fill=colorLT,fill opacity=0.2,mark size=11.90pt] coordinates {(0.761,0.365)};
\addplot[forget plot,only marks,mark=*,color=colorLT,fill=colorLT,fill opacity=0.2,mark size=9.15pt] coordinates {(0.761,0.365)};
\addplot[forget plot,only marks,mark=*,color=colorLT,fill=colorLT,fill opacity=0.2,mark size=6.27pt] coordinates {(0.761,0.365)};
\addplot[forget plot,only marks,mark=*,color=colorLT,fill=colorLT,fill opacity=0.2,mark size=3.36pt] coordinates {(0.761,0.365)};

\addplot[forget plot,only marks,mark=*,color=colorPT,fill=colorPT,fill opacity=0.2,mark size=1.13pt] coordinates {(0.881,0.705)};
\addplot[forget plot,only marks,mark=x,color=colorPT,fill=colorPT,fill opacity=0.2,mark size=0.18pt] coordinates {(0.881,0.588)};
\addplot[forget plot,only marks,mark=*,color=colorPT,fill=colorPT,fill opacity=0.2,mark size=0.66pt] coordinates {(0.881,0.552)};
\addplot[forget plot,only marks,mark=*,color=colorPT,fill=colorPT,fill opacity=0.2,mark size=1.33pt] coordinates {(0.881,0.495)};
\addplot[forget plot,only marks,mark=*,color=colorPT,fill=colorPT,fill opacity=0.2,mark size=1.73pt] coordinates {(0.881,0.370)};

\addplot[forget plot,only marks,mark=*,color=colorSK,fill=colorSK,fill opacity=0.2,mark size=0.18pt] coordinates {(0.904,0.773)};
\addplot[forget plot,only marks,mark=*,color=colorSK,fill=colorSK,fill opacity=0.2,mark size=3.09pt] coordinates {(0.904,0.680)};
\addplot[forget plot,only marks,mark=*,color=colorSK,fill=colorSK,fill opacity=0.2,mark size=0.80pt] coordinates {(0.904,0.645)};
\addplot[forget plot,only marks,mark=*,color=colorSK,fill=colorSK,fill opacity=0.2,mark size=5.27pt] coordinates {(0.904,0.605)};
\addplot[forget plot,only marks,mark=*,color=colorSK,fill=colorSK,fill opacity=0.2,mark size=2.23pt] coordinates {(0.904,0.462)};

\addplot[forget plot,only marks,mark=*,color=colorSI,fill=colorSI,fill opacity=0.2,mark size=8.23pt] coordinates {(0.928,0.866)};
\addplot[forget plot,only marks,mark=*,color=colorSI,fill=colorSI,fill opacity=0.2,mark size=10.70pt] coordinates {(0.928,0.804)};
\addplot[forget plot,only marks,mark=*,color=colorSI,fill=colorSI,fill opacity=0.2,mark size=6pt] coordinates {(0.928,0.782)};
\addplot[forget plot,only marks,mark=*,color=colorSI,fill=colorSI,fill opacity=0.2,mark size=8.62pt] coordinates {(0.928,0.750)};
\addplot[forget plot,only marks,mark=*,color=colorSI,fill=colorSI,fill opacity=0.2,mark size=5.82pt] coordinates {(0.928,0.621)};

\addplot[forget plot,only marks,mark=*,color=colorES,fill=colorES,fill opacity=0.2,mark size=4.72pt] coordinates {(0.634,0.213)};
\addplot[forget plot,only marks,mark=*,color=colorES,fill=colorES,fill opacity=0.2,mark size=0.56pt] coordinates {(0.634,0.213)};
\addplot[forget plot,only marks,mark=x,color=colorES,fill=colorES,fill opacity=0.2,mark size=1.47pt] coordinates {(0.634,0.213)};
\addplot[forget plot,only marks,mark=*,color=colorES,fill=colorES,fill opacity=0.2,mark size=0.48pt] coordinates {(0.634,0.213)};
\addplot[forget plot,only marks,mark=x,color=colorES,fill=colorES,fill opacity=0.2,mark size=1.31pt] coordinates {(0.634,0.213)};

\addplot[forget plot,only marks,mark=x,color=colorEE,fill=colorEE,fill opacity=0.2,mark size=1.07pt] coordinates {(0.919,0.857)};
\addplot[forget plot,only marks,mark=*,color=colorEE,fill=colorEE,fill opacity=0.2,mark size=5.44pt] coordinates {(0.919,0.740)};
\addplot[forget plot,only marks,mark=*,color=colorEE,fill=colorEE,fill opacity=0.2,mark size=4.10pt] coordinates {(0.919,0.691)};
\addplot[forget plot,only marks,mark=*,color=colorEE,fill=colorEE,fill opacity=0.2,mark size=0.71pt] coordinates {(0.919,0.629)};
\addplot[forget plot,only marks,mark=*,color=colorEE,fill=colorEE,fill opacity=0.2,mark size=1.11pt] coordinates {(0.919,0.450)};

\addplot[forget plot,only marks,mark=*,color=colorBE,fill=colorBE,fill opacity=0.2,mark size=1.95pt] coordinates {(0.939,0.880)};
\addplot[forget plot,only marks,mark=x,color=colorBE,fill=colorBE,fill opacity=0.2,mark size=2.03pt] coordinates {(0.939,0.850)};
\addplot[forget plot,only marks,mark=*,color=colorBE,fill=colorBE,fill opacity=0.2,mark size=2.72pt] coordinates {(0.939,0.827)};
\addplot[forget plot,only marks,mark=*,color=colorBE,fill=colorBE,fill opacity=0.2,mark size=0.41pt] coordinates {(0.939,0.786)};
\addplot[forget plot,only marks,mark=*,color=colorBE,fill=colorBE,fill opacity=0.2,mark size=2.74pt] coordinates {(0.939,0.637)};

\end{axis}
\end{tikzpicture}

%% file: images/bubble_plots/newformat/bubble_fl_acc.tex
\begin{tikzpicture}
\begin{axis}[
    width=\linewidth,
    height=5.5cm,
    font=\footnotesize\sansmath\sffamily,
    tick label style={font=\footnotesize\sansmath\sffamily},
    xlabel={Gini coefficient ($C_\text{G}$)},
    ylabel={\phantom{\textbf{Accuracy}}\\Bhattacharyya distance ($D_\text{B}$)},
    ylabel style={align=center},
    grid=both,
    grid style={dashed, gray!30},
    xmin=0.6, xmax=0.96,
    ymin=0.2, ymax=1.0,
    legend pos=outer north east,
    legend cell align=left,
    yticklabel style={
        font=\footnotesize\sansmath\sffamily,
        anchor=east,
        yshift=0.5ex 
    },
]

\addplot[forget plot,only marks,mark=*,color=colorLV,fill=colorLV,fill opacity=0.2,mark size=7.68pt] coordinates {(0.910,0.736)};
\addplot[forget plot,only marks,mark=*,color=colorLV,fill=colorLV,fill opacity=0.2,mark size=13.46pt] coordinates {(0.910,0.716)};
\addplot[forget plot,only marks,mark=*,color=colorLV,fill=colorLV,fill opacity=0.2,mark size=13.85pt] coordinates {(0.910,0.688)};
\addplot[forget plot,only marks,mark=*,color=colorLV,fill=colorLV,fill opacity=0.2,mark size=10.98pt] coordinates {(0.910,0.655)};
\addplot[forget plot,only marks,mark=*,color=colorLV,fill=colorLV,fill opacity=0.2,mark size=3.98pt] coordinates {(0.910,0.539)};

\addplot[forget plot,only marks,mark=*,color=colorLT,fill=colorLT,fill opacity=0.2,mark size=8.53pt] coordinates {(0.761,0.365)};
\addplot[forget plot,only marks,mark=*,color=colorLT,fill=colorLT,fill opacity=0.2,mark size=12.07pt] coordinates {(0.761,0.365)};
\addplot[forget plot,only marks,mark=*,color=colorLT,fill=colorLT,fill opacity=0.2,mark size=9.80pt] coordinates {(0.761,0.365)};
\addplot[forget plot,only marks,mark=*,color=colorLT,fill=colorLT,fill opacity=0.2,mark size=6.21pt] coordinates {(0.761,0.365)};
\addplot[forget plot,only marks,mark=*,color=colorLT,fill=colorLT,fill opacity=0.2,mark size=2.83pt] coordinates {(0.761,0.365)};

\addplot[forget plot,only marks,mark=*,color=colorPT,fill=colorPT,fill opacity=0.2,mark size=2.20pt] coordinates {(0.881,0.705)};
\addplot[forget plot,only marks,mark=x,color=colorPT,fill=colorPT,fill opacity=0.2,mark size=0.20pt] coordinates {(0.881,0.588)};
\addplot[forget plot,only marks,mark=*,color=colorPT,fill=colorPT,fill opacity=0.2,mark size=1.39pt] coordinates {(0.881,0.552)};
\addplot[forget plot,only marks,mark=*,color=colorPT,fill=colorPT,fill opacity=0.2,mark size=1.57pt] coordinates {(0.881,0.495)};
\addplot[forget plot,only marks,mark=*,color=colorPT,fill=colorPT,fill opacity=0.2,mark size=1.77pt] coordinates {(0.881,0.370)};

\addplot[forget plot,only marks,mark=*,color=colorSK,fill=colorSK,fill opacity=0.2,mark size=0.40pt] coordinates {(0.904,0.773)};
\addplot[forget plot,only marks,mark=*,color=colorSK,fill=colorSK,fill opacity=0.2,mark size=3.27pt] coordinates {(0.904,0.680)};
\addplot[forget plot,only marks,mark=*,color=colorSK,fill=colorSK,fill opacity=0.2,mark size=0.25pt] coordinates {(0.904,0.645)};
\addplot[forget plot,only marks,mark=*,color=colorSK,fill=colorSK,fill opacity=0.2,mark size=5.84pt] coordinates {(0.904,0.605)};
\addplot[forget plot,only marks,mark=*,color=colorSK,fill=colorSK,fill opacity=0.2,mark size=2.25pt] coordinates {(0.904,0.462)};

\addplot[only marks,mark=*,color=colorSI,fill=colorSI,fill opacity=0.2,mark size=11.70pt] coordinates {(0.928,0.866)};
\addplot[only marks,mark=*,color=colorSI,fill=colorSI,fill opacity=0.2,mark size=16.02pt] coordinates {(0.928,0.804)};
\addplot[only marks,mark=*,color=colorSI,fill=colorSI,fill opacity=0.2,mark size=9.42pt] coordinates {(0.928,0.782)};
\addplot[only marks,mark=*,color=colorSI,fill=colorSI,fill opacity=0.2,mark size=12.12pt] coordinates {(0.928,0.750)};
\addplot[only marks,mark=*,color=colorSI,fill=colorSI,fill opacity=0.2,mark size=7.87pt] coordinates {(0.928,0.621)};

\addplot[forget plot,only marks,mark=*,color=colorES,fill=colorES,fill opacity=0.2,mark size=9.21pt] coordinates {(0.634,0.213)};
\addplot[forget plot,only marks,mark=*,color=colorES,fill=colorES,fill opacity=0.2,mark size=2.02pt] coordinates {(0.634,0.213)};
\addplot[forget plot,only marks,mark=x,color=colorES,fill=colorES,fill opacity=0.2,mark size=2.91pt] coordinates {(0.634,0.213)};
\addplot[forget plot,only marks,mark=*,color=colorES,fill=colorES,fill opacity=0.2,mark size=0.98pt] coordinates {(0.634,0.213)};
\addplot[forget plot,only marks,mark=x,color=colorES,fill=colorES,fill opacity=0.2,mark size=1.49pt] coordinates {(0.634,0.213)};

\addplot[forget plot,only marks,mark=*,color=colorBE,fill=colorBE,fill opacity=0.2,mark size=2.42pt] coordinates {(0.939,0.880)};
\addplot[forget plot,only marks,mark=x,color=colorBE,fill=colorBE,fill opacity=0.2,mark size=4.31pt] coordinates {(0.939,0.850)};
\addplot[forget plot,only marks,mark=*,color=colorBE,fill=colorBE,fill opacity=0.2,mark size=1.60pt] coordinates {(0.939,0.827)};
\addplot[forget plot,only marks,mark=*,color=colorBE,fill=colorBE,fill opacity=0.2,mark size=0.71pt] coordinates {(0.939,0.786)};
\addplot[forget plot,only marks,mark=*,color=colorBE,fill=colorBE,fill opacity=0.2,mark size=2.46pt] coordinates {(0.939,0.637)};

\addplot[forget plot,only marks,mark=x,color=colorEE,fill=colorEE,fill opacity=0.2,mark size=1.56pt] coordinates {(0.919,0.857)};
\addplot[forget plot,only marks,mark=*,color=colorEE,fill=colorEE,fill opacity=0.2,mark size=6.88pt] coordinates {(0.919,0.740)};
\addplot[forget plot,only marks,mark=*,color=colorEE,fill=colorEE,fill opacity=0.2,mark size=5.70pt] coordinates {(0.919,0.691)};
\addplot[forget plot,only marks,mark=*,color=colorEE,fill=colorEE,fill opacity=0.2,mark size=1.86pt] coordinates {(0.919,0.629)};
\addplot[forget plot,only marks,mark=*,color=colorEE,fill=colorEE,fill opacity=0.2,mark size=1.56pt] coordinates {(0.919,0.450)};

\end{axis}
\end{tikzpicture}

%% file: images/bubble_plots/newformat/bubble_fl_f1.tex
\begin{tikzpicture}
\begin{axis}[
    width=\linewidth,
    height=5.5cm,
    font=\footnotesize\sansmath\sffamily,
    tick label style={font=\footnotesize\sansmath\sffamily},
    xlabel={Gini coefficient ($C_\text{G}$)},
    ylabel={\phantom{\textbf{F1-score}}\\Bhattacharyya distance ($D_\text{B}$)},
    ylabel style={align=center},
    grid=both,
    grid style={dashed, gray!30},
    xmin=0.6, xmax=0.96,
    ymin=0.2, ymax=1.0,
    legend pos=outer north east,
    legend cell align=left,
    yticklabel style={
        font=\footnotesize\sansmath\sffamily,
        anchor=east,
        yshift=0.5ex 
    },
]

\addplot[forget plot,only marks,mark=x,color=colorBE,fill=colorBE,fill opacity=0.2,mark size=0.42pt] coordinates {(0.939,0.880)};
\addplot[forget plot,only marks,mark=*,color=colorBE,fill=colorBE,fill opacity=0.2,mark size=0.01pt] coordinates {(0.939,0.850)};
\addplot[forget plot,only marks,mark=*,color=colorBE,fill=colorBE,fill opacity=0.2,mark size=0.63pt] coordinates {(0.939,0.827)};
\addplot[forget plot,only marks,mark=x,color=colorBE,fill=colorBE,fill opacity=0.2,mark size=0.67pt] coordinates {(0.939,0.786)};
\addplot[forget plot,only marks,mark=*,color=colorBE,fill=colorBE,fill opacity=0.2,mark size=1.10pt] coordinates {(0.939,0.637)};

\addplot[forget plot,only marks,mark=*,color=colorEE,fill=colorEE,fill opacity=0.2,mark size=0.13pt] coordinates {(0.919,0.857)};
\addplot[forget plot,only marks,mark=*,color=colorEE,fill=colorEE,fill opacity=0.2,mark size=0.31pt] coordinates {(0.919,0.740)};
\addplot[forget plot,only marks,mark=x,color=colorEE,fill=colorEE,fill opacity=0.2,mark size=1.56pt] coordinates {(0.919,0.691)};
\addplot[forget plot,only marks,mark=x,color=colorEE,fill=colorEE,fill opacity=0.2,mark size=1.34pt] coordinates {(0.919,0.629)};
\addplot[forget plot,only marks,mark=x,color=colorEE,fill=colorEE,fill opacity=0.2,mark size=1.42pt] coordinates {(0.919,0.450)};

\addplot[forget plot,only marks,mark=x,color=colorLV,fill=colorLV,fill opacity=0.2,mark size=0.52pt] coordinates {(0.910,0.736)};
\addplot[forget plot,only marks,mark=*,color=colorLV,fill=colorLV,fill opacity=0.2,mark size=0.73pt] coordinates {(0.910,0.716)};
\addplot[forget plot,only marks,mark=x,color=colorLV,fill=colorLV,fill opacity=0.2,mark size=0.85pt] coordinates {(0.910,0.688)};
\addplot[forget plot,only marks,mark=x,color=colorLV,fill=colorLV,fill opacity=0.2,mark size=1.81pt] coordinates {(0.910,0.655)};
\addplot[forget plot,only marks,mark=x,color=colorLV,fill=colorLV,fill opacity=0.2,mark size=2.18pt] coordinates {(0.910,0.539)};

\addplot[forget plot,only marks,mark=*,color=colorLT,fill=colorLT,fill opacity=0.2,mark size=0.51pt] coordinates {(0.761,0.365)};
\addplot[forget plot,only marks,mark=*,color=colorLT,fill=colorLT,fill opacity=0.2,mark size=0.88pt] coordinates {(0.761,0.365)};
\addplot[forget plot,only marks,mark=x,color=colorLT,fill=colorLT,fill opacity=0.2,mark size=2.55pt] coordinates {(0.761,0.365)};
\addplot[forget plot,only marks,mark=*,color=colorLT,fill=colorLT,fill opacity=0.2,mark size=0.90pt] coordinates {(0.761,0.365)};
\addplot[forget plot,only marks,mark=*,color=colorLT,fill=colorLT,fill opacity=0.2,mark size=0.64pt] coordinates {(0.761,0.365)};

\addplot[forget plot,only marks,mark=x,color=colorPT,fill=colorPT,fill opacity=0.2,mark size=0.38pt] coordinates {(0.881,0.705)};
\addplot[forget plot,only marks,mark=*,color=colorPT,fill=colorPT,fill opacity=0.2,mark size=0.04pt] coordinates {(0.881,0.588)};
\addplot[forget plot,only marks,mark=*,color=colorPT,fill=colorPT,fill opacity=0.2,mark size=0.07pt] coordinates {(0.881,0.552)};
\addplot[forget plot,only marks,mark=*,color=colorPT,fill=colorPT,fill opacity=0.2,mark size=1.00pt] coordinates {(0.881,0.495)};
\addplot[forget plot,only marks,mark=*,color=colorPT,fill=colorPT,fill opacity=0.2,mark size=0.93pt] coordinates {(0.881,0.370)};

\addplot[forget plot,only marks,mark=x,color=colorSK,fill=colorSK,fill opacity=0.2,mark size=1.44pt] coordinates {(0.904,0.773)};
\addplot[forget plot,only marks,mark=*,color=colorSK,fill=colorSK,fill opacity=0.2,mark size=0.91pt] coordinates {(0.904,0.680)};
\addplot[forget plot,only marks,mark=*,color=colorSK,fill=colorSK,fill opacity=0.2,mark size=1.22pt] coordinates {(0.904,0.645)};
\addplot[forget plot,only marks,mark=*,color=colorSK,fill=colorSK,fill opacity=0.2,mark size=0.69pt] coordinates {(0.904,0.605)};
\addplot[forget plot,only marks,mark=x,color=colorSK,fill=colorSK,fill opacity=0.2,mark size=2.13pt] coordinates {(0.904,0.462)};

\addplot[forget plot,only marks,mark=*,color=colorSI,fill=colorSI,fill opacity=0.2,mark size=0.83pt] coordinates {(0.928,0.866)};
\addplot[forget plot,only marks,mark=*,color=colorSI,fill=colorSI,fill opacity=0.2,mark size=0.18pt] coordinates {(0.928,0.804)};
\addplot[forget plot,only marks,mark=x,color=colorSI,fill=colorSI,fill opacity=0.2,mark size=0.53pt] coordinates {(0.928,0.782)};
\addplot[forget plot,only marks,mark=*,color=colorSI,fill=colorSI,fill opacity=0.2,mark size=0.17pt] coordinates {(0.928,0.750)};
\addplot[forget plot,only marks,mark=x,color=colorSI,fill=colorSI,fill opacity=0.2,mark size=1.38pt] coordinates {(0.928,0.621)};

\addplot[forget plot,only marks,mark=x,color=colorES,fill=colorES,fill opacity=0.2,mark size=0.12pt] coordinates {(0.634,0.213)};
\addplot[forget plot,only marks,mark=x,color=colorES,fill=colorES,fill opacity=0.2,mark size=1.69pt] coordinates {(0.634,0.213)};
\addplot[forget plot,only marks,mark=*,color=colorES,fill=colorES,fill opacity=0.2,mark size=1.15pt] coordinates {(0.634,0.213)};
\addplot[forget plot,only marks,mark=*,color=colorES,fill=colorES,fill opacity=0.2,mark size=0.29pt] coordinates {(0.634,0.213)};
\addplot[forget plot,only marks,mark=x,color=colorES,fill=colorES,fill opacity=0.2,mark size=0.34pt] coordinates {(0.634,0.213)};

\end{axis}
\end{tikzpicture}

%% file: images/bubble_plots/legend.tex

\begin{tikzpicture}
  \begin{axis}[
    hide axis,
    scale only axis,
    xmin=0, xmax=1,
    ymin=0, ymax=1,
    width=0.95\linewidth,
    height=0.5cm, 
    legend style={                
        draw=none,                
        legend columns=-1, 
        legend cell align={left},
        font=\footnotesize\sffamily,
        column sep=5pt, 
        nodes={inner xsep=2pt}, 
        /tikz/every legend image/.append style={yshift=0.5ex}, 
    },
  ]
    \addlegendimage{only marks,mark=*,color=colorBE,fill=colorBE,fill opacity=0.7}
    \addlegendentry{Belgium}
    \addlegendimage{only marks,mark=*,color=colorEE,fill=colorEE,fill opacity=0.7}
    \addlegendentry{Estonia}
    \addlegendimage{only marks,mark=*,color=colorLV,fill=colorLV,fill opacity=0.7}
    \addlegendentry{Latvia}
    \addlegendimage{only marks,mark=*,color=colorLT,fill=colorLT,fill opacity=0.7}
    \addlegendentry{Lithuania}
    \addlegendimage{only marks,mark=*,color=colorPT,fill=colorPT,fill opacity=0.7}
    \addlegendentry{Portugal}
    \addlegendimage{only marks,mark=*,color=colorSK,fill=colorSK,fill opacity=0.7}
    \addlegendentry{Slovakia}
    \addlegendimage{only marks,mark=*,color=colorSI,fill=colorSI,fill opacity=0.7}
    \addlegendentry{Slovenia}
    \addlegendimage{only marks,mark=*,color=colorES,fill=colorES,fill opacity=0.7}
    \addlegendentry{Spain}
  \end{axis}
\end{tikzpicture}

%% file: images/bubble_plots/newformat/bubble_ce_lv3_acc.tex
\begin{tikzpicture}
\begin{axis}[
    width=\linewidth,
    height=5.5cm,
    font=\footnotesize\sansmath\sffamily,
    tick label style={font=\footnotesize\sansmath\sffamily},
    xlabel={Gini coefficient ($C_\text{G}$)},
    ylabel={\textbf{Accuracy}\\Bhattacharyya distance ($D_\text{B}$)},
    ylabel style={align=center},
    grid=both,
    grid style={dashed, gray!30},
    xmin=0.6, xmax=0.96,
    ymin=0.2, ymax=1.0,
    legend pos=outer north east,
    legend cell align=left,
    yticklabel style={
        font=\footnotesize\sansmath\sffamily,
        anchor=east,
        yshift=0.5ex 
    },
]

\addplot[forget plot,only marks,mark=*,color=colorEE,fill=colorEE,fill opacity=0.2,mark size=15.27pt] coordinates {(0.919,0.857)};
\addplot[forget plot,only marks,mark=*,color=colorEE,fill=colorEE,fill opacity=0.2,mark size=13.42pt] coordinates {(0.919,0.740)};
\addplot[forget plot,only marks,mark=*,color=colorEE,fill=colorEE,fill opacity=0.2,mark size=11.54pt] coordinates {(0.919,0.691)};
\addplot[forget plot,only marks,mark=*,color=colorEE,fill=colorEE,fill opacity=0.2,mark size=10.55pt] coordinates {(0.919,0.629)};
\addplot[forget plot,only marks,mark=*,color=colorEE,fill=colorEE,fill opacity=0.2,mark size=5.56pt] coordinates {(0.919,0.450)};

\addplot[forget plot,only marks,mark=*,color=colorLT,fill=colorLT,fill opacity=0.2,mark size=0.20pt] coordinates {(0.761,0.365)};
\addplot[forget plot,only marks,mark=*,color=colorLT,fill=colorLT,fill opacity=0.2,mark size=8.98pt] coordinates {(0.761,0.365)};
\addplot[forget plot,only marks,mark=*,color=colorLT,fill=colorLT,fill opacity=0.2,mark size=2.43pt] coordinates {(0.761,0.365)};
\addplot[forget plot,only marks,mark=*,color=colorLT,fill=colorLT,fill opacity=0.2,mark size=3.57pt] coordinates {(0.761,0.365)};
\addplot[forget plot,only marks,mark=*,color=colorLT,fill=colorLT,fill opacity=0.2,mark size=2.26pt] coordinates {(0.761,0.365)};

\addplot[forget plot,only marks,mark=x,color=colorPT,fill=colorPT,fill opacity=0.2,mark size=0.45pt] coordinates {(0.881,0.705)};
\addplot[forget plot,only marks,mark=*,color=colorPT,fill=colorPT,fill opacity=0.2,mark size=1.85pt] coordinates {(0.881,0.588)};
\addplot[forget plot,only marks,mark=*,color=colorPT,fill=colorPT,fill opacity=0.2,mark size=1.87pt] coordinates {(0.881,0.552)};
\addplot[forget plot,only marks,mark=*,color=colorPT,fill=colorPT,fill opacity=0.2,mark size=1.62pt] coordinates {(0.881,0.495)};
\addplot[forget plot,only marks,mark=*,color=colorPT,fill=colorPT,fill opacity=0.2,mark size=0.15pt] coordinates {(0.881,0.370)};

\addplot[forget plot,only marks,mark=*,color=colorSK,fill=colorSK,fill opacity=0.2,mark size=1.97pt] coordinates {(0.904,0.773)};
\addplot[forget plot,only marks,mark=*,color=colorSK,fill=colorSK,fill opacity=0.2,mark size=3.04pt] coordinates {(0.904,0.680)};
\addplot[forget plot,only marks,mark=*,color=colorSK,fill=colorSK,fill opacity=0.2,mark size=0.56pt] coordinates {(0.904,0.645)};
\addplot[forget plot,only marks,mark=*,color=colorSK,fill=colorSK,fill opacity=0.2,mark size=3.74pt] coordinates {(0.904,0.605)};
\addplot[forget plot,only marks,mark=*,color=colorSK,fill=colorSK,fill opacity=0.2,mark size=1.38pt] coordinates {(0.904,0.462)};

\addplot[forget plot,only marks,mark=*,color=colorSI,fill=colorSI,fill opacity=0.2,mark size=9.36pt] coordinates {(0.928,0.866)};
\addplot[forget plot,only marks,mark=*,color=colorSI,fill=colorSI,fill opacity=0.2,mark size=16.23pt] coordinates {(0.928,0.804)};
\addplot[forget plot,only marks,mark=*,color=colorSI,fill=colorSI,fill opacity=0.2,mark size=14.67pt] coordinates {(0.928,0.782)};
\addplot[forget plot,only marks,mark=*,color=colorSI,fill=colorSI,fill opacity=0.2,mark size=13.04pt] coordinates {(0.928,0.750)};
\addplot[forget plot,only marks,mark=*,color=colorSI,fill=colorSI,fill opacity=0.2,mark size=7.23pt] coordinates {(0.928,0.621)};

\addplot[forget plot,only marks,mark=*,color=colorBE,fill=colorBE,fill opacity=0.2,mark size=2.46pt] coordinates {(0.939,0.880)};
\addplot[forget plot,only marks,mark=*,color=colorBE,fill=colorBE,fill opacity=0.2,mark size=13.00pt] coordinates {(0.939,0.850)};
\addplot[forget plot,only marks,mark=*,color=colorBE,fill=colorBE,fill opacity=0.2,mark size=10.11pt] coordinates {(0.939,0.827)};
\addplot[forget plot,only marks,mark=*,color=colorBE,fill=colorBE,fill opacity=0.2,mark size=7.39pt] coordinates {(0.939,0.786)};
\addplot[forget plot,only marks,mark=*,color=colorBE,fill=colorBE,fill opacity=0.2,mark size=6.48pt] coordinates {(0.939,0.637)};

\addplot[forget plot,only marks,mark=*,color=colorES,fill=colorES,fill opacity=0.2,mark size=9.91pt] coordinates {(0.634,0.213)};
\addplot[forget plot,only marks,mark=*,color=colorES,fill=colorES,fill opacity=0.2,mark size=2.13pt] coordinates {(0.634,0.213)};
\addplot[forget plot,only marks,mark=*,color=colorES,fill=colorES,fill opacity=0.2,mark size=1.34pt] coordinates {(0.634,0.213)};
\addplot[forget plot,only marks,mark=*,color=colorES,fill=colorES,fill opacity=0.2,mark size=3.03pt] coordinates {(0.634,0.213)};
\addplot[forget plot,only marks,mark=*,color=colorES,fill=colorES,fill opacity=0.2,mark size=0.80pt] coordinates {(0.634,0.213)};

\addplot[forget plot,only marks,mark=*,color=colorLV,fill=colorLV,fill opacity=0.2,mark size=0.92pt] coordinates {(0.910,0.736)};
\addplot[forget plot,only marks,mark=x,color=colorLV,fill=colorLV,fill opacity=0.2,mark size=1.15pt] coordinates {(0.910,0.716)};
\addplot[forget plot,only marks,mark=*,color=colorLV,fill=colorLV,fill opacity=0.2,mark size=1.99pt] coordinates {(0.910,0.688)};
\addplot[forget plot,only marks,mark=*,color=colorLV,fill=colorLV,fill opacity=0.2,mark size=3.79pt] coordinates {(0.910,0.655)};
\addplot[forget plot,only marks,mark=*,color=colorLV,fill=colorLV,fill opacity=0.2,mark size=0.00pt] coordinates {(0.910,0.539)};


\end{axis}
\end{tikzpicture}

%% file: images/bubble_plots/newformat/bubble_ce_lv3_f1.tex
\begin{tikzpicture}
\begin{axis}[
    width=\linewidth,
    height=5.5cm,
    font=\footnotesize\sansmath\sffamily,
    tick label style={font=\footnotesize\sansmath\sffamily},
    xlabel={Gini coefficient ($C_\text{G}$)},
    ylabel={\textbf{F1-score}\\Bhattacharyya distance ($D_\text{B}$)},
    ylabel style={align=center},
    grid=both,
    grid style={dashed, gray!30},
    xmin=0.6, xmax=0.96,
    ymin=0.2, ymax=1.0,
    legend pos=outer north east,
    legend cell align=left,
    yticklabel style={
        font=\footnotesize\sansmath\sffamily,
        anchor=east,
        yshift=0.5ex 
    },
]

\addplot[forget plot,only marks,mark=x,color=colorBE,fill=colorBE,fill opacity=0.2,mark size=0.64pt] coordinates {(0.939,0.880)};
\addplot[forget plot,only marks,mark=*,color=colorBE,fill=colorBE,fill opacity=0.2,mark size=3.94pt] coordinates {(0.939,0.850)};
\addplot[forget plot,only marks,mark=*,color=colorBE,fill=colorBE,fill opacity=0.2,mark size=1.71pt] coordinates {(0.939,0.827)};
\addplot[forget plot,only marks,mark=*,color=colorBE,fill=colorBE,fill opacity=0.2,mark size=0.97pt] coordinates {(0.939,0.786)};
\addplot[forget plot,only marks,mark=x,color=colorBE,fill=colorBE,fill opacity=0.2,mark size=3.53pt] coordinates {(0.939,0.637)};

\addplot[forget plot,only marks,mark=*,color=colorEE,fill=colorEE,fill opacity=0.2,mark size=6.37pt] coordinates {(0.919,0.857)};
\addplot[forget plot,only marks,mark=*,color=colorEE,fill=colorEE,fill opacity=0.2,mark size=5.38pt] coordinates {(0.919,0.740)};
\addplot[forget plot,only marks,mark=*,color=colorEE,fill=colorEE,fill opacity=0.2,mark size=3.85pt] coordinates {(0.919,0.691)};
\addplot[forget plot,only marks,mark=*,color=colorEE,fill=colorEE,fill opacity=0.2,mark size=3.88pt] coordinates {(0.919,0.629)};
\addplot[forget plot,only marks,mark=*,color=colorEE,fill=colorEE,fill opacity=0.2,mark size=0.66pt] coordinates {(0.919,0.450)};

\addplot[forget plot,only marks,mark=*,color=colorLV,fill=colorLV,fill opacity=0.2,mark size=1.94pt] coordinates {(0.910,0.736)};
\addplot[forget plot,only marks,mark=x,color=colorLV,fill=colorLV,fill opacity=0.2,mark size=0.29pt] coordinates {(0.910,0.716)};
\addplot[forget plot,only marks,mark=*,color=colorLV,fill=colorLV,fill opacity=0.2,mark size=0.12pt] coordinates {(0.910,0.688)};
\addplot[forget plot,only marks,mark=*,color=colorLV,fill=colorLV,fill opacity=0.2,mark size=2.33pt] coordinates {(0.910,0.655)};
\addplot[forget plot,only marks,mark=x,color=colorLV,fill=colorLV,fill opacity=0.2,mark size=0.75pt] coordinates {(0.910,0.539)};

\addplot[forget plot,only marks,mark=*,color=colorLT,fill=colorLT,fill opacity=0.2,mark size=1.01pt] coordinates {(0.761,0.365)};
\addplot[forget plot,only marks,mark=*,color=colorLT,fill=colorLT,fill opacity=0.2,mark size=3.37pt] coordinates {(0.761,0.365)};
\addplot[forget plot,only marks,mark=*,color=colorLT,fill=colorLT,fill opacity=0.2,mark size=0.73pt] coordinates {(0.761,0.365)};
\addplot[forget plot,only marks,mark=*,color=colorLT,fill=colorLT,fill opacity=0.2,mark size=2.47pt] coordinates {(0.761,0.365)};
\addplot[forget plot,only marks,mark=*,color=colorLT,fill=colorLT,fill opacity=0.2,mark size=0.28pt] coordinates {(0.761,0.365)};

\addplot[forget plot,only marks,mark=x,color=colorPT,fill=colorPT,fill opacity=0.2,mark size=1.57pt] coordinates {(0.881,0.705)};
\addplot[forget plot,only marks,mark=*,color=colorPT,fill=colorPT,fill opacity=0.2,mark size=1.34pt] coordinates {(0.881,0.588)};
\addplot[forget plot,only marks,mark=*,color=colorPT,fill=colorPT,fill opacity=0.2,mark size=1.67pt] coordinates {(0.881,0.552)};
\addplot[forget plot,only marks,mark=*,color=colorPT,fill=colorPT,fill opacity=0.2,mark size=3.12pt] coordinates {(0.881,0.495)};
\addplot[forget plot,only marks,mark=x,color=colorPT,fill=colorPT,fill opacity=0.2,mark size=0.84pt] coordinates {(0.881,0.370)};

\addplot[forget plot,only marks,mark=*,color=colorSK,fill=colorSK,fill opacity=0.2,mark size=1.75pt] coordinates {(0.904,0.773)};
\addplot[forget plot,only marks,mark=*,color=colorSK,fill=colorSK,fill opacity=0.2,mark size=1.35pt] coordinates {(0.904,0.680)};
\addplot[forget plot,only marks,mark=*,color=colorSK,fill=colorSK,fill opacity=0.2,mark size=0.00pt] coordinates {(0.904,0.645)};
\addplot[forget plot,only marks,mark=*,color=colorSK,fill=colorSK,fill opacity=0.2,mark size=2.80pt] coordinates {(0.904,0.605)};
\addplot[forget plot,only marks,mark=*,color=colorSK,fill=colorSK,fill opacity=0.2,mark size=0.51pt] coordinates {(0.904,0.462)};

\addplot[forget plot,only marks,mark=*,color=colorSI,fill=colorSI,fill opacity=0.2,mark size=3.84pt] coordinates {(0.928,0.866)};
\addplot[forget plot,only marks,mark=*,color=colorSI,fill=colorSI,fill opacity=0.2,mark size=7.05pt] coordinates {(0.928,0.804)};
\addplot[forget plot,only marks,mark=*,color=colorSI,fill=colorSI,fill opacity=0.2,mark size=5.72pt] coordinates {(0.928,0.782)};
\addplot[forget plot,only marks,mark=*,color=colorSI,fill=colorSI,fill opacity=0.2,mark size=5.30pt] coordinates {(0.928,0.750)};
\addplot[forget plot,only marks,mark=*,color=colorSI,fill=colorSI,fill opacity=0.2,mark size=2.59pt] coordinates {(0.928,0.621)};

\addplot[forget plot,only marks,mark=*,color=colorES,fill=colorES,fill opacity=0.2,mark size=4.76pt] coordinates {(0.634,0.213)};
\addplot[forget plot,only marks,mark=*,color=colorES,fill=colorES,fill opacity=0.2,mark size=0.93pt] coordinates {(0.634,0.213)};
\addplot[forget plot,only marks,mark=*,color=colorES,fill=colorES,fill opacity=0.2,mark size=1.12pt] coordinates {(0.634,0.213)};
\addplot[forget plot,only marks,mark=*,color=colorES,fill=colorES,fill opacity=0.2,mark size=1.62pt] coordinates {(0.634,0.213)};
\addplot[forget plot,only marks,mark=*,color=colorES,fill=colorES,fill opacity=0.2,mark size=0.85pt] coordinates {(0.634,0.213)};


\end{axis}
\end{tikzpicture}

%% file: images/bubble_plots/newformat/bubble_fl_lv3_acc.tex
\begin{tikzpicture}
\begin{axis}[
    width=\linewidth,
    height=5.5cm,
    font=\footnotesize\sansmath\sffamily,
    tick label style={font=\footnotesize\sansmath\sffamily},
    xlabel={Gini coefficient ($C_\text{G}$)},
    ylabel={\phantom{\textbf{Accuracy}}\\Bhattacharyya distance ($D_\text{B}$)},
    ylabel style={align=center},
    grid=both,
    grid style={dashed, gray!30},
    xmin=0.6, xmax=0.96,
    ymin=0.2, ymax=1.0,
    legend pos=outer north east,
    legend cell align=left,
    yticklabel style={
        font=\footnotesize\sansmath\sffamily,
        anchor=east,
        yshift=0.5ex 
    },
]

\addplot[forget plot,only marks,mark=*,color=colorLV,fill=colorLV,fill opacity=0.2,mark size=4.64pt] coordinates {(0.910,0.736)};
\addplot[forget plot,only marks,mark=*,color=colorLV,fill=colorLV,fill opacity=0.2,mark size=9.18pt] coordinates {(0.910,0.716)};
\addplot[forget plot,only marks,mark=*,color=colorLV,fill=colorLV,fill opacity=0.2,mark size=11.33pt] coordinates {(0.910,0.688)};
\addplot[forget plot,only marks,mark=*,color=colorLV,fill=colorLV,fill opacity=0.2,mark size=9.85pt] coordinates {(0.910,0.655)};
\addplot[forget plot,only marks,mark=*,color=colorLV,fill=colorLV,fill opacity=0.2,mark size=3.54pt] coordinates {(0.910,0.539)};

\addplot[forget plot,only marks,mark=*,color=colorLT,fill=colorLT,fill opacity=0.2,mark size=6.22pt] coordinates {(0.761,0.365)};
\addplot[forget plot,only marks,mark=*,color=colorLT,fill=colorLT,fill opacity=0.2,mark size=8.03pt] coordinates {(0.761,0.365)};
\addplot[forget plot,only marks,mark=*,color=colorLT,fill=colorLT,fill opacity=0.2,mark size=7.90pt] coordinates {(0.761,0.365)};
\addplot[forget plot,only marks,mark=*,color=colorLT,fill=colorLT,fill opacity=0.2,mark size=4.97pt] coordinates {(0.761,0.365)};
\addplot[forget plot,only marks,mark=*,color=colorLT,fill=colorLT,fill opacity=0.2,mark size=1.32pt] coordinates {(0.761,0.365)};

\addplot[forget plot,only marks,mark=*,color=colorPT,fill=colorPT,fill opacity=0.2,mark size=0.51pt] coordinates {(0.881,0.705)};
\addplot[forget plot,only marks,mark=x,color=colorPT,fill=colorPT,fill opacity=0.2,mark size=0.50pt] coordinates {(0.881,0.588)};
\addplot[forget plot,only marks,mark=*,color=colorPT,fill=colorPT,fill opacity=0.2,mark size=0.30pt] coordinates {(0.881,0.552)};
\addplot[forget plot,only marks,mark=x,color=colorPT,fill=colorPT,fill opacity=0.2,mark size=0.30pt] coordinates {(0.881,0.495)};
\addplot[forget plot,only marks,mark=*,color=colorPT,fill=colorPT,fill opacity=0.2,mark size=0.93pt] coordinates {(0.881,0.370)};

\addplot[forget plot,only marks,mark=*,color=colorSK,fill=colorSK,fill opacity=0.2,mark size=1.39pt] coordinates {(0.904,0.773)};
\addplot[forget plot,only marks,mark=*,color=colorSK,fill=colorSK,fill opacity=0.2,mark size=2.43pt] coordinates {(0.904,0.680)};
\addplot[forget plot,only marks,mark=x,color=colorSK,fill=colorSK,fill opacity=0.2,mark size=0.52pt] coordinates {(0.904,0.645)};
\addplot[forget plot,only marks,mark=*,color=colorSK,fill=colorSK,fill opacity=0.2,mark size=5.06pt] coordinates {(0.904,0.605)};
\addplot[forget plot,only marks,mark=*,color=colorSK,fill=colorSK,fill opacity=0.2,mark size=1.28pt] coordinates {(0.904,0.462)};

\addplot[forget plot,only marks,mark=*,color=colorSI,fill=colorSI,fill opacity=0.2,mark size=8.21pt] coordinates {(0.928,0.866)};
\addplot[forget plot,only marks,mark=*,color=colorSI,fill=colorSI,fill opacity=0.2,mark size=13.84pt] coordinates {(0.928,0.804)};
\addplot[forget plot,only marks,mark=*,color=colorSI,fill=colorSI,fill opacity=0.2,mark size=7.45pt] coordinates {(0.928,0.782)};
\addplot[forget plot,only marks,mark=*,color=colorSI,fill=colorSI,fill opacity=0.2,mark size=10.03pt] coordinates {(0.928,0.750)};
\addplot[forget plot,only marks,mark=*,color=colorSI,fill=colorSI,fill opacity=0.2,mark size=5.51pt] coordinates {(0.928,0.621)};

\addplot[forget plot,only marks,mark=*,color=colorES,fill=colorES,fill opacity=0.2,mark size=6.02pt] coordinates {(0.634,0.213)};
\addplot[forget plot,only marks,mark=*,color=colorES,fill=colorES,fill opacity=0.2,mark size=0.79pt] coordinates {(0.634,0.213)};
\addplot[forget plot,only marks,mark=x,color=colorES,fill=colorES,fill opacity=0.2,mark size=1.79pt] coordinates {(0.634,0.213)};
\addplot[forget plot,only marks,mark=*,color=colorES,fill=colorES,fill opacity=0.2,mark size=1.21pt] coordinates {(0.634,0.213)};
\addplot[forget plot,only marks,mark=x,color=colorES,fill=colorES,fill opacity=0.2,mark size=1.59pt] coordinates {(0.634,0.213)};

\addplot[forget plot,only marks,mark=*,color=colorBE,fill=colorBE,fill opacity=0.2,mark size=0.19pt] coordinates {(0.939,0.880)};
\addplot[forget plot,only marks,mark=x,color=colorBE,fill=colorBE,fill opacity=0.2,mark size=5.61pt] coordinates {(0.939,0.850)};
\addplot[forget plot,only marks,mark=x,color=colorBE,fill=colorBE,fill opacity=0.2,mark size=2.03pt] coordinates {(0.939,0.827)};
\addplot[forget plot,only marks,mark=*,color=colorBE,fill=colorBE,fill opacity=0.2,mark size=1.78pt] coordinates {(0.939,0.786)};
\addplot[forget plot,only marks,mark=*,color=colorBE,fill=colorBE,fill opacity=0.2,mark size=1.93pt] coordinates {(0.939,0.637)};

\addplot[forget plot,only marks,mark=x,color=colorEE,fill=colorEE,fill opacity=0.2,mark size=1.18pt] coordinates {(0.919,0.857)};
\addplot[forget plot,only marks,mark=*,color=colorEE,fill=colorEE,fill opacity=0.2,mark size=4.50pt] coordinates {(0.919,0.740)};
\addplot[forget plot,only marks,mark=*,color=colorEE,fill=colorEE,fill opacity=0.2,mark size=3.86pt] coordinates {(0.919,0.691)};
\addplot[forget plot,only marks,mark=*,color=colorEE,fill=colorEE,fill opacity=0.2,mark size=2.40pt] coordinates {(0.919,0.629)};
\addplot[forget plot,only marks,mark=*,color=colorEE,fill=colorEE,fill opacity=0.2,mark size=1.90pt] coordinates {(0.919,0.450)};

\end{axis}
\end{tikzpicture}

%% file: images/bubble_plots/newformat/bubble_fl_lv3_f1.tex
\begin{tikzpicture}
\begin{axis}[
    width=\linewidth,
    height=5.5cm,
    font=\footnotesize\sansmath\sffamily,
    tick label style={font=\footnotesize\sansmath\sffamily},
    xlabel={Gini coefficient ($C_\text{G}$)},
    ylabel={\phantom{\textbf{F1-score}}\\Bhattacharyya distance ($D_\text{B}$)},
    ylabel style={align=center},
    grid=both,
    grid style={dashed, gray!30},
    xmin=0.6, xmax=0.96,
    ymin=0.2, ymax=1.0,
    legend pos=outer north east,
    legend cell align=left,
    yticklabel style={
        font=\footnotesize\sansmath\sffamily,
        anchor=east,
        yshift=0.5ex 
    },
]

\addplot[forget plot,only marks,mark=x,color=colorBE,fill=colorBE,fill opacity=0.2,mark size=0.17pt] coordinates {(0.939,0.880)};
\addplot[forget plot,only marks,mark=x,color=colorBE,fill=colorBE,fill opacity=0.2,mark size=3.25pt] coordinates {(0.939,0.850)};
\addplot[forget plot,only marks,mark=x,color=colorBE,fill=colorBE,fill opacity=0.2,mark size=0.35pt] coordinates {(0.939,0.827)};
\addplot[forget plot,only marks,mark=x,color=colorBE,fill=colorBE,fill opacity=0.2,mark size=0.78pt] coordinates {(0.939,0.786)};
\addplot[forget plot,only marks,mark=*,color=colorBE,fill=colorBE,fill opacity=0.2,mark size=3.15pt] coordinates {(0.939,0.637)};

\addplot[forget plot,only marks,mark=*,color=colorLV,fill=colorLV,fill opacity=0.2,mark size=2.15pt] coordinates {(0.910,0.736)};
\addplot[forget plot,only marks,mark=*,color=colorLV,fill=colorLV,fill opacity=0.2,mark size=5.69pt] coordinates {(0.910,0.716)};
\addplot[forget plot,only marks,mark=*,color=colorLV,fill=colorLV,fill opacity=0.2,mark size=7.45pt] coordinates {(0.910,0.688)};
\addplot[forget plot,only marks,mark=*,color=colorLV,fill=colorLV,fill opacity=0.2,mark size=5.65pt] coordinates {(0.910,0.655)};
\addplot[forget plot,only marks,mark=*,color=colorLV,fill=colorLV,fill opacity=0.2,mark size=1.11pt] coordinates {(0.910,0.539)};

\addplot[forget plot,only marks,mark=*,color=colorLT,fill=colorLT,fill opacity=0.2,mark size=1.79pt] coordinates {(0.761,0.365)};
\addplot[forget plot,only marks,mark=*,color=colorLT,fill=colorLT,fill opacity=0.2,mark size=3.20pt] coordinates {(0.761,0.365)};
\addplot[forget plot,only marks,mark=*,color=colorLT,fill=colorLT,fill opacity=0.2,mark size=0.07pt] coordinates {(0.761,0.365)};
\addplot[forget plot,only marks,mark=*,color=colorLT,fill=colorLT,fill opacity=0.2,mark size=2.82pt] coordinates {(0.761,0.365)};
\addplot[forget plot,only marks,mark=*,color=colorLT,fill=colorLT,fill opacity=0.2,mark size=0.06pt] coordinates {(0.761,0.365)};

\addplot[forget plot,only marks,mark=*,color=colorPT,fill=colorPT,fill opacity=0.2,mark size=0.37pt] coordinates {(0.881,0.705)};
\addplot[forget plot,only marks,mark=*,color=colorPT,fill=colorPT,fill opacity=0.2,mark size=0.04pt] coordinates {(0.881,0.588)};
\addplot[forget plot,only marks,mark=*,color=colorPT,fill=colorPT,fill opacity=0.2,mark size=0.96pt] coordinates {(0.881,0.552)};
\addplot[forget plot,only marks,mark=x,color=colorPT,fill=colorPT,fill opacity=0.2,mark size=0.12pt] coordinates {(0.881,0.495)};
\addplot[forget plot,only marks,mark=*,color=colorPT,fill=colorPT,fill opacity=0.2,mark size=0.10pt] coordinates {(0.881,0.370)};

\addplot[forget plot,only marks,mark=*,color=colorSK,fill=colorSK,fill opacity=0.2,mark size=0.04pt] coordinates {(0.904,0.773)};
\addplot[forget plot,only marks,mark=*,color=colorSK,fill=colorSK,fill opacity=0.2,mark size=3.16pt] coordinates {(0.904,0.680)};
\addplot[forget plot,only marks,mark=*,color=colorSK,fill=colorSK,fill opacity=0.2,mark size=0.88pt] coordinates {(0.904,0.645)};
\addplot[forget plot,only marks,mark=*,color=colorSK,fill=colorSK,fill opacity=0.2,mark size=3.85pt] coordinates {(0.904,0.605)};
\addplot[forget plot,only marks,mark=*,color=colorSK,fill=colorSK,fill opacity=0.2,mark size=1.11pt] coordinates {(0.904,0.462)};

\addplot[forget plot,only marks,mark=*,color=colorSI,fill=colorSI,fill opacity=0.2,mark size=2.88pt] coordinates {(0.928,0.866)};
\addplot[forget plot,only marks,mark=*,color=colorSI,fill=colorSI,fill opacity=0.2,mark size=5.99pt] coordinates {(0.928,0.804)};
\addplot[forget plot,only marks,mark=*,color=colorSI,fill=colorSI,fill opacity=0.2,mark size=2.66pt] coordinates {(0.928,0.782)};
\addplot[forget plot,only marks,mark=*,color=colorSI,fill=colorSI,fill opacity=0.2,mark size=3.87pt] coordinates {(0.928,0.750)};
\addplot[forget plot,only marks,mark=*,color=colorSI,fill=colorSI,fill opacity=0.2,mark size=2.27pt] coordinates {(0.928,0.621)};

\addplot[forget plot,only marks,mark=*,color=colorES,fill=colorES,fill opacity=0.2,mark size=1.51pt] coordinates {(0.634,0.213)};
\addplot[forget plot,only marks,mark=x,color=colorES,fill=colorES,fill opacity=0.2,mark size=0.32pt] coordinates {(0.634,0.213)};
\addplot[forget plot,only marks,mark=x,color=colorES,fill=colorES,fill opacity=0.2,mark size=1.22pt] coordinates {(0.634,0.213)};
\addplot[forget plot,only marks,mark=*,color=colorES,fill=colorES,fill opacity=0.2,mark size=0.36pt] coordinates {(0.634,0.213)};
\addplot[forget plot,only marks,mark=x,color=colorES,fill=colorES,fill opacity=0.2,mark size=1.03pt] coordinates {(0.634,0.213)};

\addplot[forget plot,only marks,mark=x,color=colorEE,fill=colorEE,fill opacity=0.2,mark size=1.60pt] coordinates {(0.919,0.857)};
\addplot[forget plot,only marks,mark=*,color=colorEE,fill=colorEE,fill opacity=0.2,mark size=2.27pt] coordinates {(0.919,0.740)};
\addplot[forget plot,only marks,mark=*,color=colorEE,fill=colorEE,fill opacity=0.2,mark size=1.02pt] coordinates {(0.919,0.691)};
\addplot[forget plot,only marks,mark=*,color=colorEE,fill=colorEE,fill opacity=0.2,mark size=0.50pt] coordinates {(0.919,0.629)};
\addplot[forget plot,only marks,mark=*,color=colorEE,fill=colorEE,fill opacity=0.2,mark size=0.53pt] coordinates {(0.919,0.450)};

\end{axis}
\end{tikzpicture}

%% file: tables/runtimes/BE.tex
\robustify\bfseries
\sisetup{detect-all=true,uncertainty-mode=separate,table-align-uncertainty=true,round-mode=uncertainty,round-precision=2}

\begin{tabular}[t]{@{}Xrrrrr@{}}
\toprule
algorithm/loss & \multicolumn{5}{X@{}}{benchmark task ($k$-shot)} \\
\cmidrule(l){2-6}
{}  & \multicolumn{1}{X}{1} & \multicolumn{1}{X}{5} & \multicolumn{1}{X}{10} & \multicolumn{1}{X}{20} & \multicolumn{1}{X}{100} \\
\cmidrule(r){1-1} \cmidrule(lr){2-2} \cmidrule(lr){3-3} \cmidrule(lr){4-4} \cmidrule(lr){5-5} \cmidrule(l){6-6}
\gls{gl:ce} & \num{13.289857 +- 8.427037} & \num{11.469213 +- 4.363450} & \num{13.612350 +- 3.053609} & \num{12.309223 +- 2.472891} & \num{19.970963 +- 3.569147} \\
\gls{gl:ce} \gls{gl:dirpa} & \num{9.616047 +- 0.581898} & \num{10.417910 +- 1.413838} & \num{10.119823 +- 0.993733} & \num{12.615223 +- 2.866580} & \num{17.508493 +- 1.727554} \\
\gls{gl:fl} & \num{6.726383 +- 2.291953} & \bfseries \num{6.857050 +- 2.732554} & \num{10.840830 +- 4.093952} & \num{10.562220 +- 2.915440} & \num{13.426667 +- 5.082452} \\
\gls{gl:fl} \gls{gl:dirpa} & \bfseries \num{6.672557 +- 1.450298} & \num{7.142610 +- 3.073632} & \bfseries \num{9.461057 +- 1.558183} & \bfseries \num{8.670053 +- 1.461380} & \bfseries \num{12.663447 +- 1.661755} \\
\bottomrule
\end{tabular}

%% file: tables/runtimes/EE.tex
\robustify\bfseries
\sisetup{detect-all=true,uncertainty-mode=separate,table-align-uncertainty=true,round-mode=uncertainty,round-precision=2}

\begin{tabular}[t]{@{}Xrrrrr@{}}
\toprule
algorithm/loss & \multicolumn{5}{X@{}}{benchmark task ($k$-shot)} \\
\cmidrule(l){2-6}
{}  & \multicolumn{1}{X}{1} & \multicolumn{1}{X}{5} & \multicolumn{1}{X}{10} & \multicolumn{1}{X}{20} & \multicolumn{1}{X}{100} \\
\cmidrule(r){1-1} \cmidrule(lr){2-2} \cmidrule(lr){3-3} \cmidrule(lr){4-4} \cmidrule(lr){5-5} \cmidrule(l){6-6}
\gls{gl:ce} & \num{15.114293 +- 13.133837} & \num{7.352643 +- 0.798790} & \num{8.896243 +- 1.606300} & \num{9.896347 +- 1.570817} & \num{17.153410 +- 5.647102} \\
\gls{gl:ce} \gls{gl:dirpa} & \bfseries \num{9.110967 +- 2.644530} & \bfseries \num{7.040727 +- 1.124475} & \bfseries \num{8.513653 +- 1.809453} & \bfseries \num{8.628690 +- 1.207103} & \bfseries \num{12.403610 +- 2.363247} \\
\gls{gl:fl} & \num{9.097793 +- 2.744415} & \num{47.793257 +- 36.073263} & \num{37.295497 +- 36.027621} & \num{53.842527 +- 34.654026} & \num{31.593223 +- 30.398930} \\
\gls{gl:fl} \gls{gl:dirpa} & \num{51.352183 +- 32.016492} & \num{52.421357 +- 56.770861} & \num{36.577673 +- 34.753552} & \num{21.357673 +- 31.188609} & \num{19.374983 +- 9.001035} \\
\bottomrule
\end{tabular}

%% file: tables/runtimes/LT.tex
\robustify\bfseries
\sisetup{detect-all=true,uncertainty-mode=separate,table-align-uncertainty=true,round-mode=uncertainty,round-precision=2}

\begin{tabular}[t]{@{}Xrrrrr@{}}
\toprule
algorithm/loss & \multicolumn{5}{X@{}}{benchmark task ($k$-shot)} \\
\cmidrule(l){2-6}
{}  & \multicolumn{1}{X}{1} & \multicolumn{1}{X}{5} & \multicolumn{1}{X}{10} & \multicolumn{1}{X}{20} & \multicolumn{1}{X}{100} \\
\cmidrule(r){1-1} \cmidrule(lr){2-2} \cmidrule(lr){3-3} \cmidrule(lr){4-4} \cmidrule(lr){5-5} \cmidrule(l){6-6}
\gls{gl:ce} & \bfseries \num{8.044280 +- 2.712212} & \num{11.501870 +- 6.405188} & \bfseries \num{8.765760 +- 1.987778} & \bfseries \num{7.654137 +- 3.540995} & \num{10.395847 +- 1.682871} \\
\gls{gl:ce} \gls{gl:dirpa} & \num{15.541147 +- 3.231581} & \num{12.152577 +- 3.396038} & \num{13.179780 +- 0.185752} & \num{11.130960 +- 2.382089} & \bfseries \num{10.203390 +- 5.444159} \\
\gls{gl:fl} & \num{14.149937 +- 4.741221} & \num{11.469900 +- 3.070765} & \num{12.451517 +- 4.235579} & \num{13.019090 +- 3.546511} & \num{11.540210 +- 2.920380} \\
\gls{gl:fl} \gls{gl:dirpa} & \num{12.818767 +- 3.097913} & \bfseries \num{11.385537 +- 2.469036} & \num{9.637880 +- 2.430279} & \num{12.292803 +- 1.662080} & \num{11.855353 +- 0.876824} \\
\bottomrule
\end{tabular}

%% file: tables/runtimes/LV.tex
\robustify\bfseries
\sisetup{detect-all=true,uncertainty-mode=separate,table-align-uncertainty=true,round-mode=uncertainty,round-precision=2}

\begin{tabular}[t]{@{}Xrrrrr@{}}
\toprule
algorithm/loss & \multicolumn{5}{X@{}}{benchmark task ($k$-shot)} \\
\cmidrule(l){2-6}
{}  & \multicolumn{1}{X}{1} & \multicolumn{1}{X}{5} & \multicolumn{1}{X}{10} & \multicolumn{1}{X}{20} & \multicolumn{1}{X}{100} \\
\cmidrule(r){1-1} \cmidrule(lr){2-2} \cmidrule(lr){3-3} \cmidrule(lr){4-4} \cmidrule(lr){5-5} \cmidrule(l){6-6}
\gls{gl:ce} & \num{9.260703 +- 1.906495} & \num{12.164123 +- 5.219105} & \num{11.136137 +- 1.216544} & \num{13.004923 +- 2.372812} & \num{19.428480 +- 6.522352} \\
\gls{gl:ce} \gls{gl:dirpa} & \num{9.881603 +- 1.103402} & \num{10.352373 +- 2.155403} & \bfseries \num{8.529727 +- 2.712491} & \num{10.799650 +- 2.958940} & \bfseries \num{13.654000 +- 5.134761} \\
\gls{gl:fl} & \num{9.108660 +- 0.733540} & \num{8.990780 +- 1.093702} & \num{9.304413 +- 0.295043} & \num{11.506287 +- 4.570783} & \num{17.666677 +- 3.567740} \\
\gls{gl:fl} \gls{gl:dirpa} & \bfseries \num{8.983157 +- 1.382304} & \bfseries \num{8.673490 +- 0.917893} & \num{11.096097 +- 1.634672} & \bfseries \num{9.940173 +- 2.497455} & \num{15.615883 +- 1.249053} \\
\bottomrule
\end{tabular}

%% file: tables/runtimes/PT.tex
\robustify\bfseries
\sisetup{detect-all=true,uncertainty-mode=separate,table-align-uncertainty=true,round-mode=uncertainty,round-precision=2}

\begin{tabular}[t]{@{}Xrrrrr@{}}
\toprule
algorithm/loss & \multicolumn{5}{X@{}}{benchmark task ($k$-shot)} \\
\cmidrule(l){2-6}
{}  & \multicolumn{1}{X}{1} & \multicolumn{1}{X}{5} & \multicolumn{1}{X}{10} & \multicolumn{1}{X}{20} & \multicolumn{1}{X}{100} \\
\cmidrule(r){1-1} \cmidrule(lr){2-2} \cmidrule(lr){3-3} \cmidrule(lr){4-4} \cmidrule(lr){5-5} \cmidrule(l){6-6}

\smaller \gls{gl:ce} & \num{6.106120 +- 2.279087} & \bfseries \num{7.078273 +- 1.320107} & \num{7.638620 +- 2.143625} & \bfseries \num{7.073157 +- 2.193797} & \num{11.256137 +- 6.112456} \\
\gls{gl:ce} \gls{gl:dirpa} & \bfseries \num{4.985393 +- 2.996504} & \num{7.376027 +- 2.614819} & \bfseries \num{6.348770 +- 2.135994} & \num{7.485227 +- 1.915216} & \bfseries \num{9.924320 +- 4.455447} \\
\gls{gl:fl}  & \num{21.017800 +- 28.339115} & \num{27.89435 +- 29.843895} & \num{35.990027 +- 38.109397} & \num{49.005560 +- 58.671489} & \num{14.716073 +- 5.491213} \\
\gls{gl:fl} \gls{gl:dirpa} & \num{7.043510 +- 3.451348} & \num{20.804973 +- 31.671793} & \num{23.449157 +- 26.096266} & \num{34.064837 +- 39.974590} & \num{12.887087 +- 8.249299} \\
\bottomrule
\end{tabular}

%% file: tables/runtimes/SK.tex
\robustify\bfseries
\sisetup{detect-all=true,uncertainty-mode=separate,table-align-uncertainty=true,round-mode=uncertainty,round-precision=2}

\begin{tabular}[t]{@{}Xrrrrr@{}}
\toprule
algorithm/loss & \multicolumn{5}{X@{}}{benchmark task ($k$-shot)} \\
\cmidrule(l){2-6}
{}  & \multicolumn{1}{X}{1} & \multicolumn{1}{X}{5} & \multicolumn{1}{X}{10} & \multicolumn{1}{X}{20} & \multicolumn{1}{X}{100} \\
\cmidrule(r){1-1} \cmidrule(lr){2-2} \cmidrule(lr){3-3} \cmidrule(lr){4-4} \cmidrule(lr){5-5} \cmidrule(l){6-6}
\gls{gl:ce} & \num{7.892167 +- 0.820982} & \num{9.969560 +- 3.554896} & \num{9.700100 +- 1.685056} & \num{10.950930 +- 2.928832} & \num{16.145813 +- 5.758777} \\
\gls{gl:ce} \gls{gl:dirpa} & \num{7.176353 +- 1.424698} & \bfseries \num{7.148257 +- 3.316162} & \bfseries \num{8.040400 +- 2.045451} & \bfseries \num{7.555813 +- 2.496166} & \num{14.083130 +- 3.668841} \\
\gls{gl:fl} & \num{9.434570 +- 2.473063} & \num{11.301977 +- 2.598286} & \num{12.092953 +- 6.063345} & \num{12.717100 +- 2.652926} & \num{22.179613 +- 15.588062} \\
\gls{gl:fl} \gls{gl:dirpa} & \bfseries \num{6.637677 +- 2.609201} & \num{8.825837 +- 4.043449} & \num{9.456963 +- 3.426530} & \num{11.444237 +- 1.362571} & \bfseries \num{12.155097 +- 2.278300} \\
\bottomrule
\end{tabular}

%% file: tables/runtimes/SI.tex
\robustify\bfseries
\sisetup{detect-all=true,uncertainty-mode=separate,table-align-uncertainty=true,round-mode=uncertainty,round-precision=2}

\begin{tabular}[t]{@{}Xrrrrr@{}}
\toprule
algorithm/loss & \multicolumn{5}{X@{}}{benchmark task ($k$-shot)} \\
\cmidrule(l){2-6}
{}  & \multicolumn{1}{X}{1} & \multicolumn{1}{X}{5} & \multicolumn{1}{X}{10} & \multicolumn{1}{X}{20} & \multicolumn{1}{X}{100} \\
\cmidrule(r){1-1} \cmidrule(lr){2-2} \cmidrule(lr){3-3} \cmidrule(lr){4-4} \cmidrule(lr){5-5} \cmidrule(l){6-6}
\gls{gl:ce} & \bfseries \num{6.782167 +- 2.156214} & \bfseries \num{9.548483 +- 0.938359} & \bfseries \num{11.378630 +- 4.238550} & \bfseries \num{10.645473 +- 5.584847} & \num{21.102253 +- 6.188302} \\
\gls{gl:ce} \gls{gl:dirpa} & \num{12.463270 +- 2.008205} & \num{12.086047 +- 1.238249} & \num{11.507993 +- 1.089948} & \num{11.931230 +- 4.793913} & \bfseries \num{14.319293 +- 4.398825} \\
\gls{gl:fl} & \num{8.585217 +- 4.065608} & \num{10.547860 +- 2.004268} & \num{11.688130 +- 2.862822} & \num{14.868950 +- 8.515425} & \num{19.446147 +- 9.068916} \\
\gls{gl:fl} \gls{gl:dirpa} & \num{10.181313 +- 2.305979} & \num{12.295833 +- 3.925255} & \num{13.253137 +- 2.696096} & \num{12.952990 +- 1.425347} & \num{17.279127 +- 3.316301} \\
\bottomrule
\end{tabular}

%% file: tables/runtimes/ES.tex
\robustify\bfseries
\sisetup{detect-all=true,uncertainty-mode=separate,table-align-uncertainty=true,round-mode=uncertainty,round-precision=2}

\begin{tabular}[t]{@{}Xrrrrr@{}}
\toprule
algorithm/loss & \multicolumn{5}{X@{}}{benchmark task ($k$-shot)} \\
\cmidrule(l){2-6}
{}  & \multicolumn{1}{X}{1} & \multicolumn{1}{X}{5} & \multicolumn{1}{X}{10} & \multicolumn{1}{X}{20} & \multicolumn{1}{X}{100} \\
\cmidrule(r){1-1} \cmidrule(lr){2-2} \cmidrule(lr){3-3} \cmidrule(lr){4-4} \cmidrule(lr){5-5} \cmidrule(l){6-6}
\gls{gl:ce} & \bfseries \num{6.555870 +- 0.842287} & \bfseries \num{6.420410 +- 0.587695} & \num{7.218750 +- 1.291122} & \num{6.710627 +- 1.368406} & \num{8.282793 +- 3.141645} \\
\gls{gl:ce} \gls{gl:dirpa} & \num{16.003783 +- 6.129671} & \num{31.394197 +- 28.310336} & \num{41.817613 +- 38.153626} & \num{39.423680 +- 32.378595} & \num{52.329827 +- 31.291299} \\
\gls{gl:fl} & \num{8.398957 +- 2.644772} & \num{8.238637 +- 2.344832} & \bfseries \num{6.637683 +- 2.244511} & \bfseries \num{7.226997 +- 1.730941} & \bfseries \num{7.079583 +- 1.902185} \\
\gls{gl:fl} \gls{gl:dirpa} & \num{10.218090 +- 3.601519} & \num{10.818533 +- 3.932724} & \num{11.390987 +- 5.333990} & \num{9.201033 +- 4.338430} & \num{10.061707 +- 4.423198} \\
\bottomrule
\end{tabular}